\documentclass[10pt,journal,twocolumn,cspaper,compsoc]{IEEEtran}
\ifCLASSOPTIONcompsoc
  \usepackage[nocompress]{cite}
\else
  \usepackage{cite}
\fi
\ifCLASSINFOpdf
\else
\fi

\usepackage{graphicx}
\usepackage{amsmath,amssymb} % define this before the line numbering.
\usepackage{color}

\usepackage[utf8]{inputenc}
\usepackage{times}
\usepackage{epsfig}

\usepackage{url}

\usepackage{stfloats}

\usepackage[font=small,labelfont=bf]{caption}

\usepackage{xcolor}
\usepackage{amsfonts}
\usepackage[vlined, ruled]{algorithm2e}
\usepackage{setspace}
\usepackage{caption}
\usepackage{subcaption}
\usepackage{bm}
\usepackage{isomath}
\usepackage{algorithm2e}
\usepackage[numbers]{natbib}
\usepackage{bbold}
\usepackage{dsfont}

\usepackage[pagebackref=true,breaklinks=true,letterpaper=true,colorlinks,bookmarks=false]{hyperref}
\usepackage{footmisc}

\usepackage{multirow}
\usepackage{ragged2e}

\usepackage{pbox}
\usepackage{xpatch}

\usepackage{tcolorbox}

\usepackage{enumitem}

\definecolor{beaublue}{rgb}{0.84, 0.9, 0.95}
\definecolor{blackish}{rgb}{0.2, 0.2, 0.2}

\definecolor{beaublue2}{rgb}{0.84, 0.9, 0.95}
\definecolor{blackish2}{rgb}{0.2, 0.2, 0.2}

\makeatletter
\newcommand\fs@nobottomruled{\def\@fs@cfont{\bfseries}\let\@fs@capt\floatc@ruled
  \def\@fs@pre{}% \hrule height.8pt depth0pt \kern2pt
  \def\@fs@post{}% Formerly \def\@fs@post{\kern2pt\hrule\relax}%
  \def\@fs@mid{\kern2pt\hrule\kern2pt}%
  \let\@fs@iftopcapt\iftrue}
\makeatother

\newcommand{\iu}{{i\mkern1mu}}

\renewcommand\vec[1]{\ensuremath\boldsymbol{#1}}
\renewcommand\cdots{...}

\newcommand{\tY}{\vec{\mathcal{Y}}}

\newcommand{\mW}{\mathbf{W}}

\newcommand{\vy}{\mathbf{y}}
\newcommand{\valpha}{\bm{\alpha}}
\newcommand{\tX}{\vec{\mathcal{X}}}

\newcommand{\mX}{\mathbf{X}}
\newcommand{\vx}{\mathbf{x}}

\newcommand{\mbr}[1]{\mathbb{R}^{#1}}

\newcommand{\vbeta}{\vec{\beta}}
\newcommand{\vv}{\mathbf{v}}

\newcommand{\cE}{{\mathcal{E}}}
\newcommand{\cEH}{{\mathcal{\hat{E}}}}
\newcommand{\tV}{\vec{\mathcal{V}}}
\newcommand{\tVnb}{\mathcal{V}}
\newcommand{\tXnb}{\mathcal{X}}
\newcommand{\tYnb}{\mathcal{Y}}
\newcommand{\tE}{\vec{\mathcal{E}}}
\newcommand{\tEH}{\vec{\mathcal{\hat{E}}}}
\newcommand{\tGH}{\vec{\mathcal{\hat{G}}}}
\newcommand{\tVH}{\vec{\mathcal{\bar{V}}}}
\newcommand{\tVT}{\vec{\mathcal{\hat{V}}}}
\newcommand{\idx}[1]{\mathcal{I}_{#1}}

\newcommand{\semipd}[1]{\mathcal{S}_{+}^{#1}}

\newcommand{\vu}{\mathbf{u}}

\newcommand{\vub}{\mathbf{\bar{u}}}
\newcommand{\vz}{\mathbf{z}}
\newcommand{\vzeta}{\boldsymbol{\zeta}}

\newcommand{\vphi}{\boldsymbol{\phi}}

\newcommand{\bigoh}{\mathcal{O}}

\newcommand{\vj}{\vec{j}}

\newcommand{\enorm}[1]{\left\|{#1}\right\|_2}
\newcommand{\fnorm}[1]{\left\|{#1}\right\|_F}

\newcommand{\set}[1]{\left\{#1\right\}}

\DeclareMathOperator*{\kronstack}{\uparrow\!\otimes}

\DeclareMathOperator*{\avg}{Avg}
\DeclareMathOperator*{\sgn}{Sgn}
\DeclareMathOperator*{\hosvd}{HOSVD}

\DeclareMathOperator*{\diag}{daig}

\newcommand{\suptensor}[1]{\mathfrak{S}^{d}}

\newcommand{\mLambda}{\bm{\lambda}}
\newcommand{\mU}{\bm{U}}
\newcommand{\mV}{\bm{V}}

\newcommand{\piA}{{\Pi_A}}
\newcommand{\piB}{{\Pi_B}}

\newcommand{\cf}[1]{{{f}_{#1}}}

\newcommand{\vs}{\boldsymbol{s}}
\newcommand{\vw}{\boldsymbol{w}}

\def\eg{\emph{e.g.}}

\newcommand{\mygthree}[1]{\boldsymbol{\mathcal{G}}\!\left(\!#1\!\right)}

\newcommand{\tG}{\boldsymbol{\mathcal{G}}}

\newcommand{\vi}{\mathbf{i}}
\newcommand{\idxJ}{\mathcal{J}}

\newcommand{\RaiseBiggBar}[2]{\ensuremath{\raisebox{#1}{$\Bigg|#2$}}}

\renewcommand{\comment}[1]{}

\newcommand\MyLLBrace[2]{$\left\{\rule{0pt}{#1}\right.\text{#2}$}
\newcommand\MyLRBrace[2]{$\left.\rule{0pt}{#1}\right\}\text{#2}$}

\newcommand\revised[1]{{{#1}}}

\makeatletter
\DeclareRobustCommand\onedot{\futurelet\@let@token\bmv@onedotaux}
\def\bmv@onedotaux{\ifx\@let@token.\else.\null\fi\xspace}
\def\eg{\emph{e.g}\onedot} 
\def\ie{\emph{i.e}\onedot} 
\def\cf{\emph{c.f}\onedot} 
\def\etc{\emph{etc}\onedot} \def\vs{\emph{vs}\onedot}
\def\wrt{w.r.t\onedot} 
\def\etal{\emph{et al}\onedot}
\def\bigoh{\mathcal{O}}
\makeatother

\begin{document}
%\title{Tensor Representations via Kernel Linearization for Action Recognition from 3D Skeletons and Videos (with a Twist on Fine-grained Problems)}
\title{Tensor Representations for Action Recognition}

\author{Piotr~Koniusz\qquad Lei~Wang  %,~\IEEEmembership{Member,~IEEE,}
        \qquad~Anoop~Cherian %,~\IEEEmembership{Member,~IEEE}% <-this % stops a space
\IEEEcompsocitemizethanks{\IEEEcompsocthanksitem P. Koniusz and L. Wang are with Data61/CSIRO (former NICTA) and the Australian National University, Canberra, Australia, ACT2601.
\protect\\
E-mail: see http://claret.wikidot.com
\IEEEcompsocthanksitem A. Cherian is with Mitsubishi Electric Research Labs (MERL), Cambridge, MA, USA.
}% <-this % stops an unwanted space
\thanks{Manuscript submitted Dec-2018. Manuscript accepted by TPAMI on 24-Dec-2019.}
}


\markboth{IEEE Transactions on Pattern Analysis and Machine Intelligence,~Submitted, December~2018,~Accepted, December~2019}%
{Shell \MakeLowercase{\textit{et al.}}: Bare Demo of IEEEtran.cls for Computer Society Journals}



\IEEEtitleabstractindextext{%
%\IEEEcompsoctitleabstractindextext{
\begin{justify}
\begin{abstract}
Human actions in video sequences are characterized by the complex interplay between spatial features and their temporal dynamics. In this paper, we propose novel tensor representations for compactly capturing such higher-order relationships between visual features for the task of action recognition. We propose two tensor-based feature representations, viz. (i) \emph{sequence compatibility kernel} (SCK) and (ii) \emph{dynamics compatibility kernel} (DCK). SCK builds on the spatio-temporal correlations between features, whereas DCK explicitly models the action dynamics of a sequence. We also explore generalization of SCK, coined SCK$\,\oplus$, that operates on subsequences to capture the local-global interplay of correlations, which can incorporate  multi-modal inputs \eg., skeleton 3D body-joints and per-frame classifier scores obtained from deep learning models trained on videos. We introduce linearization of these kernels that lead to compact and fast descriptors.  We provide experiments on (i) 3D skeleton action sequences, (ii) fine-grained video sequences, and (iii) standard non-fine-grained videos. %We show state-of-the-art results substantiating the effectiveness of our representations.
As our final representations are tensors that capture higher-order relationships of features, they relate to co-occurrences for robust fine-grained recognition \cite{lin2017improved,koniusz2018deeper}. We use higher-order tensors and so-called Eigenvalue Power Normalization (EPN) which have been long speculated to perform spectral detection of higher-order occurrences \cite{me_tensor2,me_tensor}, thus detecting fine-grained relationships of features rather than merely count features in action sequences. We prove that a tensor of order $r$, built from $Z_*$ dimensional features, coupled with EPN indeed detects if at least one higher-order occurrence is `projected' into one of its $\binom{Z_*}{r}$  subspaces of dim. $r$ represented by the tensor, thus forming a Tensor Power Normalization metric endowed with $\binom{Z_*}{r}$ such `detectors'.
\end{abstract}
\end{justify}
\begin{IEEEkeywords}
CNN, 3D Skeletons, Action Recognition, Aggregation, Kernels, Higher-order Tensors, HOSVD, Power Normalization.
\end{IEEEkeywords}}

\maketitle

%\IEEEdisplaynontitleabstractindextext

% \fi
%s
\IEEEpeerreviewmaketitle

\section{Introduction}\label{sec:intro}
Human action recognition is a central problem in computer vision with potential impact in surveillance, human-robot interaction, elderly assistance systems,  \etc. While there have been significant advancements in this area over the past few years, action recognition in unconstrained settings still remains a challenge. Some papers simplify the problem from using RGB cameras to the use of Microsoft Kinect or the OpenPose library \cite{cao2018openpose} to localize human body-parts, produce moving 3D skeletons~\cite{shotton2013real} and use them  for recognition. However, skeletons can be noisy due to badly localized body-parts, self-occlusions, and sensor errors.
Similarly, a popular strategy of classifying RGB frames into actions followed by average/max-pooling fails as only correlations of some features are informative \cite{joint_preds_mpii,anoop_generalized,action_second}. Such observations motivate the need for higher-order reasoning on 3D skeletons/frame-wise CNN classifier scores taking action recognition toward fine-grained modeling.

Recent approaches which work with skeletons can be mainly divided into two perspectives, namely (i) generative models that assume the skeleton points are produced by a latent dynamic model~\cite{turaga2009locally} corrupted by noise and (ii) discriminative approaches that generate compact representations of sequences on which classifiers are trained~\cite{presti20153d}. Due to the huge configuration space of 3D actions and the unavailability of sufficient training data, discriminative approaches have been more successful. In this line of research, the main idea is to compactly represent the spatio-temporal evolution of 3D skeletons, and later train classifiers on these representations to recognize actions. Fortunately, there is a definitive structure to motions of 3D joints relative to each other due to the connectivity and length constraints of body-parts. Such constraints have been used with the Lie Algebra~\cite{vemulapalli_SE3}, positive definite matrices~\cite{harandi2014bregman,hussein_action}, torus manifold~\cite{elgammal2009tracking}, Hanklet representations~\cite{li2012cross}, \etc. While modeling actions with explicit manifold assumptions is useful, it is computationally costly. %seen to be computationally expensive.

However, action recognition from videos \cite{twostream,tran2014c3d,karpathy2014large,donahue2014long} does not require elaborate skeletal models. %as videos are more tolerant to body-joint recognition and localization errors. 
A two-stream CNN framework \cite{twostream} uses two streams to model RGB frames and optical flow. 
Tran \etal \cite{tran2014c3d} use CNNs to learn spatio-temporal filters. Karpathy \etal \cite{karpathy2014large} apply RGB and optical-flow fusion, whereas approach \cite{donahue2014long} combines CNNs with LSTM to model temporal flow. \revised{Wang  \etal \cite{temp_seg_net} apply a long-range temporal structure modeling. Tran \etal \cite{spat_convs} study several forms of spatiotemporal convolutions. 
Recent works on fine-grained 
activity recognition use CNNs \cite{cheron2015p,ji20133d} and the human pose estimation for high-level fine-grained reasoning \cite{rohrbach2012database,wang2013approach,zuffi2013puppet,cheron2015p}. Finally, the recent I3D model \cite{i3d_net}  `inflates' 2D CNN filters pretrained on ImageNet to spatio-temporal 3D filters yielding state-of-the-art results.} %, and implements temporal pooling across the inception module.

In contrast to these approaches, %we investigate general higher-order pooling to model spatio-temporal/fine-grained relations of features in videos.
we present a novel  representation of actions based on 3D skeleton sequences and the CNN classifier score sequences. We avoid assumptions about the data manifold by capturing higher-order statistics of the body-joints and the classifier score interactions per  sequence. To this end, our scheme combines positive definite kernels and higher-order tensors, with the goal of obtaining rich and compact representations that  benefit from the non-linearity of radial basis functions (RBF). Such a scheme captures higher-order data statistics \cite{me_tensor}, complex action dynamics \cite{tensor_eccv,hok} and fine-grained relations \cite{lin2017improved,koniusz2018deeper}.

We present two representations for classification of 3D skeletons. Our first representation, \emph{sequence compatibility kernel} (SCK), captures the spatio-temporal compatibility of body-joints between two sequences. To this end, we present an RBF kernel formulation that jointly captures the spatial and temporal similarity of each body-pose (normalized with respect to the hip position) in a sequence against those in another. We show that tensors generated from third-order outer-products of the linearizations of these kernels are a simple yet powerful representation capturing higher-order statistics of body-parts. % to yield high results.

Our second representation, termed~\emph{dynamics compatibility kernel} (DCK), represents spatio-temporal dynamics of each sequence explicitly. %In order to achieve this, 
We present a novel RBF kernel formulation that captures the similarity between a pair of body-poses in a given sequence explicitly, and then compare it against such body-pose pairs in other sequences. Such spatio-temporal modeling could be expensive due to the volumetric nature of space and time. However, we show that using an appropriate kernel model can shrink the time-related variable into a small  representation of constant size after kernel linearization. With this approach, we can model both spatial and temporal variations in the form of co-occurrences which could otherwise be prohibitive. We show empirically that SCK and DCK are complementary.

%We further show through experiments that the above two representations in fact capture complementary statistics regarding the actions, and combining them leads to significant benefits. 

As SCK/DCK work on entire sequences, we formulate an SCK-like kernel over multiple length subsequences as some of subsequences capture the gist of performed actions better than full sequences. To show the versatility of the extended SCK, we apply it to capture spatio-temporal compatibility of frame-wise CNN classifier scores from videos (regular and fine-grained actions).

We present experiments on seven standard datasets, namely (i) UTKinect-Actions~\cite{xia_utkinect}, (ii) Florence3D-Actions~\cite{seidenari_florence3d}, (iii) MSR-Action3D \cite{li_msraction3d} and (iv) HMDB-51\cite{kuehne2011hmdb} datasets as well as two fine-grained datasets (v) NTU RGB+D~\cite{shahroudy2016ntu}, (vi) MPII Cooking Activities~\cite{rohrbach2012database} and (vii) Kinetics \cite{kinetics_400}. We use the first three datasets as a source of 3D body joint sequences (as well as Kinetics), NTU for both 3D body joint sequences, and videos with RGB frames and optical flow frames, and HMDB-51 and MPII Cooking Activities for videos with RGB and optical flow frames. We show that our extensions can still achieve state-of-the-art accuracy two years after SCK/DCK were proposed \cite{tensor_eccv}. 
To summarize: %the main contributions of this paper are:
\renewcommand{\labelenumi}{\roman{enumi}.}
\vspace{-0.3mm}
%\hspace{-1.5cm}
\begin{enumerate}[leftmargin=0.5cm]
\item We design  sequence and  dynamics compatibility kernels that capture spatio-temporal evolution of 3D skeleton body-joints.
\item We derive linearizations of these kernels by tensors.
\item We extend these kernels to aggregation over multiple subsequences and CNN classifier scores.
\item We conduct a novel theoretical analysis of Tensor Power Normalization which connects it to subspace methods.
We are the first to conduct a theoretical analysis of higher-order pooling with Tensor Power Normalization in Section \hyperref[{sec:epn_interp}]{D}, and use it for generic/fine-grained action recognition.
\end{enumerate}

%\revised{The first two contributions are based on our papers \cite{tensor_eccv,hok} while the last two contributions are newly introduced.}

%\vspace{-0.3cm}
\section{Related Work}
\label{sec:related_work}
%This paper is an extension of our early works on higher-order action recognition \cite{tensor_eccv,hok}. In the sequel, we will review some of the more recent related approaches to the problem.% to which our scheme is most related.
 %Below we review some of the more recent related approaches to the problem

In the first part of our paper, we focus on action recognition from an articulated set of connected body-joints that evolve in time~\cite{zatsiorsky_body}. A temporal evolution of the human skeleton is very informative for action recognition as shown by Johansson in his seminal experiment involving the moving lights display~\cite{johansson_lights}. At the simplest level, the human body can be represented as a set of 3D points corresponding to body-joints such as elbow, wrist, knee, ankle, \etc. Action dynamics has been modeled using the motion of such 3D points in~\cite{hussein_action,lv_3daction}, using joint orientations with respect to a reference axis~\cite{parameswaran_viewinvariance} and even relative body-joint positions~\cite{wu_actionlets,yang_eigenjoints}. In contrast, we represent these 3D body-joints by kernels whose linearization results in higher-order tensors capturing complex statistics.  We also note parts-based approaches that use connected body segments~\cite{yacoob_activities,ohn_hog2,ofli_infjoints,vemulapalli_SE3}. For details, see a survey \cite{presti20153d}.
% -- the strength of our representations in contrast to existing works.

We also handle the temporal domain differently to other methods. 3D joint locations are modeled as temporal hierarchy of coefficients in \cite{hussein_action}. Pairwise relative positions of joints were modeled in \cite{wu_actionlets} and combined with a hierarchy of Fourier coefficients to capture temporal evolution of actions. %Moreover, this approach uses multiple kernel learning to select discriminative joint combinations. 
In \cite{yang_eigenjoints}, the relative joint positions and their temporal displacements are modeled with respect to the initial frame. In \cite{vemulapalli_SE3}, the displacements and angles between the body parts are represented as a collection of matrices belonging to SE(3), a special Euclidean group. The temporal domain is handled by the dynamic time warping and Fourier temporal pyramid matching. % on a sequence of such matrices. 
In contrast, we avoid expensive time warping by modeling the temporal domain with an RBF kernel  invariant to local temporal shifts. % and multiple-kernel learning.
 
Our scheme also differs from works such as kernel descriptors \cite{ker_des} that %aggregate orientations of gradients for recognition. Their approach exploits sums over the product of at most two RBF kernels handling two cues \eg, 
sum gradient orientations over image patches, %and apply Kernel PCA and Nystr\"{o}m techniques. %Similarly, convolutional kernel networks \cite{ckn} consider stacked layers of a variant of kernel descriptors \cite{ker_des}.
action recognition via kernelized covariances % which are obtained in Nystr\"{o}m-like process.
%\cite{cavazza_kercov,lei_kermats}.
\cite{lei_kermats,cavazza_kercov,lei_sice}, and
a time series kernel \cite{gaidon_timekern} %between auto-correlation matrices 
which extracts spatio-temporal autocorrelations.
%\todo{unclear! Split the following sentences and rephrase!} %\textcolor{blue}{In contrast, our scheme allows sum kernels over multiple products and sums of RBF kernels which in turn lead to higher-order statistics captured in fourth-order tensors. This is achieved by linearizing and combining obtained feature maps with a polynomial kernel of order $r\!\geq2$ according to the derivations provided in the text. }
In contrast, our scheme sums over several multiplicative and additive RBF kernels. We capture higher-order statistics by linearizing a polynomial kernel and avoid evaluating costly kernels directly. % in contrast to kernel trick.

Third-order tensors have been used to form spatio-temporal tensors on videos in \cite{tensoraction2007}. Non-negative tensor factorization is used for image denoising \cite{shashua2005non}, tensors are used for texture rendering \cite{vasilescu2004tensortextures} and for face recognition \cite{vasilescu2002multilinear}. A survey of multi-linear algebraic methods for tensor subspace learning is available in~\cite{lu2011survey}. These methods use a single tensor, whereas we use tensors as descriptors \cite{me_tensor2,me_tensor,sparse_tensor_cvpr,zhao2012comprehensive}. However, we use  third-order tensors for action recognition, which poses a  set of new challenges. %In this spirit, our kernel formulation is completely different from these prior works.

For fine-grained action recognition, high-level sophisticated action reasoning~\cite{rohrbach2012database,wang2013approach,zuffi2013puppet,cheron2015p} is typically used together with  pose estimation systems \cite{wei2016cpm,Insafutdinov2016}. However, these approaches scale poorly to millions of video frames. Human-object interactions in the videos are analyzed in \cite{zhou2015interaction}. Correlations between space-time features are proposed in~\cite{shechtman2005space}.

Power Normalization approaches \cite{me_tensor2,me_tensor,sparse_tensor_cvpr,koniusz2018deeper,deeper_look2} speculate that Eigenvalue Power Normalization  prevents so-called burstiness, thus performing spectral detection of higher-order occurrences of features \cite{me_tensor2,me_tensor}, which can be paraphrased as `\emph{do a knife, a hand and a chopping board co-occur together?}' rather than `\emph{how many knifes, hands and chopping boards appear in the scene?}'

Moreover, first-order pooling was successfully used for representing action recognition via hallucination \cite{i3d_halluc}. Papers \cite{koniusz2018deeper,deeper_look2} study second-order pooling, power normalizing functions and their taxonomy while fast pooling methods are proposed in \cite{lin2017improved,deeper_look2,pk_maji}. 

Finally, second-order pooling was successfully used for few-shot action recognition \cite{fsl_action}, few-shot classification \cite{sosn, christian_subs}, few-shot segmentation \cite{Zhang_2020_ACCV}, modulating  optimization \cite{christian_modulate}, style transfer \cite{fatima_dicta,fatima_wacv18,fatima_wacv19,fatima_ijcv} and action self-supervision \cite{i3d_halluc2}. Noteworthy are also graph convolutional networks \cite{stgcn,uai_ke,zhu2021simple} and embeddings \cite{refine} easily applicable to 3D skeleton action recognition.

%In contrast, we are the first to conduct a theoretical analysis of higher-order pooling with Tensor Power Normalizations (Section \hyperref[{sec:epn_interp}]{D}) and use it for generic/fine-grained action recognition.

\section{Preliminaries}
\label{sec:framework}
%In this section, we review our notations and the necessary background on shift-invariant kernels and their linearizations, which will be useful for deriving kernels. % on 3D skeletons for action recognition.
In this section, we review our notations and the necessary background on shift-invariant kernels and their linearizations.

\subsection{Tensor Notations}
\label{sec:not} 
\begin{figure}[t]%htbp % left bottom right top
\centering%%%%\vspace{-0.3cm}
\hspace{-0.5cm}
\begin{subfigure}[b]{0.495\linewidth}
\centering\includegraphics[trim=0 0 0 0, clip=true, height=2.0cm]{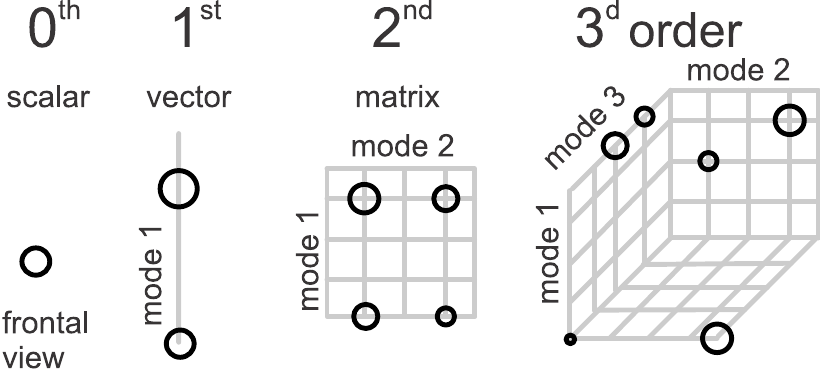}%2.6
%\phantomcaption
\caption{\label{fig:ten0}}
\end{subfigure}
\begin{subfigure}[b]{0.495\linewidth}
\centering\includegraphics[trim=0 0 0 0, clip=true, height=2.0cm]{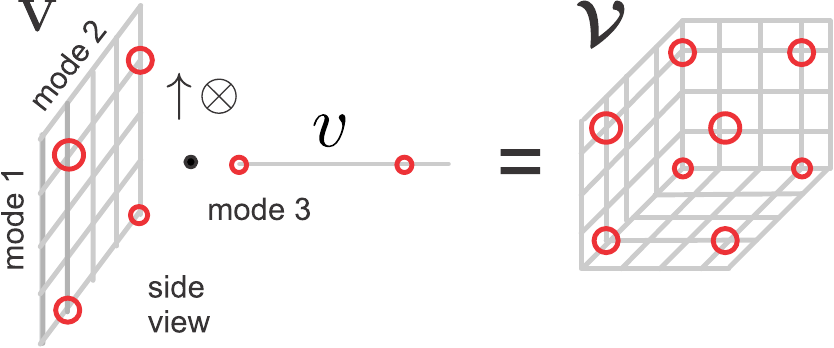}%1.42
%\phantomcaption
\caption{\label{fig:ten1}}
\end{subfigure}
%
%%%%\vspace{-0.2cm}
\caption{\revised{Figure \ref{fig:ten0} illustrates the notion of tensors, their order and modes. Figure \ref{fig:ten1} illustrates the matrix-vector order outer-product.
}}\vspace{-0.3cm}
\end{figure}

Figure \ref{fig:ten0} illustrates the notion of tensors, their order and modes. Let $\tV\in\mbr{d_1\times d_2\times d_3}$ denote a third-order tensor. Using the Matlab notation, we refer to the $k$-th slice of this tensor as $\tV_{:,:,k}$, which is a $d_1\times d_2$ matrix. For a matrix $\mV\in\mbr{d_1\times d_2}$ and a vector $\vv\in\mbr{d_3}$, the notation  $\tV\!=\!\mV\kronstack\vv$ produces a tensor $\tV\!\in\!\mbr{d_1\times d_2\times d_3}$ \revised{whose $k$-th slice is given by $\mV\!\cdot\!v_k$, $v_k$ being the $k$-th coefficient of $\vv$. Figure \ref{fig:ten1} illustrates such an outer-product. Symmetric third-order tensors of rank one are formed by the outer-product of a vector $\vv\in\mbr{d}$ in three modes, that is, a rank-one $\tV\in\mbr{d\times d\times d}$ is obtained from $\vv$ as $\tV\!=\!({\kronstack}_3\vv\!\triangleq\!(\vv\vv^T)\kronstack\vv)$ which yields $\tVnb_{ijk}\!=\!v_i\cdot\!v_j\!\cdot\!v_k$, where $\tVnb_{ijk}$ represents the $ijk$-th element of $\tV$. Matrices have two modes: the first and second mode correspond to the row and column indexes $i$ and $j$, respectively. Order $r$ tensors have $r$ modes addressed by $\tVnb_{i_1\cdots i_r}$ where $\tV\!\in\!\mbr{d_1\!\times\!\cdots\!\times\!d_k\!\times\!\cdots\!\times\!d_r}$ and $k$ indicates the mode $k$. Concatenation of $n$ tensors in mode $k$ is simply stacking them along mode $k$, denoted as $\left[\tV_i\right]_{i\in\idx{n}}^{\oplus_k}\!\equiv\!\text{numpy.concatenate}((\tV_1,\cdots,\tV_n), \text{axis}\!=\!k\!-\!1))$. $\idx{n}$ is an index sequence $1,2,\cdots, n$}. We define the Frobenius norm $\fnorm{\tV} = \sqrt{\sum_{i,j,k} \tVnb_{ijk}^2}$ and the inner-product between $\tX$ and $\tY$ as $\left\langle\tX,\tY\right\rangle=\sum_{ijk}\tXnb_{ijk}\tYnb_{ijk}$. Also, $\vec{e}_z$ are spanning bases of $\mbr{Z}$. %$\vOnes$ is a vector with coeff. equal 1.
%of the sum of element-wise squares: $||\tV||_F\!=\!\Big({\sum\limits_{i,j,k}\tVnb_{ijk}^2}\Big)^{0.5}\!\!$ and an associated with it dot-product  $\left<\tX,\tY\right>\!=\!\sum\limits_{i,j,k}\!\tXnb_{ijk}\tYnb_{ijk}$.
\revised{Further basics on tensors and tensor algebra can be found in \cite{ten_basics}.}

\subsection{Kernel Linearization}
\label{sec:kernel_linearization}
Let $G_{\sigma}(\vu-\vub)=\exp(-\enorm{\vu - \vub}^2/{2\sigma^2})$ denote a standard Gaussian RBF kernel centered at $\vub$ and having a bandwidth $\sigma$. Kernel linearization refers to rewriting this $G_{\sigma}$ as an inner-product of two infinite-dimensional feature maps. To obtain these maps, we use a fast approximation method based on probability product kernels \cite{jebara_prodkers}. Specifically, we employ the inner product of $d'$-dimensional isotropic Gaussians given $u,u'\!\!\in\!\mbr{d'}\!$. Thus, we have: %The resulting approximation can be written as:
% where $\vu\in\mbr{d}$ and $\vv\in\mbr{d}$:
%which underpins the subsequent formulations and we employ a linearization of Gaussian kernels $G_{\sigma}$ to obtain feature maps that express these kernels by the dot-product. 
%
\begin{align}
&G_{\sigma}\!\left(\vu\!-\!\vub\right)\!\!=\!\!\left(\frac{2}{\pi\sigma^2}\right)^{\!\!\frac{d'}{2}}\!\!\!\!\!\!\int\limits_{\vzeta\in\mbr{d'}}\!\!\!\!G_{\sigma/\sqrt{2}}\!\!\left(\vu\!-\!\vzeta\right)G_{\sigma/\sqrt{2}}(\vub\!\!-\!\vzeta)\,\mathrm{d}\vzeta.
\label{eq:gauss_lin}
\end{align}%\\[-15pt]
Eq. \eqref{eq:gauss_lin} is then approximated by replacing the integral with the sum over $Z$ pivots $\vzeta_1,\cdots,\vzeta_Z$. Thus, we obtain a feature map $\vphi$:
\begin{align}
&\vphi(\vu; \{\vzeta_i\}_{i\in\idx{Z}})=\left[{G}_{\sigma/\sqrt{2}}(\vu-\vzeta_1),\cdots,{G}_{\sigma/\sqrt{2}}(\vu-\vzeta_Z)\right]^T\!\!\!,\!\!\label{eq:gauss_lin2a}\\
\text{ and } & G_{\sigma}(\vu\!-\!\vub)\approx\left<\sqrt{c}\vphi(\vu), \sqrt{c}\vphi(\vub)\right>,
\label{eq:gauss_lin2}
\end{align}
where $c$ is a const. \revised{Eq. \eqref{eq:gauss_lin2} is the linearization of the RBF kernel. Eq. \eqref{eq:gauss_lin2a} is the feature map. $\{\vzeta_i\}_{i\in\idx{Z}}$ are  pivots. As we use 1 dim. signals, we simply cover interval $[-1;1]$ (or $[0;1]$) with $Z$ equally spaced pivots. For clarity, we drop $\{\vzeta_i\}_{i\in\idx{Z}}$ and write $\vphi(\vu)$, \etc.
}

\revised{
\subsection{Equivalence between Polynomial Kernels and the Dot-product of Tensors \cite{me_tensor}}
\label{sec:kernel_linearization_eq}
For any two $Z'\!$ dim. feature vectors $\vphi,\bar{\vphi}\!\in\!\mbr{Z'}$, we have:
\begin{align}
&\!\!\!\!\!\!\!\!\!\!\left<\vphi,\bar{\vphi}\right>^r\!\!\!=\!\!%\left(\sum_{i=1}^{Z'} v_{i} \bar{v}_{i} \right)^r{\!\!\!\!=\!\!}
{\sum_{i_1=1}^{Z'}}{\!\cdots\!}{\sum_{i_r=1}^{Z'}} \phi_{i_1}{\bar{\phi}_{i_1}}\!\cdot\!{\cdots}\!\cdot\!\phi_{i_r} \bar{\phi}_{i_r}%\nonumber\\
%& \qquad{}\qquad{}\qquad{}\;\;\;
%&\qquad\,= 
\!\!=\!\left<{\kronstack}_r \vphi, {\kronstack}_r \bar{\vphi} \right>\!,%\nonumber\\[-10pt]
&
\label{eq:polexpand}
\end{align}
where $\tX\!=\!\left({\kronstack}_r\vphi\right)$ is defined as $\mathcal{X}_{i_1\cdots i_r}\!=\!\phi_{i_1}\!\cdot\!\cdots\!\cdot\!\phi_{i_r}$. %A similar expansion and its details can be found in \cite{me_tensor}.
}

\section{Proposed Approach}
Below, we formulate the problem of action recognition from 3D skeleton sequences, which precedes an exposition of our two kernel formulations for describing actions, followed by their tensor reformulations through kernel linearization. We also introduce Eigenvalue Power Normalization and our improved kernels used for action recognition based on skeletons and/or classifier scores obtained from videos passed via CNNs.

\revised{
\subsection{Statistical Motivation}
Before we outline our higher-order tensor representations, below we motivate the use of higher-order statistics. To compare skeleton sequences/videos, we want to capture distribution of local features/descriptors per sequence \eg, body joints or receptive fields in CNN. The characteristic function $\varphi_\Phi(\boldsymbol{\omega})\!=\!\mathbb{E}_{\vphi\sim\Phi}\left(\exp(\iu\boldsymbol{\omega}^T\!\vphi)\right)$ describes the probability density $f_\Phi(\vphi)$ of a skeleton sequence/video  (local features/descriptors $\vphi\!\sim\!\Phi$). 

%\vspace{-0.05cm}
\begin{tcolorbox}[width=1.0\linewidth, colframe=blackish, colback=beaublue, boxsep=0mm, arc=3mm, left=1mm, right=1mm, right=1mm, top=1mm, bottom=1mm]
Taylor expansion of the characteristic function per sequence is:
\vspace{-0.1cm}
\begin{align}
&\!\!\!\!\mathbb{E}_{\vphi\sim\Phi}\Big(\sum\limits_{r=0}^\infty\frac{\iu^r}{r!}\left<\vphi,\boldsymbol{\omega}\right>^r\Big)\!\approx\!\!\frac{1}{N}\sum\limits_{n=0}^N\sum\limits_{r=0}^\infty\frac{\iu^r}{r!}\left<{\kronstack}_r\vphi_n,{\kronstack}_r\boldsymbol{\omega}\right>\!\!\!\\
%\end{equation}
%
%\begin{equation}
&\!\!\!\!%\frac{1}{N}\sum\limits_{n=0}^N
=\sum\limits_{r=0}^\infty\frac{\iu^r}{r!}\Big<\frac{1}{N}\sum\limits_{n=0}^N{\kronstack}_r\vphi_n,{\kronstack}_r\boldsymbol{\omega}\Big>\!=\!\sum\limits_{r=0}^\infty\Big<\tX^{(r)},\frac{\iu^r}{r!}{\kronstack}_r\boldsymbol{\omega}\Big>.\nonumber
\end{align}

\vspace{-0.2cm}
\noindent{Symbol} $\tX^{(r)}\!=\!\frac{1}{N}\sum\limits_{n=0}^N{\kronstack}_r\vphi_n$ defines a tensor descriptor while $\iu$ is the imaginary number. In principle, with infinite data and infinite moments, one can fully capture  $f_\Phi(\vphi)$ which is intractable. In practice, third-order moments work well in what follows while second-order moments are somewhat insufficient.
\end{tcolorbox}
\vspace{-0.15cm}
}

\subsection{Problem Formulation}
Suppose we are given a set of 3D human pose skeleton sequences, each pose consisting of $J$ body-keypoints. Further, to simplify our notations, we assume each sequence consists of $N$ skeletons, one per frame\footnote{\label{foot:foo0}We assume that all sequences have $N$ frames for simplification of presentation. Our formulations are applicable to sequences of arbitrary lengths \eg,~$M$ and $N$. Thus, we apply in practice $G_{\sigma_3}(\frac{s}{M}-\frac{t}{N})$ in Eq. \eqref{eq:ker1a}.}. We  define such a pose sequence $\Pi$ as:
\begin{equation}
\Pi = \set{\vx_{is}\in\mbr{3},i\in\idx{J}, s\in\idx{N}}.
\end{equation}
Further, let each such a sequence $\Pi$ be associated with one of $K$ action class labels $\ell\in\idx{K}$. Our goal is to use the skeleton sequence $\Pi$ and generate an action descriptor for this sequence that can be used in a classifier for recognizing the action class. In what follows, we will present two such action descriptors, namely (i) sequence compatibility kernel and (ii) dynamics compatibility kernel, which are formulated using kernel linearization and tensor algebra theories. We present both these kernel formulations next.

\begin{figure*}[t]%htbp % left bottom right top
\centering%%%%\vspace{-0.3cm}
%\hspace{-1.5cm}
\begin{subfigure}[b]{0.210\linewidth}
\centering\includegraphics[trim=0 0 0 0, clip=true, width=3.64cm]{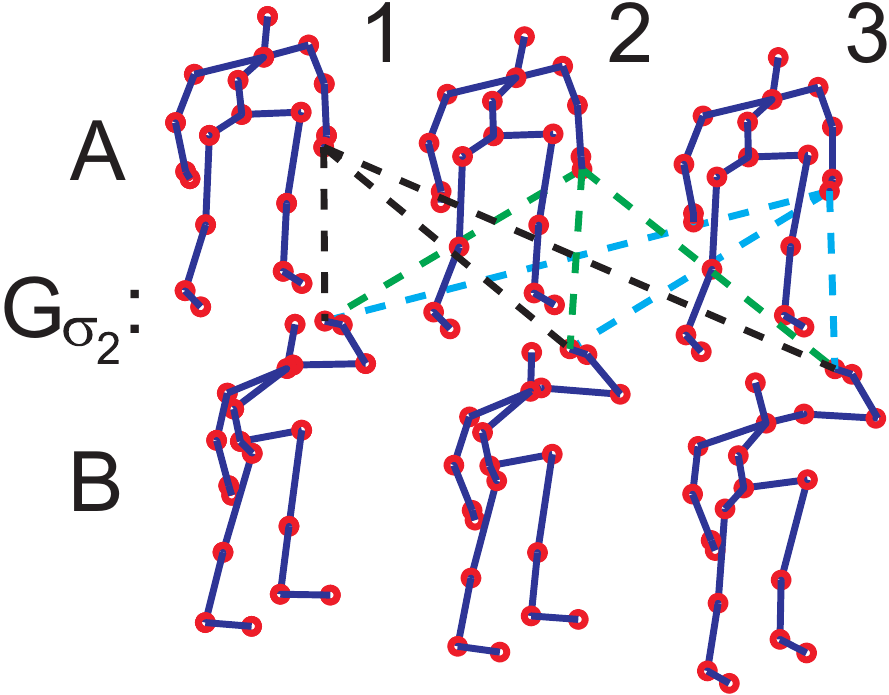}%2.6
%\phantomcaption
\vspace{-0.3cm}
\caption{\label{fig:ker0a}}
\end{subfigure}
\begin{subfigure}[b]{0.12\linewidth}
\centering\includegraphics[trim=0 0 0 0, clip=true, width=1.988cm]{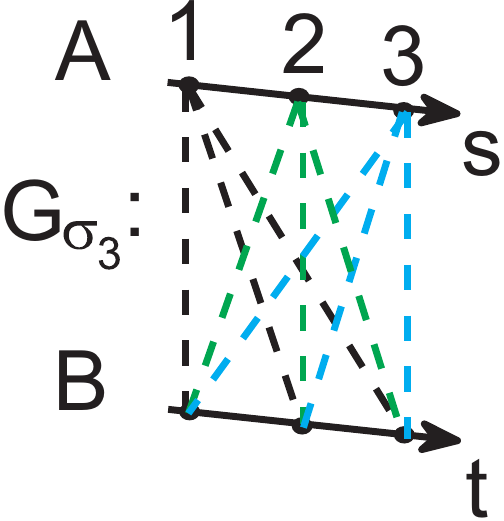}%1.42
%\phantomcaption
\vspace{-0.3cm}
\caption{\label{fig:ker0a2}}
\end{subfigure}
\begin{subfigure}[b]{0.65\linewidth}
%\centering\includegraphics[trim=15 15 0 15, clip=true, width=4.7cm]{images/conc2d.pdf}
%\centering\includegraphics[trim=0 0 0 0, clip=true, width=11.2cm]{images/conc1dd.pdf}%8.0
\centering\includegraphics[trim=0 0 0 0, clip=true, width=11.2cm]{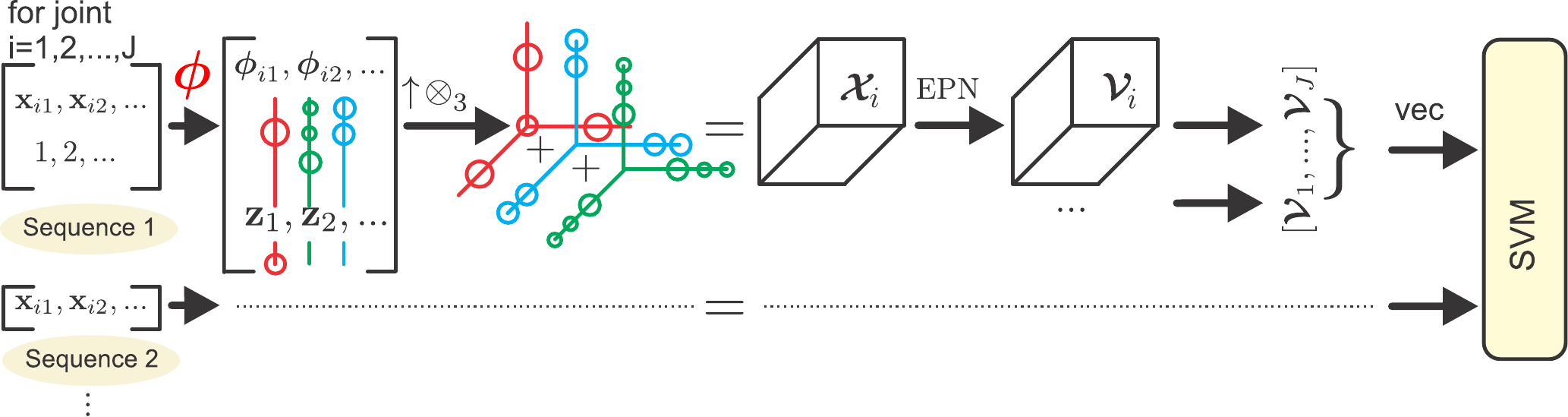}%8.0
%\phantomcaption
\vspace{-0.3cm}
\caption{\label{fig:ker0b}}
\end{subfigure}
%%%%\vspace{-0.2cm}
\caption{Figures \ref{fig:ker0a} and \ref{fig:ker0a2} show how SCK works -- kernel $G_{\sigma_2}$ compares exhaustively \eg~hand-related joint $i$ for every frame in sequence $A$ with every frame in sequence $B$. Kernel $G_{\sigma_3}$ compares exhaustively the frame indexes. \revised{Figure \ref{fig:ker0b}} shows this burden is avoided by linearization -- third-order statistics on feature maps $\vphi(\vx_{is})$ and $\vz(s/N)$ for joint $i$ are captured in tensor $\tX_i$ and whitened by EPN to obtain $\tV_i$ which are concatenated over $i\!=\!1,\cdots,J$ to represent a sequence. \revised{The final sequence tensors are vectorized per video by `vec' and fed to an SVM}.
%are now compared only between the corresponding frames by the dot-product.
}\vspace{-0.3cm}
\end{figure*}

\begin{figure}[t]%htbp % left bottom right top
\centering%%%%\vspace{-0.3cm}
%\hspace{-1.5cm}
%\begin{subfigure}[b]{0.210\linewidth}
%\centering\includegraphics[trim=0 0 0 0, clip=true, width=3.64cm]{images/conc1da.pdf}%2.6
%%\phantomcaption
%\caption{\label{fig:ker0a}}
%\end{subfigure}
%
%%%%\vspace{-0.2cm}
\includegraphics[trim=0 0 0 0, clip=true, width=9cm]{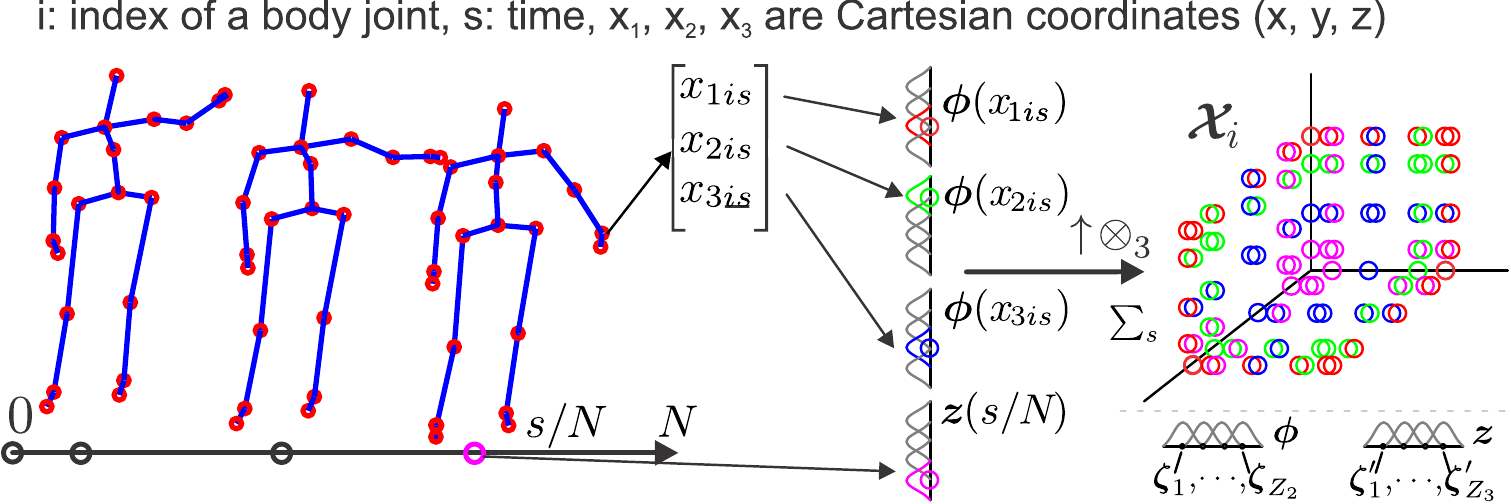}%2.6
\caption{\revised{Order $r$ statistics from Eq. \eqref{eq:ker1a} can be understood by studying the linearization in Eq. \eqref{eq:ker1b}. For a given joint $i$ at time $s/N$ (normalized frame number), we embed a 3D joint coordinate $\vx_{is}$ (all centered \wrt hip) via function $\vphi(\cdot)$ into a non-linear Hilbert space representing an RBF kernel according to Eq. \eqref{eq:gauss_lin2a}. Similarly, we embed the time $s/N$ via function $\vz(\cdot)$ (also by Eq. \eqref{eq:gauss_lin2a}). Finally, $\otimes_{r}$ performs the third-order outer-product on concatenated embeddings aggregated next over frames $s$ (note $\sum_s$). The interpretation: the Gaussians `soft-divide' the the Cartesian coordinate system along $x$, $y$, $z$ direction, resp., and time $s/N$. Thus, triplets $(x,y,z)$, $(x,y,s/N)$, $(x,z,s/N)$ and $(y,z,s/N)$ assigned into such a `soft-divided' space capture locally three-way occurrences. They factor out one spatial (or time) variable at a time (note invariance to such a variable).}}
\label{fig:phis1}
\vspace{-0.4cm}
\end{figure}

\subsection{Sequence Compatibility Kernel}
\label{sec:ker1}
As alluded to earlier, the main idea of this kernel is to measure the compatibility between two action sequences in terms of the similarity between their skeletons and their temporal order. To this end, we assume each skeleton is centered with respect to one of the body-joints (say, hip). Suppose we are given two such sequences $\Pi_A$ and $\Pi_B$, each with $J$ joints, and $N$ frames. Further, let $\vx_{is}\!\in\!\mbr{3}$ and $\vy_{jt}\!\in\!\mbr{3}$ correspond to the  body-joint coordinates of $\Pi_A$ and $\Pi_B$, respectively. 

\begin{tcolorbox}[width=1.0\linewidth, colframe=blackish, colback=beaublue, boxsep=0mm, arc=3mm, left=1mm, right=1mm, right=1mm, top=1mm, bottom=1mm]
We define our~\emph{sequence compatibility kernel} (SCK) between $\Pi_A$ and $\Pi_B$ as\footref{foot:foo0}:
\vspace{-0.5cm}
\begin{align}
& \qquad\qquad\qquad\qquad K_S(\Pi_A,\Pi_B) =\label{eq:ker1a}\\
&\!\!\frac{1}{\Lambda}\!\!\! \sum\limits_{(i,s)\in\idxJ}\!\!\!\sum\limits_{\;\;(j,t)\in\idxJ\!\!}\!\!\!\!G_{\sigma_1}(i\!-\!j)\Big(\beta_1 G_{\sigma_2}\!\left(\vx_{is} \!-\! \vy_{jt}\right) \!+\! \beta_2\, G_{\sigma_3}({\displaystyle\frac{s-t}{N}})\Big)^r\!\!.\nonumber
\end{align}
\end{tcolorbox}
\vspace{-0.2cm}

\noindent{Symbol} $\Lambda$ is a normalization constant and $\idxJ=\idx{J}\times\idx{N}$. As is clear, this kernel involves three different compatibility subkernels, namely (i) $G_{\sigma_1}$, capturing the compatibility between joint-types $i$ and $j$, (ii) $G_{\sigma_2}$, capturing the compatibility between joint locations $\vx$ and $\vy$, and (iii) $G_{\sigma_3}$, measuring the temporal alignment of two poses in two sequences. We also introduce weighting factors $\beta_1,\beta_2\geq 0$ that adjust the importance of the body-joint compatibility against the temporal alignment, where $\beta_1+\beta_2=1$. Figures \ref{fig:ker0a} and \ref{fig:ker0a2} illustrate how this kernel works. It might come as a surprise that we use kernel $G_{\sigma_1}$. Note that our skeletons may be noisy and there is a possibility that some  keypoints are detected incorrectly (for example, elbows and wrists). Thus, this kernel allows incorporating a degree of uncertainty into the alignment of such joints. To simplify our formulation, in this paper, we will assume that such errors are absent from our skeletons, and thus $G_{\sigma_1}(i-j)=\delta(i-j)$. Furthermore,  standard deviations $\sigma_2$ and $\sigma_3$ control the joint-coordinate selectivity and temporal shift-invariance, respectively. That is, for $\sigma_3\rightarrow 0$, two sequences will have to match perfectly in the temporal sense. For $\sigma_3\rightarrow\infty$, the algorithm is invariant to any permutations of the frames. As will be clear in the sequel, parameter $r$ determines the order of statistics of our kernel (we use $r=3$).

Next, we present linearization of our kernel using the method %proposed in
from Sections~\ref{sec:kernel_linearization}, \ref{sec:kernel_linearization_eq}, and Eq.~\eqref{eq:gauss_lin2}, 
so that kernel $G_{\sigma_2}(\vx-\vy)\approx \vphi(\vx)^T\vphi(\vy)$ (see note\footnote{\label{foot:foob}\revised{In practice, Cartesian coordinates of joints $\vx,\vy\!\in\!\mbr{3}$ are fed into a kernel. Thus, in place of kernel $G_{\sigma_2}$, we use the sum kernel $G^{'}_{\sigma_2}(\vx\!-\!\vy)\!=\!G_{\sigma_2}(x_1\!\!-\!y_1)\!+\!G_{\sigma_2}(x_2\!\!-\!y_2)\!+\!G_{\sigma_2}(x_3\!\!-\!y_3)$ whose approximation is given as:
$G^{'}_{\sigma_2}(\vx\!-\!\vy)\approx[\vphi(x_1;\{\vzeta_i\}_{i\in\idx{Z_2}}\!); \vphi(x_2; \{\vzeta_i\}_{i\in\idx{Z_2}}); \vphi(x_3; \{\vzeta_i\}_{i\in\idx{Z_2}})]^T[\vphi(y_1;$ $\{\vzeta_i\}_{i\in\idx{Z_2}}); \vphi(y_2; \{\vzeta_i\}_{i\in\idx{Z_2}}); \vphi(y_3; \{\vzeta_i\}_{i\in\idx{Z_2}})]$
but for simplicity we refer to it in our formulations by its generic form $G_{\sigma_2}(\vx\!-\!\vy)\!\approx\!\vphi(\vx)^T\vphi(\vy)$ 
because we can define $\phi(\vx)\!=\![\phi(x_1\!); \phi(x_2\!); \phi(x_3\!)]$.
 %Note that $(x)$, $(y)$, $(z)$ are the spatial xyz-components of joints.
}}) while $G_{\sigma_3}(\frac{s-t}{N})\approx\vz(s/N)^T\vz(t/N)$ \revised{(see note\footnote{\label{foot:temp}\revised{Feature maps $\vz(\cdot)\!\equiv\!\vphi(\cdot)$ from Eq. \eqref{eq:gauss_lin2a}. We simply write $\vz$ rather than $\vphi$ to denote these feat. maps as they encode the time/frame number (\cf the body joints). Note that $\vz(\cdot; \{\vzeta'_i\}_{i\in\idx{Z_3}})$ uses $Z_3$ pivots $\{\vzeta'_i\}_{i\in\idx{Z_3}}$ (see Figure \ref{fig:phis1}).}})}. With these approximations and simplification to $G_{\sigma_1}\!$ described above, we rewrite our sequence compatibility kernel as:
\begin{align}
&K_S(\Pi_A,\Pi_B) \approx\nonumber\\
&\!\frac{1}{\Lambda}\!\!\sum\limits_{i\in\idx{J}}\sum\limits_{s\in\idx{N}}\!\sum\limits_{t\in\idx{N}}\!\!\!\left(
\begin{bmatrix}
\sqrt{\beta_1}\,\vphi(\vx_{is}),\text{ (see note\footref{foot:foob})}\\
\sqrt{\beta_2}\,\vz(s/N),\text{ (see note\footref{foot:temp})}\\[3pt]
\end{bmatrix}^T\!\!\!\cdot
\begin{bmatrix}
\sqrt{\beta_1}\vphi(\vy_{it})\\
\sqrt{\beta_2}\vz(t/N)\\[3pt]
\end{bmatrix}\right)^r\label{eq:ker1exp1}\\
&=\!\frac{1}{\Lambda}\!\!\sum\limits_{i\in\idx{J}}\sum\limits_{s\in\idx{N}}\!\sum\limits_{t\in\idx{N}}\!\!\!\left<
{\kronstack}_r\!\begin{bmatrix}
\sqrt{\beta_1}\,\vphi(\vx_{is})\\
\sqrt{\beta_2}\,\vz(s/N)\\[3pt]
\end{bmatrix}\!,
{\kronstack}_r\!\begin{bmatrix}
\sqrt{\beta_1}\vphi(\vy_{it})\\
\sqrt{\beta_2}\vz(t/N)\\[3pt]
\end{bmatrix}\right>\label{eq:ker1exp2}\\
&=\!\!\!\sum\limits_{i\in\idx{J}}\!\!\left<
\!\frac{1}{\sqrt{\Lambda}}\!\!\sum\limits_{s\in\idx{N}}\!\!{\kronstack}_r\!\begin{bmatrix}
\sqrt{\beta_1}\,\vphi(\vx_{is})\\
\sqrt{\beta_2}\vz(s/N)\\[3pt]
\end{bmatrix}\!,
\frac{1}{\sqrt{\Lambda}}\!\!\sum\limits_{t\in\idx{N}}\!\!{\kronstack}_r\!\begin{bmatrix}
\sqrt{\beta_1}\vphi(\vy_{it})\\
\sqrt{\beta_2}\vz(t/N)\\[3pt]
\end{bmatrix}\right>.
\label{eq:ker1b}
\end{align}

\revised{Expansion of Eq. \eqref{eq:ker1exp1} into Eq. \eqref{eq:ker1exp2} simply follows the notion of equivalence between the polynomial kernels and tensor outer-products as detailed in Eq. \eqref{eq:polexpand}. Similarly, the summations in Eq. \eqref{eq:ker1exp2} can be absorbed into the dot-product in Eq. \eqref{eq:ker1b} because the inner-product is a linear operation in each of its arguments \eg, $\left<\vv_1\!+\!\vv_2,\bar{\vv}\right>\!=\!\left<\vv_1,\bar{\vv}\right>\!+\!\left<\vv_2,\bar{\vv}\right>$.}
\revised{The physical  meaning of the above equation is detailed in Figure \ref{fig:phis1}. While the first-, second- and third-order outer-products are connected to the sample mean, covariance and co-skewness of features, our tensors are not mere counts of features, as explained next. }
As is clear,~\eqref{eq:ker1b} expresses $K_S(\Pi_A,\Pi_B)$ as a sum of inner-products on third-order tensors ($r=3$), as shown in Figure \ref{fig:ker0b}. While, using the dot-product as the inner-product is an option, other alternatives for tensors of order $r\geq 2$ can act on their spectrum, leading to better representations. An example is the so-called \emph{burstiness}~\cite{jegou_bursts}, which is a commonly encountered property that a given feature appears more/less often in a sequence than a statistically independent model predicts. Robust descriptors must be invariant \wrt the length of actions \eg, a prolonged \emph{hand waving} represents the same action as a short \emph{hand wave}. % (or short vs. long \emph{head nodding}). 
Eigenvalue Power Normalization (EPN)~\cite{me_tensor}  suppresses burstiness by acting on higher-order statistics (see Fig.~\ref{fig:ker0b}). By incorporating EPN, we generalize~\eqref{eq:ker1b} as:
\vspace{-0.4cm}
\begin{align}
&
\!K_S^{*}(\piA,\piB)\!=\!\!\!\sum\limits_{i\in\idx{J}}\!\!\left<
\!\mygthree{\frac{1}{\sqrt{\Lambda}}\!\!\sum\limits_{s\in\idx{N}}\!\!\!{\kronstack}_r\!\!\begin{bmatrix}
\!\sqrt{\beta_1}\vphi(\vx_{is})\\
\sqrt{\beta_2}\vz(s/N)\\[3pt]
\end{bmatrix}}\!\!,\right.\nonumber\\
&\qquad\qquad\qquad\qquad\;\,\left.\mygthree{\frac{1}{\sqrt{\Lambda}}\!\!\sum\limits_{t\in\idx{N}}\!\!\!{\kronstack}_r\!\!\begin{bmatrix}
\sqrt{\beta_1}\vphi(\vy_{it})\\
\sqrt{\beta_2}\vz(t/N)\\[3pt]
\end{bmatrix}}\!\!\right>\!,\label{eq:ker1c}
\end{align}
where the operator $\tG$ performs EPN by applying power normalization to the spectrum of the third-order tensor (by taking the higher-order SVD). Note that in general $K_S^{*}(\piA,\piB)\!\not\approx\!K_S(\piA,\piB)$ as $\tG$ is intended to manipulate the spectrum of $\tX$. 

\begin{tcolorbox}[width=1.0\linewidth, colframe=blackish, colback=beaublue, boxsep=0mm, arc=3mm, left=1mm, right=1mm, right=1mm, top=1mm, bottom=1mm]
The final representation for linearized SCK becomes:
\vspace{-0.1cm}
\begin{align}
& \tV_i\!=\!\mygthree{\tX_i}\!,\text{ where } \tX_i\!=\!\!\frac{1}{\sqrt{\Lambda}}\!\!\!\sum\limits_{s\in\idx{N}}\!\!\!{\kronstack}_r\!\!\begin{bmatrix}
\!\sqrt{\beta_1}\,\vphi(\vx_{is})\\
\sqrt{\beta_2}\vz(s/N)\\[3pt]
\end{bmatrix}\!.\label{eq:ker1d}
\end{align}
\noindent{We} replace the sum over the body-joint indexes in \eqref{eq:ker1c} by concatenating $\tV_i$ in ~\eqref{eq:ker1d} along the fourth tensor mode, thus defining $\tV = \big[\tV_i\big]_{i\in\idx{J}}^{\oplus_4}$. Suppose $\tV_A$ and $\tV_B$ are the corresponding fourth order tensors for $\Pi_A$ and $\Pi_B$ respectively. 
Then, we obtain:
\vspace{-0.3cm}
\begin{align}
&\qquad K_S^{*}(\piA,\piB)=\left<\tV_A, \tV_B\right>.
%\left<\big[\tV_i\big]_{i\in\idx{J}}^{\oplus_4}\!,\big[\tVH_i\big]_{i\in\idx{J}}^{\oplus_4}\right>.\label{eq:ker1e}
\end{align}
\end{tcolorbox}
\vspace{-0.1cm}

Note that tensors $\tX$ have the following properties: (i) super-symmetry $\tX_{i,j,k}\!=\!\tX_{\pi(i,j,k)}$ for indexes $i,j,k$ and their permutation given by $\pi,\;\forall\pi$, and (ii) positive semi-definiteness of every slice, that is, $\tX_{:,:,s}\!\in\!\semipd{d},$ for $s\!\in\!\idx{d}$. Thus, we use only the upper-simplices of $\tV_i$ which consist of $\binom{d+r-1}{r}$ coefficients (which is the total size of our final representation times the number of body-joints) rather than $d^r\!$, where $d$ is the side-dimension of $\tV_i$ \ie, $d\!=\!3Z_2\!+\!Z_3$ (see notes\footref{foot:foob}\textsuperscript{,}\footref{foot:temp}), and $Z_2$ and $Z_3$ are the numbers of pivots used in the approximation of $G_{\sigma_2}$ and $G_{\sigma_3}$ (see notes\footref{foot:foob}\textsuperscript{,}\footref{foot:temp}). % respectively. 

%As we want to preserve the above listed properties in tensors $\tV$, we employ slice-wise EPN which is induced by the Power-Euclidean distance and involves rising matrices to a power $\gamma$. Finally, we re-stack these slices along the third mode as:

Next, we pass tensors $\tX$ via (i) slice-wise EPN (sEPN) operator or (ii) HOSVD-based tensor whitening EPN (tEPN) \cite{me_tensor}. sEPN is faster but tEPN uses the entire tensor spectrum, thus being more accurate. 
The slice-wise EPN uses the Power-Euclidean dist. for rising matrices, slices of tensor tensor $\tX$, to the power of $\gamma$. Power norm. and re-stacking slices along the third mode yields:
\begin{align}
& \mygthree{\tX}\!=\![\tX_{:,:,s}^{\gamma}]_{s\in\idx{d}}^{\oplus_3}, \text{ for } 0\!<\gamma\!\leq\!1.\label{eq:epn1}
\end{align}
We note that $\mygthree{\tX}$ preserves listed earlier properties of tensors $\tX$ and it forms our final tensors $\tV$ for the action sequence.

%\vspace{-0.05cm}
\begin{tcolorbox}[width=1.0\linewidth, colframe=blackish, colback=beaublue, boxsep=0mm, arc=3mm, left=1mm, right=1mm, right=1mm, top=1mm, bottom=1mm]
The HOSVD-based tensor whitening EPN, proposed in \cite{me_tensor}, is defined by the following operator $\tG$:
\vspace{-0.15cm}
\begin{align}
&\label{eq:rawcod3}{\left(\tE; \vec{A}_1,\cdots,\vec{A}_r\right)}=\hosvd(\tX),\\
&\label{eq:rawcod4}\tEH=\sgn\!\left(\tE\right)\!\,\left|\!\,\tE\right|^{\gamma}\!\!\!,\quad\qquad\qquad\left(\text{generally }\tEH\!=\!\tGH\!\left(\tE\right)\right)\\
&\label{eq:rawcod5}\tVT=((\tEH\times_{1}\!\vec{A}_1)\,\cdots)\times_{r}\!\vec{A}_r,\!\!\!,\quad\left(\text{think }\tVT=\tX^{\frac{1}{2}}\right)\\
&\label{eq:rawcod6}\tG(\tX)=\sgn(\tVT)\,|\!\tVT|^{\gamma^{*}}\!\!\!\!.%\\
\end{align}
\end{tcolorbox}
\vspace{-0.15cm}

\noindent{In} the above equations, we distinguish the core tensor $\tE$, its power-normalized variant $\tEH$ with factor weights evened out by rising them to the power $0\!<\!\gamma\!\leq\!1$, singular vector matrices $\vec{A}_1,\cdots,\vec{A}_r$ and operation $\times_r$ which is the so-called tensor-product in mode $r$.

As our tensors $\tX$ are super-symmetric, we note that $\vec{A}_1\!=\!\vec{A}_2\!=\!\cdots\!=\!\vec{A}_r$. However, the kernel which is proposed in Section \ref{sec:ker2} leads to a non-symmetric tensor representation. 
We refer the reader to paper \cite{me_tensor} for the detailed description of the above steps.

%\vspace{0.1cm}\hspace{-0.55cm}
%\fbox{\begin{minipage}{25.2em}{
\begin{tcolorbox}[width=1.0\linewidth, colframe=blackish, colback=beaublue, boxsep=0mm, arc=3mm, left=1mm, right=1mm, right=1mm, top=1mm, bottom=1mm]
Eq. \eqref{eq:rawcod4} has a more general form $\tEH\!=\!\tGH\!\left(\tE\right)$, where $\tGH$ can be any power normalizing function \cite{koniusz2018deeper}. In Sec. \hyperref[{sec:epn_interp}]{D}, we derive the exact interpretation of Eq. (\ref{eq:rawcod3}-\ref{eq:rawcod6}) for $\tGH\!=\!\sgn\!\left(\tE\right)(1-(1-\left|\!\,\tE\right|)^{\bar{N}})$ for which $\sgn\!\left(\tE\right)\!\,\left|\!\,\tE\right|^{\gamma}$ is an approximation \cite{koniusz2018deeper}. We prove in Sec. 
%\refapd{sec:epn_interp}{D}
\hyperref[{sec:epn_interp}]{D}
%{\color{red}{D}} 
 that EPN performs in fact a spectral detection of higher-order occurrences of features, the base of fine-grained systems \cite{lin2017improved,koniusz2018deeper}. Figure \ref{fig:epn-dist} illustrates details of such a spectral detection.
\end{tcolorbox}
\vspace{-0.25cm}
%}\end{minipage}}

\begin{figure*}[t]%htbp % left bottom right top
\centering%%%%\vspace{-0.3cm}
\begin{subfigure}[b]{0.195\linewidth}
\centering
\includegraphics[trim=0 0 0 0, clip=true, width=4.13cm]{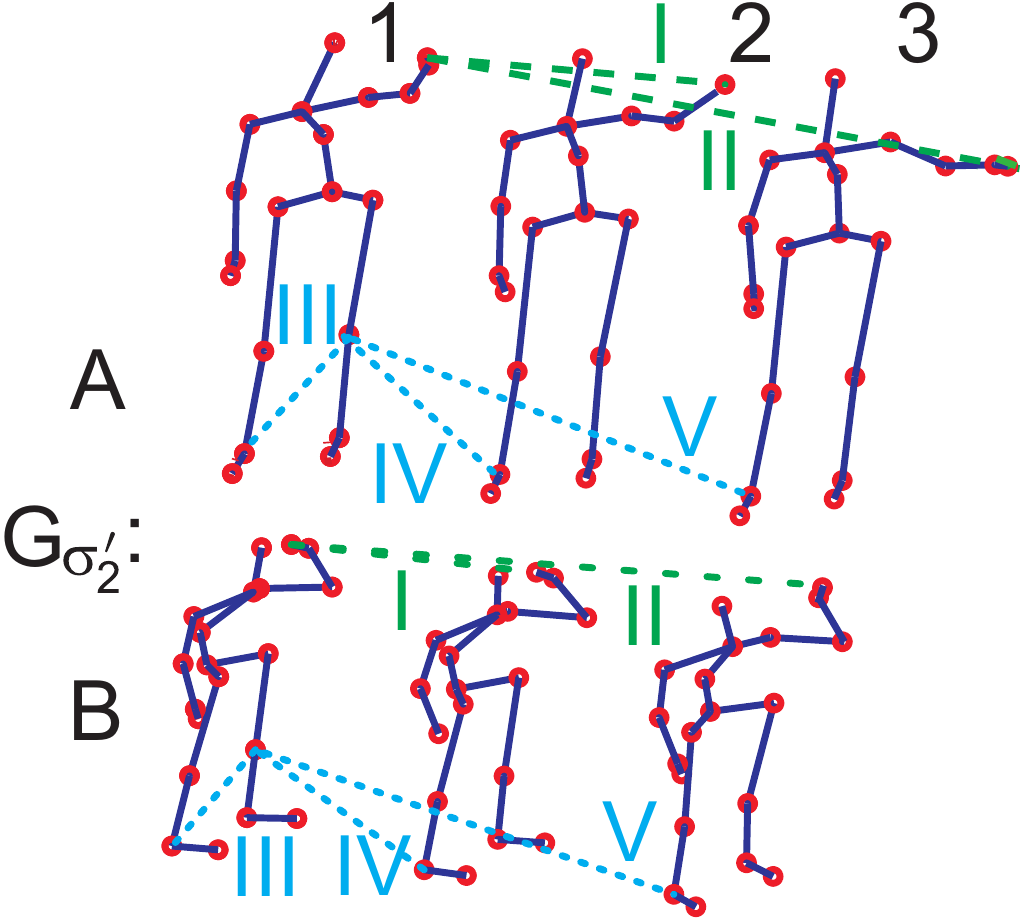}%2.95
%\phantomcaption
\vspace{-0.1cm}
\caption{\label{fig:ker2a}}
\end{subfigure}
\begin{subfigure}[b]{0.135\linewidth}
\centering
\includegraphics[trim=0 0 0 0, clip=true, width=2.24cm]{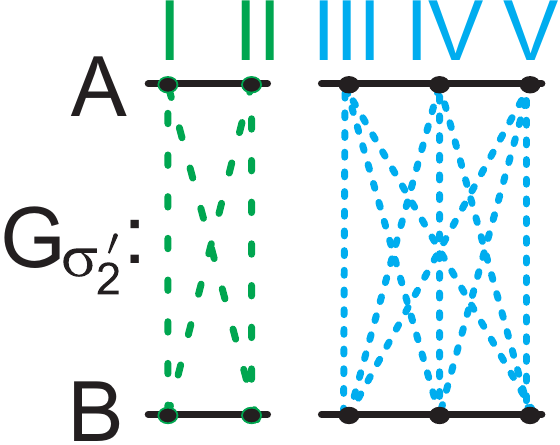}%1.6
\vspace{-0.1cm}
%\phantomcaption
\caption{\label{fig:ker2b}}
\end{subfigure}
\begin{subfigure}[b]{0.65\textwidth}
\centering
\includegraphics[trim=0 0 0 0, clip=true, width=11.2cm]{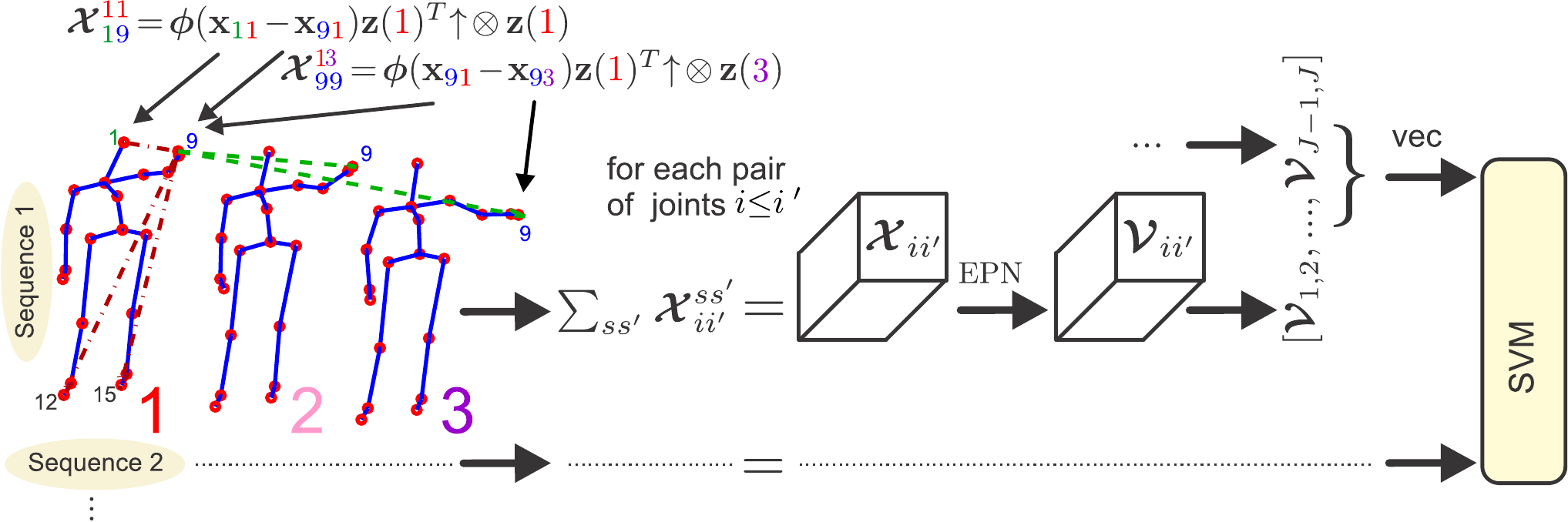}%8.0%ker2c2.pdf
%\phantomcaption
\vspace{-0.1cm}
\caption{\label{fig:ker2c}}
\end{subfigure}
%%%%\vspace{-0.2cm}
\caption{Figure \ref{fig:ker2a} shows that kernel $G_{\sigma'_2}$ in DCK captures spatio-temporal dynamics by measuring displacement vectors from any given body-joint to remaining joints spatially- and temporally-wise (\ie~see dashed lines). Figure \ref{fig:ker2b} shows that comparisons performed by $G_{\sigma'_2}$ for any selected two joints are performed all-against-all temporally-wise which is computationally expensive. Figure \ref{fig:ker2c} shows the encoding steps in the proposed linearization which is fastn. \revised{We collect all $\tX_{ii'}$ for joints $i\!\leq\!i'$, whiten them by EPN to obtain $\tV_{ii'}$, concatenate, vectorize them per video with `vec' and fed to an SVM. We introduced color-coded body joints/frame numbers to show how we assemble a single $\tX_{ii'}$.}}%%%%\vspace{-0.35cm}
\vspace{-0.3cm}
\end{figure*}

\subsection{Dynamics Compatibility Kernel}
\label{sec:ker2}
The SCK kernel that we described above captures the inter-sequence alignment, whereas the intra-sequence spatio-temporal dynamics is lost. %In order to capture these temporal dynamics
Thus, we propose a novel {\em dynamics compatibility kernel} (DCK). In what follows, we use the absolute coordinates of the joints in our kernel and follow notations from the prev. section.

\begin{tcolorbox}[width=1.0\linewidth, colframe=blackish, colback=beaublue, boxsep=0mm, arc=3mm, left=1mm, right=1mm, right=1mm, top=1mm, bottom=1mm]
DCK for two action sequences $\Pi_A$ and $\Pi_B$ is defined as:
\vspace{-0.2cm}
\begin{align}
& \!\!\!\!K_D(\piA,\piB)=\nonumber\\
&\!\!\!\!\frac{1}{\Lambda}\!\!\!\!\!\sum\limits_{\substack{(i,s)\in\idxJ\!,\\(i',s')\in\idxJ\!,\\i'\!\!\neq\!i\!,s'\!\!\neq\!s}}\!\!\!\!\!\!\!\sum\limits_{\quad\;\substack{(\!j,t)\in\idxJ\!,\\(\!j'\!\!,t'\!)\in\idxJ,\\j'\!\!\neq\!j\!,t'\!\!\neq\!t}\!\!\!\!\!\!}\!\!\!\!\!\!\!G'_{\sigma'_1}\!(i\!-\!j\!, i'\!\!-\!j'\!)\,G_{\sigma'_2}\!\left(\left(\vx_{is}\!-\!\vx_{i's'}\!\right)\!-\!\left(\vy_{jt}\!-\!\vy_{j't'}\right)\right)\cdot\nonumber\\[-22pt]
& \qquad\qquad\qquad\quad\;\cdot G'_{\sigma'_3}(\frac{s\!-\!t}{N},\!\frac{s'\!\!-\!t'}{N})\,G'_{\sigma'_4}(s\!-\!s'\!,t\!-\!t'\!).\label{eq:ker2a}
\end{align}
%
%where $G'_{\sigma}(\valpha,\vbeta)\!=\!G_{\sigma}(\valpha)G_{\sigma}(\vbeta)$. 
In contrast to  SCK in~\eqref{eq:ker1a}, the DCK kernel uses the intra-sequence joint differences, thus capturing the dynamics, which is then compared against dynamics of other sequences.
\end{tcolorbox}
\vspace{-0.1cm}

Figures~\ref{fig:ker2a}-\ref{fig:ker2c} depict schematically how DCK captures co-occurrences. As in SCK, the first kernel, $G'_{\sigma'_1}$, captures the sensor uncertainty in body-keypoint detection, and is assumed to be a delta function in this paper. The second kernel, $G_{\sigma'_2}$, models the spatio-temporal co-occurrences of the body-joints. \revised{Temporal alignment kernels, expressed as $G'_{\sigma'_3}(\valpha,\vbeta)\!=\!G_{\sigma'_3}(\valpha)G_{\sigma'_3}(\vbeta)$, encode temporal start- and end-points from $(s,s'\!)$ and $(t,t'\!)$}. Finally, $G_{\sigma'_4}$ limits contributions of dynamics between temporal points if they are distant from each other, \ie~if $s'\!\gg\!s$ or $t'\!\gg\!t$ and $\sigma'_4$ is small. Similarly to SCK, the standard deviations $\sigma'_2$ and $\sigma'_3$ control the selectivity over spatio-temporal dynamics of body-joints and their temporal shift-invariance for the start and end points, resp. \revised{As discussed for SCK, the practical extensions from footnotes\footref{foot:foo0}\textsuperscript{,}\footref{foot:foob}\textsuperscript{,}\footref{foot:temp} also apply to DCK \eg, the definition of $\vz$, the pivot numbers $Z_2$ and $Z_3$ for $G_{\sigma'_2}$ and $G_{\sigma'_3}$ kernels}. %Figure \ref{fig:ker2} provides intuitions into this kernel and what it captures. 

\begin{figure}[t]%htbp % left bottom right top
\centering%%%%\vspace{-0.3cm}
%\hspace{-1.5cm}
%%%%\vspace{-0.2cm}
\includegraphics[trim=0 0 0 0, clip=true, width=9cm]{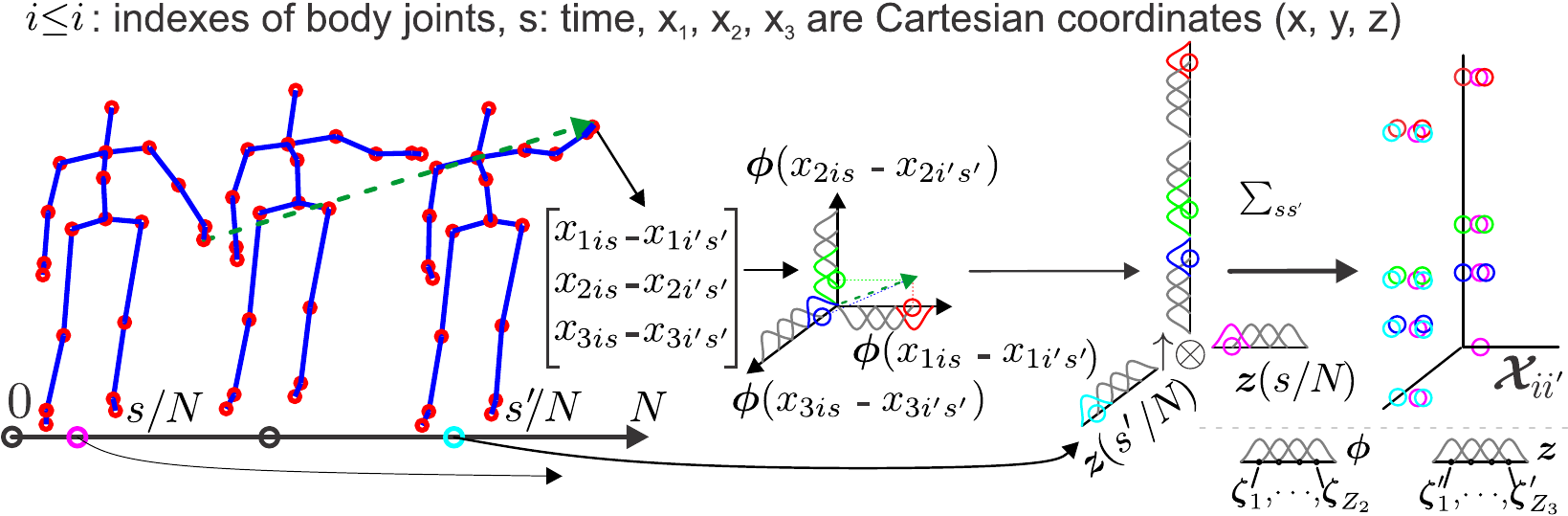}%2.6
\caption{\revised{Third-order statistics from Eq. \eqref{eq:ker2a} can be understood by studying the linearization in Eq. \eqref{eq:ker2b}. For a given pair of joints $i\!\leq\!i'$ at times $s/N$ and $s'/N$ (normalized frame numbers), we embed displacement vectors $\vx_{is}\!-\!\vx_{i's'}$ of 3D joint coordinates $\vx_{is}$and $\vx_{i's'}$ via function $\vphi(\cdot)$ into a non-linear Hilbert space representing an RBF kernel according to Eq. \eqref{eq:gauss_lin2a}. Similarly, we embed the starting and ending times $s/N$ and $s'/N$ via function $\vz(\cdot)$ (also by Eq. \eqref{eq:gauss_lin2a}). Finally, $\otimes$ performs the third-order outer-product on concatenated displacement and time embeddings aggregated next over frames $s$ and $s'$ (note $\sum_{ss'}$). The interpretation: the Gaussians `soft-divide' the Cartesian coordinate system along $x$, $y$, $z$ direction, resp., as well as time direction ($s/N$ and $s'/N$). We project displacements along $x$, $y$, $z$ directions of Cartesian coordinates and assign each projection to Gaussians. Thus, triplets $([x;y;z],s,s')$ assigned into such a `soft-divided' space capture locally displacements of pairs of joints on the time grid (3-way soft-histogram). For DCK$\,\oplus$ in Section \ref{sec:ker4} we use velocity vectors $\frac{\vx_{is}\!-\!\vx_{i's'}}{\max(1,|s'\!\!-\!s|)}$ (\cf displacement vectors) with short- and long-term estimates depending on $s'\!\!-\!s$ (3-way soft-histogram of short- and long-term speeds).}}
\label{fig:phis2}
\vspace{-0.4cm}
\end{figure}

%\!\!\!\!\!\!
%Below, we employ linearization to this kernel. 
Based on the above formulations, Section \hyperref[{sec:dck_der}]{A} shows that the linearization of DCK admits the form:
\vspace{-0.15cm}
\begin{align}
&\!K_D(\Pi_A,\Pi_B) \approx\label{eq:ker2b}\\
&\!\!\!\!\!\!\sum\limits_{\substack{i\in\idx{J}\!,\\i'\!\in\idx{J}\!:\\i'\!\neq i}}
\!\!
\left<
\!\frac{1}{\sqrt{\Lambda}}\!\!\sum\limits_{\substack{s\in\idx{N}\!,\\s'\!\!\in\idx{N}\!:\\s'\!\!\neq\!s}}\!\!
G_{\sigma'_4}(s\!-\!s'\!)\left(\vphi(\vx_{is}\!\!-\!\!\vx_{i's'})
\!\cdot\!\vz\big(\frac{s}{N}\big)^T\!\right)\!\kronstack\vz\big(\frac{s'\!}{N}\big)\!\right.,\nonumber\\[-24pt]
&\qquad\!\qquad\quad\;\;
\left.
\!\frac{1}{\sqrt{\Lambda}}\!\!\sum\limits_{\substack{t\in\idx{N}\!,\\t'\!\!\in\idx{N}\!:\\t'\!\!\neq\!t}}\!\!
G_{\sigma'_4}(t\!-\!t'\!)\Big(
\vphi(\vy_{it}\!\!-\!\!\vy_{i't'})
\!\cdot\!\vz\big(\frac{t}{N}\big)^T\!\Big)\!\kronstack\vz\big(\frac{t'\!}{N}\big)\!\right>\!.\nonumber
%\label{eq:ker2b}
\end{align}
\vspace{-0.15cm}

Equation~\eqref{eq:ker2b} expresses $K_D(\piA,\piB)$ as a sum over inner-products on third-order non-symmetric tensors (c.f. Section \ref{sec:ker1}  where the proposed kernel results in an inner-product between super-symmetric tensors). However, we can decompose each of these tensors with a variant of EPN, which involves Higher Order Singular Value Decomposition (HOSVD), into factors stored in the so-called core tensor, and equalize the contributions of these factors to prevent bursts in the  spatio-temporal co-occurrence dynamics of actions. For example, consider that a long  \emph{hand wave} versus a  short \emph{ hand wave} yield different temporal statistics, that is, the prolonged action results in bursts. However, the final representation  described below becomes invariant to bursts.

\vspace{-0.05cm}
\begin{tcolorbox}[width=1.0\linewidth, colframe=blackish, colback=beaublue, boxsep=0mm, arc=3mm, left=1mm, right=1mm, right=1mm, top=1mm, bottom=1mm]
The final representation for linearized DCK with a non-linear operator $\tG$ introduced into Eq. \eqref{eq:ker2b}
%, as in Section \ref{sec:ker1}, 
to prevent burstiness becomes:
%Thus, as in Section \ref{sec:ker1}, we introduce a non-linear operator $\tG$ into Eq. \eqref{eq:ker2b} and write our final representation \wrt $\piA$ as:
\vspace{-0.55cm}
\begin{align}
&\qquad\qquad\tV_{ii'\!}\!=\!\mygthree{\tX_{ii'\!}}\!\text{, where }\label{eq:ker2d}\\%\substack{s\in\idx{N}\!,\\s'\!\!\in\idx{N}\!:\\s'\!\!\neq\!s}
&\nonumber\\[-19pt]
&\tX_{ii'\!}\!=\!\!\frac{1}{\sqrt{\Lambda}}\!\!\!\!\!\!\!\!\!\!\!\!\!\!\sum\limits_{\qquad\;\;s,s'\!\in\idx{N}:s'\!\!\neq\!s\!\!\!\!}\!\!
\!\!\!\!\!\!\!\!\!\!\!\!G_{\sigma'_4}(s\!-\!s'\!)\left(
\vphi(\vx_{is}\!\!-\!\!\vx_{i's'})
\!\cdot\!\vz\big(\frac{s}{N}\big)^T\!\right)\!\kronstack\vz\big(\frac{s'\!}{N}\big).\nonumber
\end{align}
\vspace{-0.4cm}

\noindent{The} summation over  pairs of body-joint indexes in \eqref{eq:ker2b} is equivalent to the concatenation of $\tV_{ii'}\!$ from \eqref{eq:ker2d} along the fourth mode. Thus, we obtain tensor representations $\big[\tV_{ii'\!}\big]_{i>i'\!:\,i,i'\in\idx{J}}^{\oplus_4}\!$  for sequence $\piA$ and $\big[\tVH_{ii'\!}\big]_{i>i'\!:\,i,i'\in\idx{J}}^{\oplus_4}\!$ for sequence $\piB$.
\end{tcolorbox}
\vspace{-0.2cm}

\revised{The physical meaning of Eq. \eqref{eq:ker2d} is detailed in Figure \ref{fig:phis2}.} The dot-product can be now applied between these representations to compare them. 
Tensors $\tX$ in \eqref{eq:ker2d} are non-symmetric.
Thus, for the operator $\tG$, we choose the HOSVD-based tensor whitening EPN, that is, tEPN defined in Eq. (\ref{eq:rawcod3}-\ref{eq:rawcod6}).

\comment{ as proposed in \cite{me_tensor}. However, they work with the super-symmetric tensors, such as the one we proposed in Section \ref{sec:ker1}. We work with a general non-symmetric case in \eqref{eq:ker2d} and use the following operator $\tG$:
\begin{align}
&{\left(\tE; \vec{A}_1,\cdots,\vec{A}_r\right)}=\hosvd(\tX)\label{eq:rawcod3}\\
&\tEH=\sgn\!\left(\tE\right)\!\,\left|\!\,\tE\right|^{\gamma}\label{eq:rawcod4}\\
&\tVT=((\tEH\otimes_{1}\!\vec{A}_1)\,\cdots)\otimes_{r}\!\vec{A}_r\label{eq:rawcod5}\\
&\tG(\tX)=\sgn(\tVT)\,|\!\tVT|^{\gamma^{*}}\label{eq:rawcod6}%\\
\end{align}
In the above equations, we distinguish the core tensor $\tE$ and its power normalized variants $\tEH$ with factors that are being evened out by rising to the power $0\!<\!\gamma\!\leq\!1$, eigenvalue matrices $\vec{A}_1,\cdots,\vec{A}_r$ and operation $\otimes_r$ which represents a so-called tensor-product in mode $r$. We refer the reader to paper \cite{me_tensor} for the detailed description of the above steps.}

\subsection{Sequence Compatibility Kernel `Plus' (SCK$\,\oplus$)}
\label{sec:ker3}

Below, we extend the SCK formulation from Section \ref{sec:ker1} to aggregate over multiple subsequences extracted from the input sequence. Intuitively, this process is an equivalent of extracting local descriptors from images to attain so-called shift-invariance to the object location. As it is unlikely that relevant motion patterns stretch throughout a sequence, a specific pattern associated with some action classes may appear in one/few subsequences. Moreover, in what follows next, we will allow the aggregation to run over multiple modalities $q\!\in\!\idx{Q}$ \eg, we use 3D body-joints and/or frame-wise CNN classification scores from RGB videos and/or optical flow. Thus, we can define our multimodal pose sequence $\Pi$ as:
\vspace{-0.4cm}
\begin{equation}
\Pi = \set{\vx^{(q)}_{is}\in\mbr{W_q}, i\in\idx{J}, s\in\idx{M}, q\in\idx{Q}},
\end{equation}

\noindent{where} $W_1\!=\!3$, $J$ is the total number of  body-joints, $W_q$ for $q\!>\!1$ equals the size of modality $q$ other than body-joints. Note that if modality $q\!>\!1$ is global rather than per-joint specified, we can replicate it \eg, $\vx^{(q)}_{1s}\!=\!\cdots\!=\vx^{(q)}_{Js}$. 

%\vspace{-0.05cm}
\begin{tcolorbox}[width=1.0\linewidth, colframe=blackish, colback=beaublue, boxsep=0mm, arc=3mm, left=1mm, right=1mm, right=1mm, top=1mm, bottom=1mm]
SCK$\,\oplus$ on a pair of sequences $\Pi_A$ and $\Pi_B$ of length $M$ and $N$ is defined as:
\vspace{-0.4cm}
{\fontsize{8.5}{9}\selectfont
\begin{align}
& \qquad\qquad\qquad\quad K_{S^\oplus}(\Pi_A,\Pi_B) =\label{eq:ker3a}\\
&\!\!\frac{1}{\Lambda}\!\!\sum\limits_{i\in\idx{J}}\sum\limits_{\substack{\tau\in\mathcal{P}_A\\\tau\!'\!\in\mathcal{P}_B}}\sum\limits_{\substack{u\in\mathcal{U}_{\tau}\\u\!'\!\in\mathcal{U}_{\tau\!'}}}\sum\limits_{\substack{s\in\mathcal{S}_{\tau}\\t\in\mathcal{S}_{\tau\!'}}}\!\Big(\sum\limits_{q\in\idx{Q}}\beta^{(q)}_1 G_{\sigma_2^{(q)}}\!\left(\vx^{(q)}_{i,u\!+\!s} - \vy^{(q)}_{i,u\!'\!+\!t}\right) + \nonumber\\[-16pt]
&\qquad\qquad\qquad\qquad\qquad\qquad\;\;\;\;\beta_2\, G_{\sigma_3}(f(s,\mathcal{S}_{\tau})\!-\!f(t,\mathcal{S}_{\tau\!'})) +\nonumber\\
&\qquad\qquad\qquad\qquad\qquad\qquad\;\;\;\;\beta_3\, G_{\sigma_4}(f(u,\mathcal{U}^A_{\tau})\!-\!f(u'\!,\mathcal{U}^B_{\tau\!'})) + \nonumber\\
&\qquad\qquad\qquad\qquad\qquad\qquad\;\;\;\;\beta_4\, G_{\sigma_5}(f(\tau,\mathcal{P}_{A})\!-\!f(\tau'\!,\mathcal{P}_{B})) \Big)^r\!\!.\nonumber
\end{align}
}
\end{tcolorbox}
\vspace{-0.1cm}

\noindent{Symbols} $\mathcal{P}_{A}$ and $\mathcal{P}_{B}$ denote subsequence lengths, $\mathcal{P}_{A}\!=\!\mathcal{P}_{B}\!=\!\mathcal{P}$ is a possible assertion to make, so that \ie~$\mathcal{P}\!=\!\{8,10,12,\cdots,20\}$. Moreover, $\mathcal{U}^A_{\tau}$ and $\mathcal{U}^B_{\tau'\!}$ are sets of all positions in sequences $\pi_A$ and $\pi_B$ for subsequences of lengths $\tau$ and $\tau'\!$, respectively, \ie, if $N\!=\!100$ and $\tau\!=\!20$ then  $\mathcal{U}^A_{20}\!=\!\{1,3,5,\cdots,79\}$ is an example of a possible choice. Furthermore, $\mathcal{S}_{\tau}$ and $\mathcal{S}_{\tau'\!}$ are sets of all sampling positions in subsequences of lengths $\tau$ and $\tau'\!$, \ie, if $\tau\!=\!20$ then $\mathcal{S}_{20}\!=\!\{0,1,2,\cdots,19\}$ is an example of a possible choice. 
We define a function $f(s,\mathcal{S})=\frac{s-\mathcal{S}^{min}}{\mathcal{S}^{max}-{\pi}^{min}}$ which performs normalization on $s$ w.r.t. set $\pi$, and $\mathcal{S}^{min}$ and $\mathcal{S}^{max}$ denote the smallest and largest element of set $\mathcal{S}$, respectively. Moreover, normalizations $f(u,\mathcal{U})$ and $f(\tau,\mathcal{P})$ are defined by analogy, $\Lambda\!=\!\Lambda_A\!\cdot\!\Lambda_B\!=\!(|\idx{J}|\!\cdot\!|\mathcal{P}_A|\!\cdot\!|\mathcal{U}^A_{\tau}|\!\cdot\!|\mathcal{S}_{\tau}|)\!\cdot\!(|\idx{J}|\!\cdot\!|\mathcal{P}_B|\!\cdot\!|\mathcal{U}^B_{\tau}|\!\cdot\!|\mathcal{S}_{\tau'\!}|)$. For simplicity, we do not model the within-sequence similarity between the body joints in contrast to Eq. \eqref{eq:ker1a}, thus we skip $\!G_{\sigma_1}$. Kernels $G_{\sigma_2^{(i)}}$ capture the compatibility between body-joint locations $\vx$ and $\vy$  in a subsequence. Kernel $G_{\sigma_3}$ measures the temporal alignment of two pose snippets in the given two subsequences. Kernel $G_{\sigma_4}$ measures the temporal alignment of two subsequences in two sequences. Lastly, $G_{\sigma_5}$ measures the match of two subsequence lengths. Weight factors $\beta^{(q)}_1\!\geq\!0$ adjust the importance of each modality $q\!\in\!\idx{Q}$. Weight $\beta_2\!\geq\!0$ is the importance of the temporal alignment of snippets within subsequences. Weight $\beta_3\!\geq\!0$ is the importance of the temporal alignment of subsequences within sequences. Weight $\beta_4\!\geq\!0$ is the importance of the match of two subsequence lengths. We let $\sum_q\!\beta^{(q)}_1\!+\!\beta_2\!+\!\beta_3\!+\!\beta_4=1$. Parameters $\sigma_2^{(q)}$ in $G_{\sigma_2^{(q)}}$ and $\beta^{(q)}_1$ are set per modality \eg, for the 3D body-joints we chose $G_{\sigma_2^{(1)}}$ to be an RBF kernel, for frame-wise class predictions obtained from CNNs applied on (i) RGB and (ii) optical flow frames we choose $G_{\sigma_2^{(2)}}$ and $G_{\sigma_2^{(3)}}$ to be linear kernels (with no parameters). As previously, $r$ denotes the order of captured statistics \ie, $r\!=\!3$.

Below, we present the process of linearization of our kernel which follows the reasoning from Section~\ref{sec:kernel_linearization} and Eq.~\eqref{eq:gauss_lin2}. However, we feel it is interesting to show how various kernel components translate to various statistics encoded by the tensor: %We have the 
%
%following:
%\renewcommand{\theenumi}{\Roman{enumi}}%
\renewcommand{\labelenumi}{\roman{enumi}.}
\vspace{-0.3mm}
%\hspace{-1.5cm}
\begin{enumerate}[leftmargin=0.5cm]
\item $\!\!G_{\sigma^{(q)}_2}(\vx-\vy)\approx \vphi^{(q)}(\vx)^T\vphi^{(q)}(\vy)$ (see note\footref{foot:foob}) and, in order to reflect the choice of par. $\sigma^{(q)}_2$ for index $q$, we write $\vphi^{(q)}$,
\item $\!\!G_{\sigma_3}(f(s,\mathcal{S}_{\tau})\!-\!f(t,\mathcal{S}_{\tau\!'}))\approx\vz'(f(s,\mathcal{S}_{\tau})^T\vz'(f(t,\mathcal{S}_{\tau\!'}))$, 
\item $\!\!G_{\sigma_4}(f(u,\mathcal{U}_{\tau})\!-\!f(u'\!,\mathcal{U}_{\tau\!'}))\approx\vz''(f(u,\mathcal{U}_{\tau})^T\vz''(f(u'\!,\mathcal{U}_{\tau\!'}))$,$\!\!\!\!$
\item $\!\!G_{\sigma_5}(f(\tau,\mathcal{P}_{\!A})\!-\!f(\tau'\!,\mathcal{P}_{\!B}))\approx\vz'''(f(\tau,\mathcal{P}_{\!A})^T\vz'''(f(\tau'\!,\mathcal{P}_{\!B}))$.$\!\!\!\!$
\end{enumerate}

With these approximations at hand, we rewrite our sequence compatibility kernel `plus' as:
\vspace{-0.5cm}

{\fontsize{8.5}{9}\selectfont
\begin{align}
&K_{S^\oplus}(\Pi_A,\Pi_B) \approx\nonumber\\
&\nonumber\\[-18pt]
&\!\!\!\!\!\!\!\!\!\!\!\!\frac{1}{\Lambda}\!\!\sum\limits_{i\in\idx{J}}\sum\limits_{\substack{\tau\in\mathcal{P}_{\!A}\\\tau\!'\!\in\mathcal{P}_{\!B}}}\sum\limits_{\substack{u\in\mathcal{U}_{\tau}\\u\!'\!\in\mathcal{U}_{\tau\!'}}}\sum\limits_{\substack{s\in\mathcal{S}_{\tau}\\t\in\mathcal{S}_{\tau\!'}}}\!\!\left(\!
\begin{bmatrix}
\sqrt{\beta^{(1)}_1}\,\vphi(\vx^{(1)}_{i,u\!+\!s})\\
\cdots\\
\sqrt{\beta^{(Q)}_1}\,\vphi(\vx^{(Q)}_{i,u\!+\!s})\\
\sqrt{\beta_2}\,\vz'(f(s,\mathcal{S}_{\tau}))\\
\sqrt{\beta_3}\,\vz''(f(u,\mathcal{U}_{\tau}))\\
\!\sqrt{\beta_4}\,\vz'''(f(\tau,\mathcal{P}_{\!A}))\!\\[3pt]
\end{bmatrix}^T\!\!\!\!\cdot\!
\begin{bmatrix}
\sqrt{\beta^{(1)}_1}\vphi(\vy^{(1)}_{i,u\!'\!+\!t})\\
\cdots\\
\sqrt{\beta^{(Q)}_1}\vphi(\vy^{(Q)}_{i,u\!'\!+\!t})\\
\sqrt{\beta_2}\vz'(f(t,\mathcal{S}_{\tau\!'}))\\
\sqrt{\beta_3}\vz''(f(u'\!,\mathcal{U}_{\tau\!'}))\\
\!\sqrt{\beta_4}\vz'''(f(\tau'\!,\mathcal{P}_{\!B}))\!\\[3pt]
\end{bmatrix}\!\right)^r\!\!\!\!=\!\nonumber\\[-12pt]
&\label{eq:ker3up}\\
\comment{
&\!\!\!\!\!\!\!\!\!\!\!\!\!\!\frac{1}{\Lambda}\!\!\sum\limits_{i\in\idx{J}}\sum\limits_{\substack{\tau\in\mathcal{P}_{\!A}\\\tau\!'\!\in\mathcal{P}_{\!B}}}\sum\limits_{\substack{u\in\mathcal{U}_{\tau}\\u\!'\!\in\mathcal{U}_{\tau\!'}}}\sum\limits_{\substack{s\in\mathcal{S}_{\tau}\\t\in\mathcal{S}_{\tau\!'}}}\!\!\left<\!
\!{\kronstack}_r\!\!\begin{bmatrix}
\sqrt{\beta^{(1)}_1}\,\vphi(\vx^{(1)}_{i,u\!+\!s})\\
\cdots\\
\sqrt{\beta^{(Q)}_1}\,\vphi(\vx^{(Q)}_{i,u\!+\!s})\\
\sqrt{\beta_2}\,\vz'(f(s,\mathcal{S}_{\tau}))\\
\sqrt{\beta_3}\,\vz''(f(u,\mathcal{U}_{\tau}))\\
\!\sqrt{\beta_4}\,\vz'''(f(\tau,\mathcal{P}_{\!A}))\!\\[3pt]
\end{bmatrix}\!\!,
\!{\kronstack}_r\!\!\begin{bmatrix}
\sqrt{\beta^{(1)}_1}\vphi(\vy^{(1)}_{i,u\!'\!+\!t})\\
\cdots\\
\sqrt{\beta^{(Q)}_1}\vphi(\vy^{(Q)}_{i,u\!'\!+\!t})\\
\sqrt{\beta_2}\vz'(f(t,\mathcal{S}_{\tau\!'}))\\
\sqrt{\beta_3}\vz''(f(u'\!,\mathcal{U}_{\tau\!'}))\\
\!\sqrt{\beta_4}\vz'''(f(\tau'\!,\mathcal{P}_{\!B}))\!%\\[3pt]
\end{bmatrix}\!\right>\!.}
&\!\!\!\!\!\!\!\!\!\!\!\!\!\!\sum\limits_{i\in\idx{J}}\!\!\left<
\mygthree{\!\frac{1}{\Lambda_A}\!\!\sum\limits_{\tau\in\mathcal{P}_{\!A}}\!\sum\limits_{u\in\mathcal{U}_{\tau}}\!\sum\limits_{s\in\mathcal{S}_{\tau}}\!\!\!{\kronstack}_r\!\!\begin{bmatrix}
\sqrt{\beta^{(1)}_1}\,\vphi(\vx^{(1)}_{i,u\!+\!s})\\
\cdots\\
\sqrt{\beta^{(Q)}_1}\,\vphi(\vx^{(Q)}_{i,u\!+\!s})\\
\sqrt{\beta_2}\,\vz'(f(s,\mathcal{S}_{\tau}))\\
\sqrt{\beta_3}\,\vz''(f(u,\mathcal{U}_{\tau}))\\
\!\sqrt{\beta_4}\,\vz'''(f(\tau,\mathcal{P}_{\!A}))\!\\[3pt]
\end{bmatrix}}\right.,\!\label{eq:ker3c}\\[-28pt]
&\qquad\qquad\qquad\qquad\quad\,\left.\mygthree{\!\frac{1}{\Lambda_B}\!\!\sum\limits_{\tau\!'\!\in\mathcal{P}_{\!B}}\!\sum\limits_{u\!'\!\in\mathcal{U}_{\tau\!'}}\!\sum\limits_{t\in\mathcal{S}_{\tau\!'}}\!\!\!{\kronstack}_r\!\!\begin{bmatrix}
\sqrt{\beta^{(1)}_1}\vphi(\vy^{(1)}_{i,u\!'\!+\!t})\\
\cdots\\
\sqrt{\beta^{(Q)}_1}\vphi(\vy^{(Q)}_{i,u\!'\!+\!t})\\
\sqrt{\beta_2}\vz'(f(t,\mathcal{S}_{\tau\!'}))\\
\sqrt{\beta_3}\vz''(f(u'\!,\mathcal{U}_{\tau\!'}))\\
\!\sqrt{\beta_4}\vz'''(f(\tau'\!,\mathcal{P}_{\!B}))\!\\[3pt]
\end{bmatrix}}\right>\!\!.\nonumber
\end{align}
}

\noindent{In} the above equation, we set $\mygthree{\tX}\!=\!\tX$ for Eq. \eqref{eq:ker3c} to be equivalent to Eq. \eqref{eq:ker3up}.
However, similarly to considerations in Section \ref{sec:ker1}, a commonly encountered adversity in aggregated representations, the \emph{burstiness}, requires some suppression. To this end, we let operator $\tG$ in Eq. \eqref{eq:ker3c} perform tEPN on the spectrum of the third-order tensor.
%
%\medskip

\begin{tcolorbox}[width=1.0\linewidth, colframe=blackish, colback=beaublue, boxsep=0mm, arc=3mm, left=1mm, right=1mm, right=1mm, top=1mm, bottom=1mm]
The final representation %\eg, for a sequence $\piA$, 
for linearized SCK$\,\oplus$ becomes:
{\fontsize{8.5}{9}\selectfont
\begin{align}
& \tV_i\!=\!\mygthree{\tX_i}\!,\text{ where } \tX_i\!=\!\!\frac{1}{\Lambda_A}\!\!\sum\limits_{\tau\in\mathcal{P}_{\!A}}\sum\limits_{u\in\mathcal{U}_{\tau}}\sum\limits_{s\in\mathcal{S}_{\tau}}\!\!\!{\kronstack}_r\!\!\begin{bmatrix}
\sqrt{\beta^{(1)}_1}\,\vphi(\vx^{(1)}_{i,u\!+\!s})\\
\cdots\\
\sqrt{\beta^{(Q)}_1}\,\vphi(\vx^{(Q)}_{i,u\!+\!s})\\
\sqrt{\beta_2}\,\vz'(f(s,\mathcal{S}_{\tau}))\\
\sqrt{\beta_3}\,\vz''(f(u,\mathcal{U}_{\tau}))\\
\!\sqrt{\beta_4}\,\vz'''(f(\tau,\mathcal{P}_{\!A}))\!%\\[3pt]
\end{bmatrix}\!.\label{eq:ker3d}
\end{align}
}
\vspace{-0.1cm}

We can further replace the summation over the body-joint indexes in \eqref{eq:ker3c} by concatenating $\tV_i$ in ~\eqref{eq:ker3d} along the fourth tensor mode, thus defining $\tV = \big[\tV_i\big]_{i\in\idx{J}}^{\oplus_4}$. Suppose $\tV_A$ and $\tV_B$ are the corresponding fourth order tensors for $\Pi_A$ and $\Pi_B$, then we have: 
\vspace{-0.3cm}
\begin{align}
& K^*_{S^{\oplus}}(\piA,\piB)=\left<\tV_A, \tV_B\right>.
%\left<\big[\tV_i\big]_{i\in\idx{J}}^{\oplus_4}\!,\big[\tVH_i\big]_{i\in\idx{J}}^{\oplus_4}\right>.\label{eq:ker1e}
\end{align}
\end{tcolorbox}
\vspace{-0.1cm}

Note that in general $K^*_{S^{\oplus}}(\piA,\piB)\!\not\approx\!K_{S^{\oplus}}(\piA,\piB)$ as $\tG$ manipulates the spectrum of $\tX$.
Finally, for our final representation, we use only the upper-simplices of $\tV_i$ which consist of $\binom{d+r-1}{r}$ coefficients each, rather than $d^r\!$, where $d$ is the side-dimension of $\tV_i$ \ie, $d\!=\!3Z^{(1)}_2\!+\!\cdots\!+Z^{(Q)}_2\!+\!Z_3\!+\!Z_4\!+\!Z_5$ (see notes\footref{foot:foob}\textsuperscript{,}\footref{foot:temp}), and $Z^{(1)}_2,\cdots,Z^{(Q)}_2$ and $Z_3,Z_4,Z_5$ are the numbers of pivots used in the approximation of $G_{\sigma_2^{(1)}},\cdots,G_{\sigma_2^{(Q)}}$ and $G_{\sigma_3},G_{\sigma_4},G_{\sigma_5}$ (see notes\footref{foot:foob}\textsuperscript{,}\footref{foot:temp}).

\subsection{Dynamics Compatibility Kernel `Plus' (DCK$\,\oplus$)}
\label{sec:ker4}

Below, we apply the aggregation over subsequences to our DCK kernel. We follow the same steps as for SCK$\,\oplus$ (Section \ref{sec:ker3}) except that our subsequences for DCK$\,\oplus$  have a fixed length. 
For a pair of sequences $\Pi_A$ and $\Pi_B$ of length $M$ and $N$, we have:
\begin{align}
& K_{D^\oplus}(\Pi_A,\Pi_B) =\label{eq:ker_dyn_pp}\\
%&\qquad\frac{1}{\Lambda'}\!\!\sum\limits_{\substack{u\in\mathcal{U}_{\tau}\\u\!'\!\in\mathcal{U}_{\tau}}}\!\!\! K_{D}(\Pi'\!_{A,\tau,u},\Pi'\!_{B,\tau,u'\!})G_{\sigma_4}(f(u,\mathcal{U}^A_{\tau})\!-\!f(u'\!,\mathcal{U}^B_{\tau})),\nonumber
&\qquad\frac{1}{\Lambda'}\!\!\sum\limits_{u,u'\!\in\mathcal{U}_{\tau}}\!\!\! K_{D}(\Pi'\!_{A,\tau,u},\Pi'\!_{B,\tau,u'\!})G_{\sigma_4}(f(u,\mathcal{U}^A_{\tau})\!-\!f(u'\!,\mathcal{U}^B_{\tau})),\nonumber
\end{align}
where $\tau$ is a length of subsequences. $K_{D}(\Pi'\!_{A,\tau,u},\Pi_{B'\!,\tau,u'\!})$ is defined in Eq. \eqref{eq:ker2a}. \revised{However, we use velocity vectors $\frac{\vx_{is}\!-\!\vx_{i's'}}{\max(1,|s'\!\!-\!s|)}$ (\cf displacement vectors in DCK) with short- and long-term estimates depending on $s'\!\!-\!s$.  Figure \ref{fig:phis2} provides an interpretation of this kernel.} 
$K_{D}(\Pi'\!_{A,\tau,u},\Pi_{B'\!,\tau,u'\!})$ is evaluated over subsequences $\Pi'\!_{A,\tau,u}$ and $\Pi'\!_{B,\tau,u'\!}$ sampled from  $\Pi_A$ and $\Pi_B$ according to sets of sampling coordinates  $\mathcal{S}_{\tau,u}\!=\!\{\mathcal{S}_{\tau}\}\!+\!u$ and $\mathcal{S}_{\tau,u'\!}\!=\!\{\mathcal{S}_{\tau\!'\!}\}\!+\!u'\!$ of length $\tau$ which are shifted by locations $u$ and $u'\!$ according to $\mathcal{U}_{\tau}$. Lastly, $\Lambda'\!=\!|\mathcal{U}^A_{\tau}|\!\cdot\!|\mathcal{U}^B_{\tau}|$. The remaining symbols follow definitions in Section \ref{sec:ker3}. Kernel in Eq. \eqref{eq:ker_dyn_pp} is then linearized in the similar manner to Eq. \eqref{eq:ker2a} which results in linearization similar to Eq. \eqref{eq:ker2d} but containing an additional mode corresponding to linearization of kernel $G_{\sigma_4}$. We skip this derivation for brevity. 

\comment{
\section{Computational Complexity}
Non-linearized SCK with ker. SVM has complexity $\bigoh(JN^2T^\rho)$ given $J$ body joints, $N$ frames per sequence, $T$ sequences, and $2\!<\!\rho\!<\!3$ which concerns complexity of kernel SVM. Linearized SCK with linear SVM takes $\bigoh(JNTZ_*^r)$ for a total of $Z_*$ pivots and tensor order $r\!=\!3$. Note that $N^2T^\rho\!\gg\!NTZ_*^r$. For $N\!=\!50$ and $Z_*\!=\!20$,  this is $3.5\!\times$ (or $32\!\times$) faster than the exact kernel for $T\!=\!557$ (or $T\!=\!5000$) used in our experiments. Non-linearized DCK with kernel SVM has complexity $\bigoh(J^2N^4T^\rho)$ while linearized DCK takes $\bigoh(J^2N^2TZ^3)$ for $Z$ pivots per kernel, \eg~$Z\!=\!Z_2\!=\!Z_3$ given $G_{\sigma'_2}$ and $G_{\sigma'_3}$. As $N^4T^\rho\!\gg\!N^2TZ^3$, the linearization is $~11000\!\times$ faster than the exact kernel, for say $Z\!=\!5$. Note that EPN incurs negligible cost (see~\cite{tensor_arxiv} for details). Linearized SCK$\,\oplus$ with linear SVM also takes $\bigoh(JNTZ_*^r)$ for a total of $Z_*$, however, $Z_*\!=\!3Z_2\!+\!Z_3\!+\!Z_4\!+\!Z_5$ thus $Z_*\!=\!28$. The linearized DCK takes $\bigoh(J^2N^2TZ^3Z_6)$ where $Z_6\!=\!4$ in our experiments.
}
\section{Experiments}\label{sec:exp}
Below, we present experiments on our models on seven popular datasets. For datasets based on 3D skeletons, we use (i) the UTKinect-Action~\cite{xia_utkinect}, (ii) Florence3D-Action~\cite{seidenari_florence3d}, (iii) MSR-Action3D~\cite{li_msraction3d}, and (iv) Kinetics \cite{kinetics_400} (where stated). For datasets based on RGB frames, we use (v) the fine-grained MPII Cooking Activities~\cite{rohrbach2012database} and (vi) HMDB-51~\cite{kuehne2011hmdb} datasets. For experiments on the 3D skeletons fused with RGB frames, we use (vii) large scale NTU-RGBD~\cite{shahroudy2016ntu} dataset. We also evaluate the influence of various hyper-parameters, such as the number of pivots $Z$ used for linearizing the body-joint and temporal kernels, and the impact of Eigenvalue Power Normalization (we vary the factor equalization). We evaluate our older SCK and DCK kernels, and their newer counterparts SCK$\,\oplus$ and DCK$\,\oplus$. For skeletons, we feed them directly to our kernel representations while RGB-based datasets are firstly encoded by the two-stream CNN \cite{twostream} or the I3D \cite{i3d_net}.

\subsection{Datasets}
\label{sec:sets}
\noindent\textbf{UTKinect-Action~\cite{xia_utkinect}}  consists of 10 actions performed twice by 10 different subjects, and has 199 action sequences. The dataset provides 3D coordinate annotations of 20 body-joints for every frame. The dataset was captured with a stationary Kinect sensor and contains significant viewpoint and intra-class variations.
\\
\noindent\textbf{Florence3D-Action~\cite{seidenari_florence3d}} dataset consists of 9 actions performed 2--3$\times$ by 10 different subjects and it has 215 action sequences. 3D coordinate annotations of 15 body-joints captured with a Kinect sensor are provided. Significant intra-class variations are present \ie, the same action articulated with the left/right hand, and  actions like \emph{drinking}/\emph{performing a phone call} can be seen as fine-grained.
\\
\noindent\textbf{MSR-Action3D~\cite{li_msraction3d}} dataset is comprised of 20 actions performed 2--3$\times$ by 10 different subjects and it has 567 action sequences. 3D coordinates of 20 body-joints captured by a Kinect-like depth sensor are provided. MSR-Action3D has strong inter-class similarity.

\begin{figure}
	\centering   
	\includegraphics[width=3.5cm]{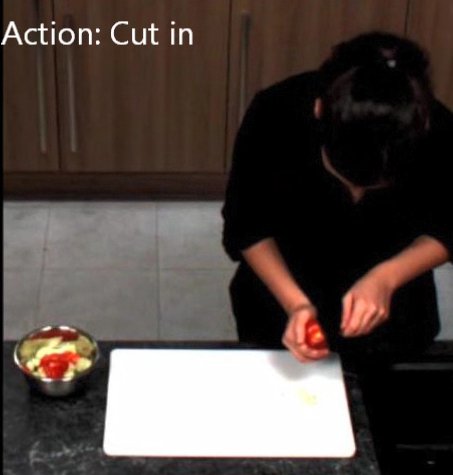}\hspace{0.3cm}
	\includegraphics[width=3.5cm]{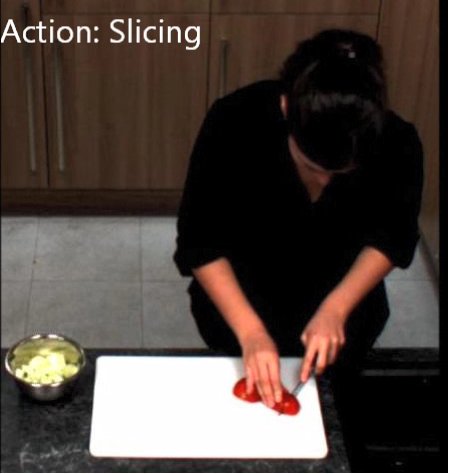}
	\caption{Fine-grained action instances (MPII Cooking Activities \cite{rohrbach2012database}) from two different action categories: \emph{cut-in} (left) and \emph{slicing} (right). %These instances are from the MPII Cooking Activities dataset~\cite{rohrbach2012database}.
	}
	\vspace{-0.2cm}
     \label{fig:mpii}
\end{figure}

In the above datasets, we use the cross-subject test setting (unless stated otherwise), in which half of the subjects  are used for training and the remaining half for testing. Similarly, we divide the training set into two halves for the purpose of training/validation. %For MSR-Action3D,  we use two protocols according to approaches~\cite{wu_actionlets} and ~\cite{li_msraction3d}, where the latter protocol groups related actions into three subsets.
\\
\noindent\textbf{NTU-RGBD~\cite{shahroudy2016ntu}} is by far the largest 3D skeleton-based video action recognition dataset. It has 56880 video sequences across 60 classes, 40 subjects, and
80 views. The videos have on average 70 frames and consist of people performing various actions. Each frame is annotated with 25 human skeletal keypoints (some videos have multiple subjects). Two evaluation protocols are used for this dataset, namely, cross-subject and cross-view evaluation. This dataset can be considered as having many fine-grained classes \eg, \emph{make a phone call}, \emph{playing with phone}, 
%\emph{put the palms together}, \emph{cross hands in front}, 
\emph{punching other person}, \emph{pushing other person}, \emph{pat on back of other person}, \etc.
\\
\noindent\textbf{MPII Cooking Activities~\cite{rohrbach2012database}} dataset consists of high-resolution videos of cooking activities/people cooking various dishes. %; each video contains a single person cooking a dish. %and overall there are 12 such videos in the dataset. 
There are 64 distinct activities spread across 3748 video clips and one background activity (1861 clips). Activities include coarse actions \eg, \emph{opening refrigerator}, and fine-grained actions \eg, \emph{peel}, \emph{slice}, \emph{cut apart} (see Figure \ref{fig:mpii}). This dataset is challenging due to (i) unbalanced action classes, %-- there are certain activities that have only about 1K frames over the entire dataset
 (ii) significant intra-class differences (each subject %are only asked to prepare one from set of 14 possible dishes, and 
 cooks according to their own style). %Further, there are neither any annotations of objects in the scene, nor the tools used for actions are clearly visible (such as spice folder, knife, \etc). 
%Figure \ref{fig:mpii} shows an example fine-grained activity from this dataset. 
We use the mean Average Precision (mAP) over 7-fold cross-validation.
\\
\noindent\textbf{HMDB-51~\cite{kuehne2011hmdb}} dataset is a popular video benchmark for human action recognition, consisting of 6766 Internet videos over 51 classes. Each video has about 20--1000 frames. We report the average classification accuracy on standard three-fold splits.
\\
\revised{
\noindent\textbf{Kinetics} \cite{kinetics_400} contains $\sim$300000 clips from YouTube which cover 400 human action classes, ranging from daily activities, sports scenes, to
complex interactions. Each clip is $\sim$10 seconds long.
}

%We follow the standard protocols in all our experiments.

\subsection{Experimental Setup}\label{sec:setup}

For our experiments, we distinguish four configurations: (i) for UTKinect-Action, Florence3D-Action and MSR-Action3D that provide 3D body-joints, we feed sequences of 3D body-joints to our kernel(s), (ii) for MPII Cooking Activities, HMDB-51 and NTU-RGBD that provide RGB frames, we train a two-stream ResNet-152 model (as in \cite{twostream}) taking  RGB frames (in the spatial stream) and a stack of optical flow frames (in the temporal stream) as input to obtain classification scores per frame per stream which are then passed to our kernel, (iii) for NTU-RGBD which contains both 3D body-joints and RGB frames, we investigate both such inputs separately as well as their combination, and (iv) for Kinetics, we use skeletons and combine ST-GCN with SCK.

For the sequence compatibility kernel on sequences of 3D body-joints, we first normalized all body-keypoints with respect to the hip joints across frames, as indicated in Section \ref{sec:ker1}. We also normalized lengths of all body-parts w.r.t. to a reference skeleton. This setup follows pre-processing of~\cite{vemulapalli_SE3}. For our dynamics compatibility kernel, we use unnormalized body-joints and assume that the displacements of body-joint coordinates across frames capture their temporal evolution implicitly. For the sequence compatibility kernel on classifier scores, we take the scores before they are passed through the logistic function and we apply a rectifier.
%We assume that, as time progresses, corresponding body-joints evolve temporally. Their displacement is captured to indicate the change in motion between frames.

\vspace{0.05cm}
\noindent{\textbf{CNN Training.}} \revised{To extract features with CNN, we train a two-stream ResNet-152 model \cite{twostream} taking  RGB frames (in the spatial stream) and a stack of optical flow frames (in the temporal stream) from a given training split as input. For optical flow, we use the Large Displacement Optical Flow (LDOF) \cite{brox_largedisp}. We use the classifier predictions from each stream as inputs to our kernels. The two streams of the CNN are trained separately on the respective input modalities against a softmax cross-entropy loss. We simply follow the standard training protocols from \cite{twostream}.} For fine-tuning, we used a fixed learning rate of $1e\!-\!4$ and a momentum of 0.9. For the MPII Cooking Activities dataset, we used the sequences from the training set for training the CNNs (1992 sequences) and  those from the validation set (615 sequences) to check for overfitting. For HMDB-51, we use three standard splits provided with the dataset. For NTU-RGBD dataset in the cross-subject evaluation, the training and testing sets have 40320 and 16560 samples, respectively. For NTU-RGBD dataset in the cross-view evaluation, the training and testing sets have
37920 and 18960 samples, respectively. We use 70\% of the training set for training and 30\% for validation. \revised{To train SVM, we simply vectorize our tensors and set  $c\!=\!1e\!-\!2$}.

\begin{figure*}[t]%htbp % left bottom right top
\centering\hspace*{-0.1cm}
\begin{subfigure}[b]{0.245\linewidth}
\centering\includegraphics[trim=0 3 0 15, clip=true, width=4.8cm]{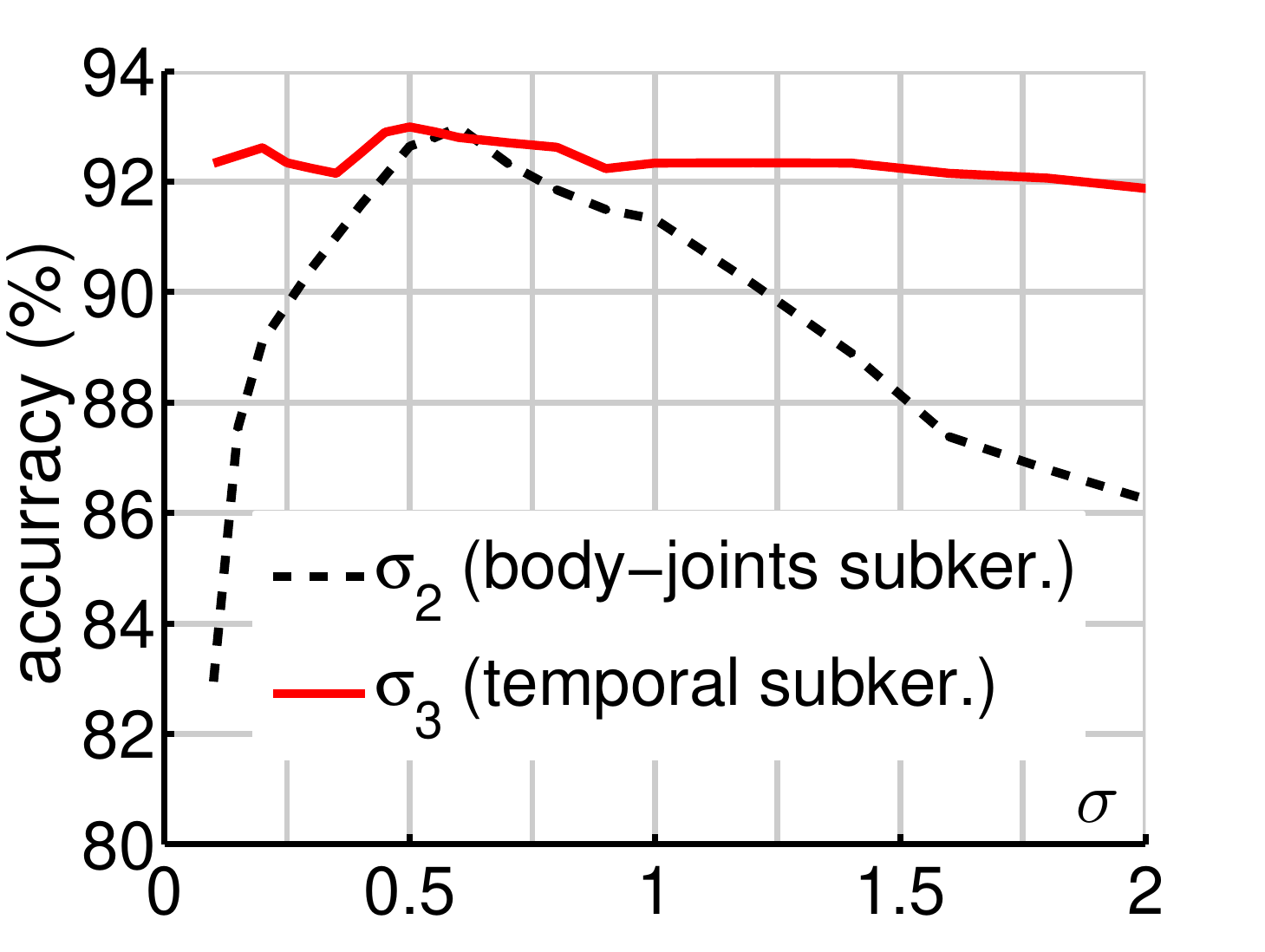}%4.35
\caption{\label{fig:piv1}}
\end{subfigure}
%\hspace*{-0.05cm}
\begin{subfigure}[b]{0.245\linewidth}
\centering\includegraphics[trim=0 3 0 15, clip=true, width=4.8cm]{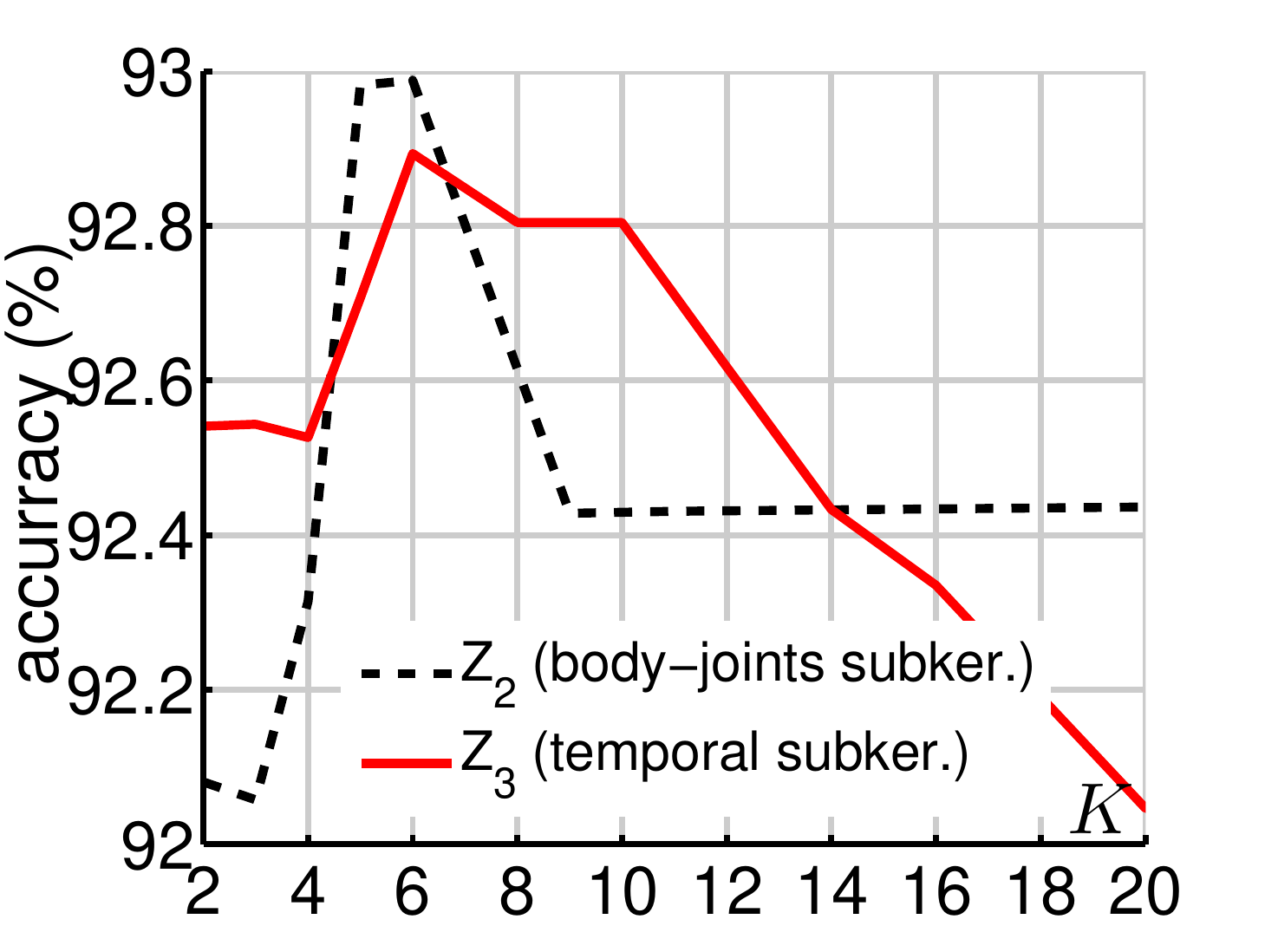}%4.35
\caption{\label{fig:piv2}}
\end{subfigure}
%\hspace*{0.2cm}
\begin{subfigure}[b]{0.245\linewidth}
\includegraphics[trim=0 3 0 15, clip=true, width=4.8cm]{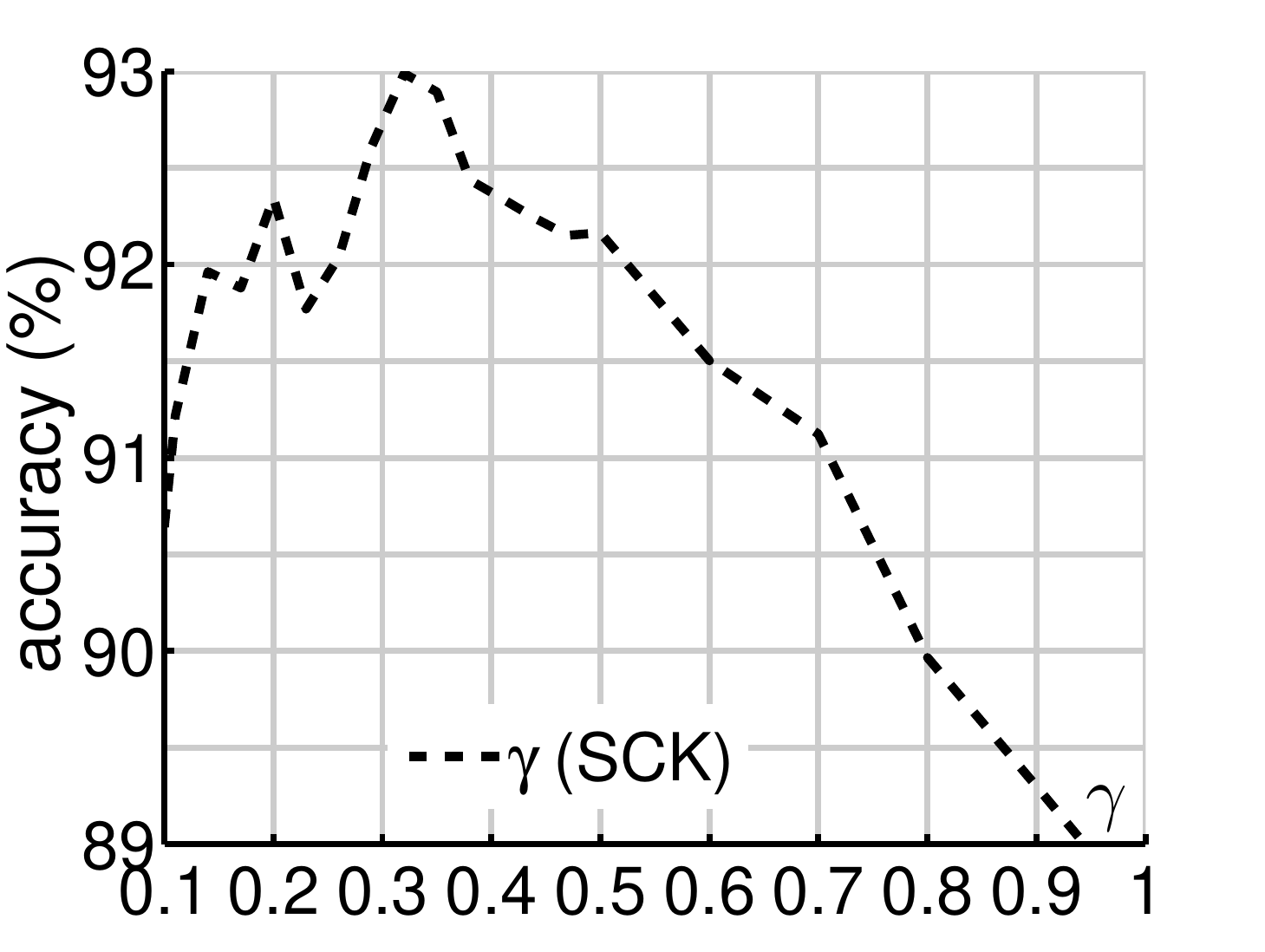}%4.35
\caption{\label{fig:piv3}}
\end{subfigure}
%\hspace*{0.2cm}
\begin{subfigure}[b]{0.245\linewidth}
\centering\includegraphics[trim=0 3 0 10, clip=true, width=4.8cm]{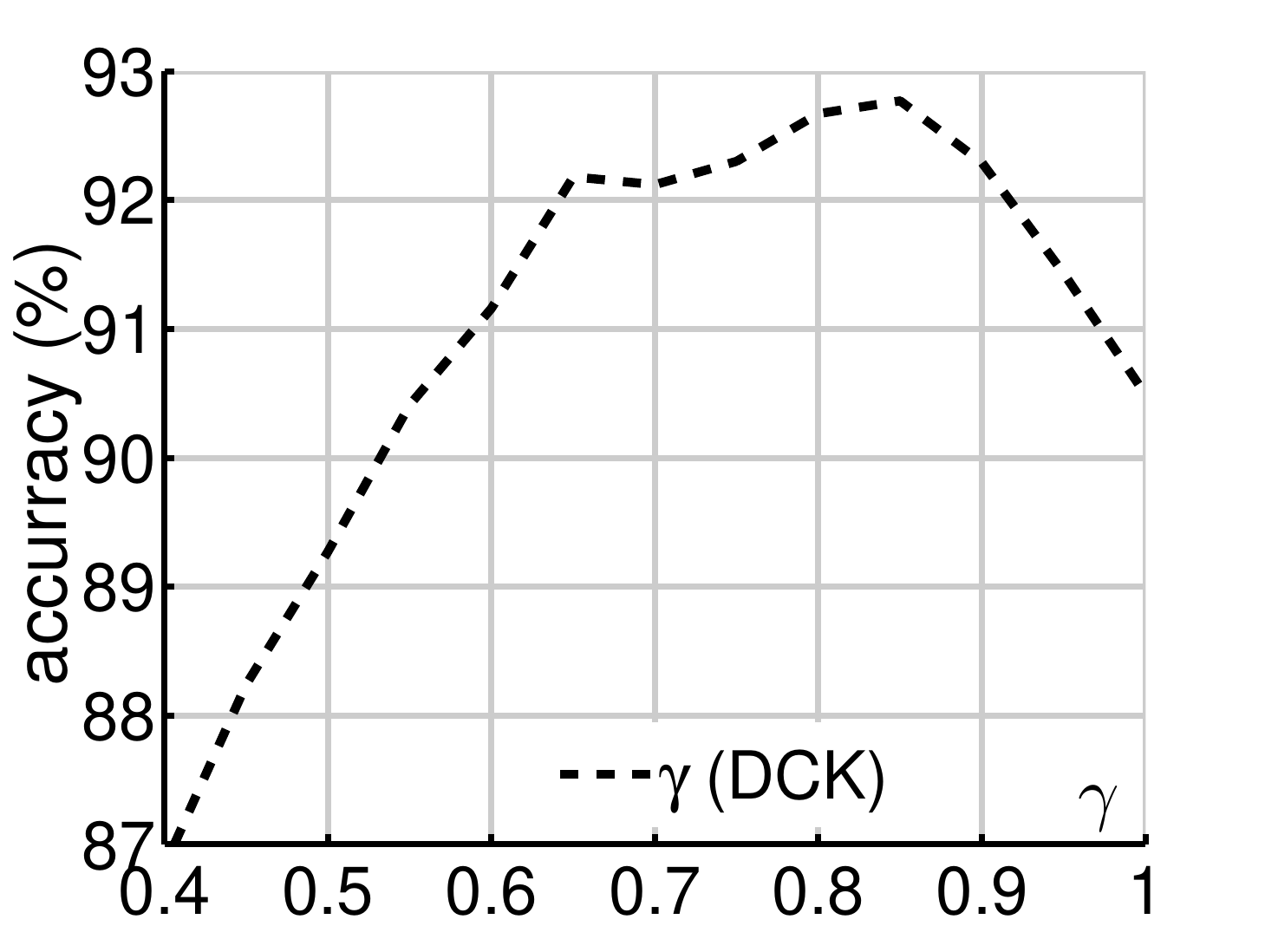}%4.35
\caption{\label{fig:piv6}}
\end{subfigure}
%\vspace{-0.1cm}
\vspace{-0.1cm}
\caption{Figure \ref{fig:piv1} illustrates the classification accuracy on Florence3d-Action for the sequence compatibility kernel when varying radii $\sigma_2$ (body-joints subkernel) and $\sigma_3$ (temporal subkernel). Figure \ref{fig:piv2} evaluates behavior of SCK w.r.t. the number of pivots $Z_2$ and $Z_3$. Figure \ref{fig:piv3} demonstrates effectiveness of our slice-wise Eigenvalue Power Normalization in tackling burstiness by varying parameter $\gamma$. Figure \ref{fig:piv6} shows effectiveness of equalizing the factors in non-symmetric tensor representation by HOSVD Eigenvalue Power Normalization by varying $\gamma$.}
\label{fig:par1}
\vspace{-0.2cm}
\end{figure*}
\begin{figure}[t]%htbp % left bottom right top
\centering\hspace{-0.22cm}
\begin{subfigure}[t]{0.495\linewidth}
\centering\includegraphics[trim=0 0 0 0, clip=true, width=3.74cm]{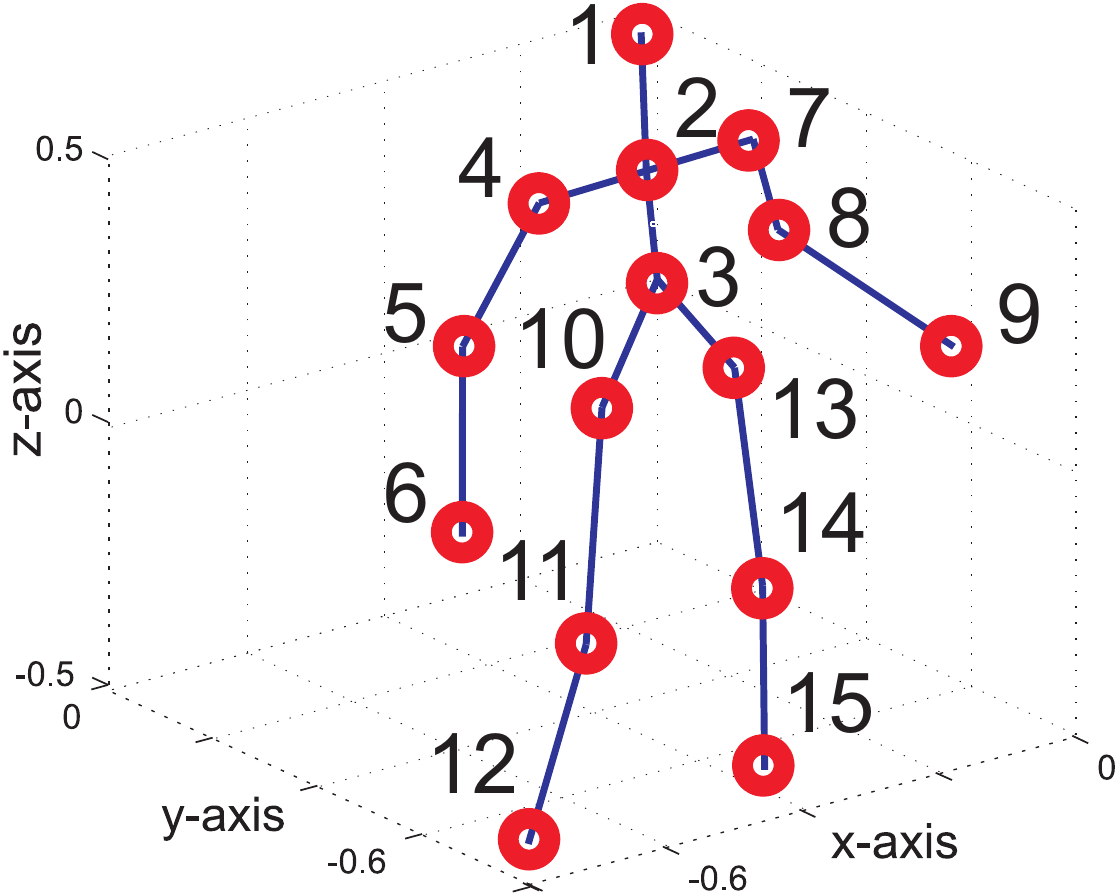}%3.2
\renewcommand{\arraystretch}{0.95}
{
\resizebox{\columnwidth}{!}{
%\tiny
%\scriptsize
%\fontsize{7.5}{8}\selectfont
\begin{tabular}{ c | c | c | c | c }
\kern-0.3em A\kern-0.3em & \kern-0.3em B\kern-0.3em & \kern-0.3em C\kern-0.3em & \kern-0.3em D\kern-0.3em & \kern-0.3em E\kern-0.3em\\
\hline
\kern-0.3em 6,9\kern-0.5em & \kern-0.3em 1,6,9\kern-0.3em & \kern-0.1em 6,9,12,15\kern-0.1em & \kern-0.1em 4,6,7,9,11,14\kern-0.1em & \kern-0.1em 4,6,7,9,\kern-0.1em\\
\cline{1-4}
F & G & H & I & \kern-0.3em 11,12,\kern-0.3em\\
\kern-0.3em 4-15\kern-0.1em & \kern-0.1em 1,4-15\kern-0.1em & \kern-0.3em 1,2,4-15\kern-0.3em & \kern-0.3em 1-15\kern-0.3em & \kern-0.3em 14,15\kern-0.3em\\
\hline
\end{tabular}
}
}
%
%\vspace{-0.4cm}
%\caption{Subsets A-I of the body-joints illustrated in Figure \ref{fig:piv4} which were used to prepare the evaluations in Figure \ref{fig:piv5}.}\label{tab:joints}\vspace{-0.4cm}
%\end{table}
%
\caption{\label{fig:piv4}}
\end{subfigure}
\hspace{-0.16cm}
\begin{subfigure}[b]{0.495\linewidth}
\centering\includegraphics[trim=0 3 0 10, clip=true, width=4.2cm]{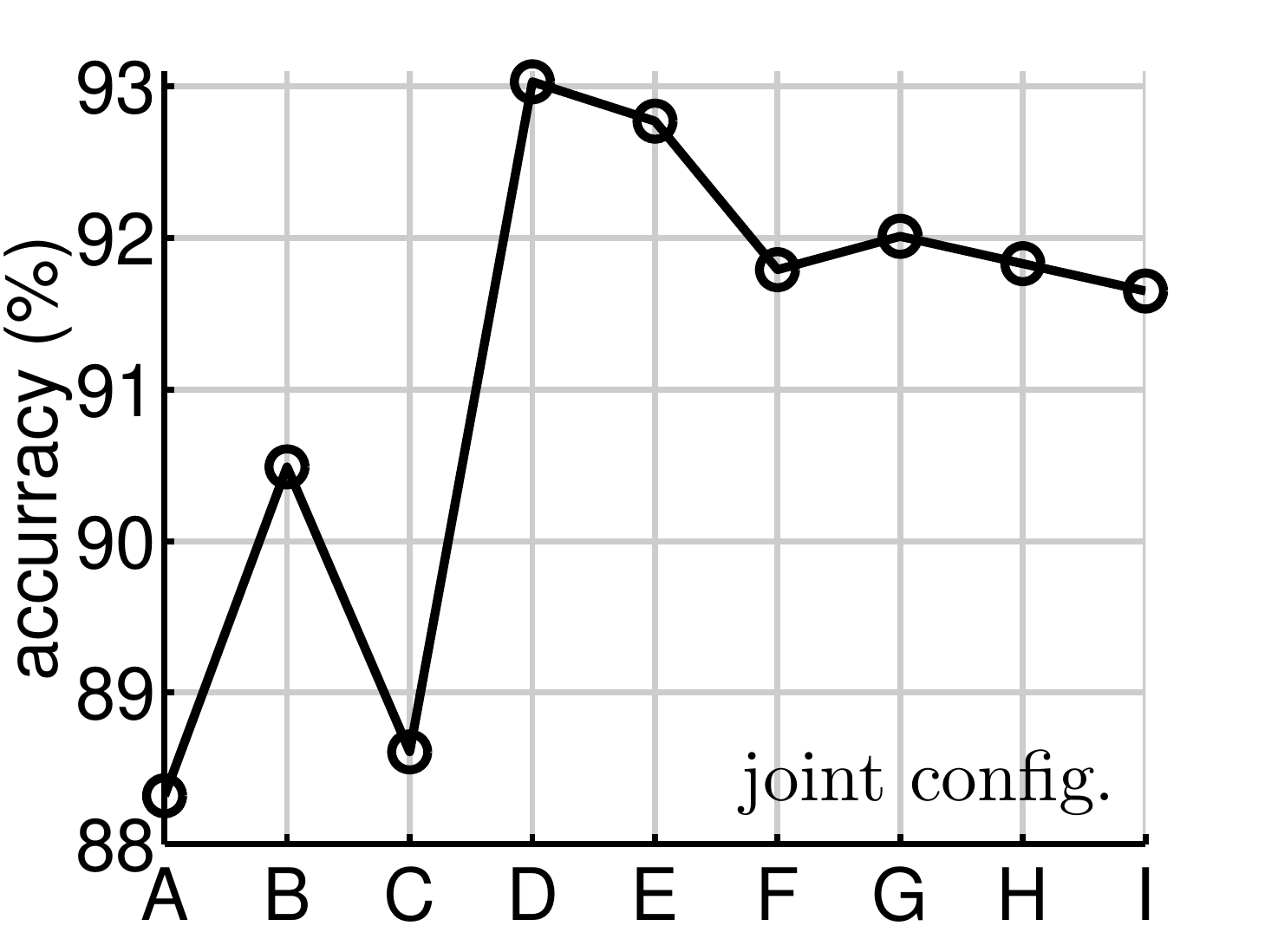}%4.35
\caption{\label{fig:piv5}}
\vspace{-1.0cm}
\end{subfigure}
%\hspace*{0.00cm}
%\begin{subfigure}[b]{0.31\linewidth}
%\centering\includegraphics[trim=0 3 0 10, clip=true, width=5.09cm]{images/piv5c.pdf}%4.35
%\caption{\label{fig:piv6}}
%\end{subfigure}
\vspace{-0.1cm}
\caption{Figure \ref{fig:piv4} enumerates the body-joints in the Florence3D-Action dataset. The table below lists subsets A-I of the body-joints used to build  representations eval. in Figure \ref{fig:piv5}, which shows the accuracy of our dynamics compatibility kernel w.r.t. these subsets.}
\label{fig:par2}
\vspace{-0.2cm}
\end{figure}

\revised{To stay competitive \wrt the state of the art, we additionally use two newer backbones such as (i) Spatial Temporal Graph Convolutional Network %for Skeleton-Based Action Recognition
 (ST-GCN) \cite{stgcn} and (ii) Two-Stream Inflated 3D ConvNet (I3D) \cite{i3d_net}. For ST-GCN, we train it on skeletal sequences from NTU and Kinetics \cite{kinetics_400} datasets following the standard protocols. For Kinetics, we follow approach \cite{kinetics_400} and use skeletons extracted with OpenPose \cite{cao2018openpose}. Finally, we combine our vectorized tensors from SCK or SCK$\,\oplus$ with the output of the last layer of ST-GCN preceding the classifier, and feed such a representation into the cross-entropy loss. As SCK is a shallow approach, we expect it to be highly complementary with ST-GCN.
For I3D network, we train it on subsequences extracted from HMDB51 and MPII. We use RGB and optical flow (LDOF) streams. We extract subsequences of length 48, 64, 80, 96 given strides 1, 2 and 3. Then, subsequences shorter than 64 are lapped. We put together all training subsequences of all lengths and all strides, and we train RGB and LDOF I3D networks separately with a learning rate $1e\!-\!4$ halved every 10 epochs.
}

\vspace{0.05cm}
\noindent{\textbf{IDT Features.}} On HMDB-51 and MPII Cooking Activities, we also report accuracy when our kernel is combined with dense trajectories \cite{dense_traj} encoded by Fisher Vectors \cite{perronnin_fisherimpr}.

%%%%\vspace{0.05cm}
\subsection{Sequence compatibility kernel.} In this section, we first present experiments evaluating the influence of parameters $\sigma_2$ and $\sigma_3$ of kernels $G_{\sigma_2}$ and $G_{\sigma_3}$ which control the degree of selectivity for the 3D body-joints and the temporal shift invariance, respectively. See Section \ref{sec:ker1} for a full definition of these parameters. 
Recall that kernels $G_{\sigma_2}$ and $G_{\sigma_3}$ are approximated via linearizations according to Eq. \eqref{eq:gauss_lin} and \eqref{eq:gauss_lin2}. The quality of these approximations and the size of our final tensor representations depend on the numbers $Z_2$ and $Z_3$ of pivots chosen. \revised{See Section \ref{sec:kernel_linearization}, Figure \ref{fig:phis1} and notes\footref{foot:foob},\footref{foot:temp} for details on pivots.} In our experiments, the pivots $\vzeta$ are spaced uniformly within interval $[-1;1]$ and $[0;1]$ for kernels $G_{\sigma_2}$ and $G_{\sigma_3}$ respectively.

Figures \ref{fig:piv1} and \ref{fig:piv2} present the results of this experiment on the Florence3D-Action dataset. % -- these are the results presented on the test set as we have also observed exactly the same trends on the validation set.
Figure \ref{fig:piv1} shows that the body-joint compatibility subkernel $G_{\sigma_2}$ requires a choice of $\sigma_2$, which is not too strict as specific body-joints (\eg, elbow) are expected to repeat across sequences in similar locations due to zero-centering \wrt hip. On the one hand, very small $\sigma_2$ leads to poor generalization. On the other hand, very large $\sigma_2$ allows big displacements of the corresponding body-joints between sequences which results in a poor discriminative power of this kernel. Furthermore, Figure \ref{fig:piv1} demonstrates that the range of $\sigma_3$ for the temporal subkernel for which we obtain very good performance is large. However, as $\sigma_3$ becomes very small or very large, extreme temporal selectivity or full temporal invariance, respectively, result in a loss of performance. For instance, $\sigma_3\!=\!4$ results in $91\%$ accuracy only.

In Figure \ref{fig:piv2}, we show the performance of our SCK kernel with respect to the number of pivots used for linearization. For the body-joint compatibility subkernel $G_{\sigma_2}$, we see that $Z_2\!=\!5$ pivots are sufficient to obtain good performance of $92.98\%$ accuracy. We have observed that this is consistent with the results on the validation set. Using more pivots, say $Z_2\!=\!20$, deteriorates the results slightly, suggesting overfitting. We make similar observations for the temporal subkernel $G_{\sigma_3}$ which demonstrates a good performance for as few as $Z_3\!=\!2$ pivots. Such a small number of pivots suggests that linearizing 1D variables and generating higher-order co-occurrences, as described in Section~\ref{sec:ker1}, constitute on a simple, robust, and effective linearization strategy.

Furthermore, Figure \ref{fig:piv3} demonstrates the effectiveness of our slice-wise Eigenvalue Power Normalization described in Eq. \eqref{eq:epn1}. When $\gamma\!=\!1$, the EPN functionality is absent. This results in a drop of performance from $92.98\%$ to $88.7\%$ accuracy. This demonstrates that statistically unpredictable bursts of actions described by  body-joints, such as long versus short \emph{hand waving}, are indeed undesirable. It is clear that in such cases, EPN is very effective, as it deals with correlated bursts, \eg~co-occurring \emph{hand wave} and associated with it elbow and neck motion. For more details regarding this concept, see~\cite{me_tensor}. For our further experiments, we choose $\sigma_2\!=\!0.6$, $\sigma_3\!=\!0.5$, $Z_2\!=\!5$, $Z_3\!=\!6$, and $\gamma\!=\!0.36$, as dictated by cross-validation.

\subsection{Dynamics compatibility kernel.}
Below, we evaluate the influence of  parameters of the DCK kernel. Our experiments are based on the 
Florence3D-Action dataset. For ablations, we present results on the test set as results on the validation set match test results closely. As this kernel considers all spatio-temporal co-occurrences of body-joints, we firstly evaluate the impact of the joint subsets we select for generating DCK, as not all body-joints need to be used for capturing actions.

\begin{table}[t]%htbp % left bottom right top
%\renewcommand{\arraystretch}{0.85}
%\footnotesize
\parbox{.99\linewidth}{
\centering	
\begin{tabular}{ c | c | c | c | c }
 & SCK & \multicolumn{2}{|c|}{DCK} & SCK+DCK\\
\hline
accuracy & $92.98\%$ & $93.03\%$ & $92.77\%$ & $\mathbf{95.23\%}$\\%95.47
size & 26,565 & 9,450 & 16,920 & 43,485\\
\hline
\end{tabular}\vspace{0.2cm}\\}
\parbox{.99\linewidth}{
\centering	
\begin{tabular}{ c | c | c | c }
 & SCK$\,\oplus$ & DCK$\,\oplus$ & SCK$\,\oplus$ + DCK$\,\oplus$\\
\hline
accuracy & $96.50\%$ & $96.41\%$ & $\mathbf{97.45\%}$\\%95.47
size & 60,900 & 37,800 & 98,700\\
\hline
\end{tabular}\vspace{0.2cm}\\}
\parbox{.99\linewidth}{
\centering
\begin{tabular}{ c | c}
\hline
Bag-of-Poses $82.00\%$ \cite{seidenari_florence3d} & Kendal Traj. $93.04\%$ \cite{kendal_traj}\\
$SE(3)$ $90.88\%$ \cite{vemulapalli_SE3} & Kernel+ResNet \cite{action_da} $95.4\%$\\
\hline
\end{tabular}%\label{tab:flor}
}
%\vspace{-0.4cm}
\caption{Evaluations of ({\em top}) SCK/DCK, ({\em middle}) our improved SCK$\,\oplus$ / DCK$\,\oplus$, ({\em bottom}) the state of the art on Florence3D-Action.}%\vspace{-0.2cm}
\label{tab:flor}
\end{table}

\begin{table}[t]%htbp % left bottom right top
%\renewcommand{\arraystretch}{0.85}
%\footnotesize
\parbox{.99\linewidth}{
\centering
\begin{tabular}{ c | c | c | c }
 & SCK & DCK & SCK+DCK\\
\hline
accuracy & $96.08\%$ & $97.5\%$ & $\mathbf{98.2\%}$\\%97.69 %98.39
size & 40,480 & 16,920 & 57,400\\
\hline
\end{tabular}\vspace{0.2cm}\\}
\parbox{.99\linewidth}{
\centering
\begin{tabular}{ c | c | c | c }
 & SCK$\,\oplus$ & DCK$\,\oplus$ & SCK$\,\oplus$ + DCK$\,\oplus$\\
\hline
accuracy & $98.50\%$ & $98.12\%$ & $\mathbf{99.2\%}$\\%97.69 %98.39
size & 81,200 & 67,680 & 148,880\\
\hline
\end{tabular}\vspace{0.2cm}\\}
\parbox{.99\linewidth}{
\centering
\begin{tabular}{ c | c }
\hline
3D joints. hist. $90.92\%$ \cite{xia_utkinect} & Kendal Traj. $97.39\%$ \cite{kendal_traj}\\
$SE(3)$ $97.08\%$ \cite{vemulapalli_SE3} & Second-order DA \cite{action_da} $98.9\%$\\
\hline
\end{tabular}
}
%\vspace{-0.4cm}
\caption{Evaluations of ({\em top}) SCK/DCK, ({\em middle}) our improved SCK$\,\oplus$ / DCK$\,\oplus$ and ({\em bottom}) the state of the art on UTKinect-Action.}\vspace{-0.2cm}
\label{tab:utk}
\end{table}

Figure \ref{fig:piv4} enumerates all body-joints that describe every 3D human skeleton on the Florence3D-Action dataset, whereas the table underneath lists the proposed body-joint subsets A-I which we use for computations of DCK. In Figure \ref{fig:piv5}, we plot the performance of our DCK kernel for each subset. The plot shows that using two body-joints associated with the hands from Configuration-A in the DCK kernel construction, we attain $88.32\%$ accuracy which highlights the informativeness of temporal dynamics. %For Configuration-D, which includes six body-joints such as the knees, elbows and hands, our performance reaches $93.03\%$. This suggests that some body-joints that were not selected for this configuration may be noisy and therefore detrimental to classification.

\vspace{-0.05cm}
\begin{tcolorbox}[width=1.0\linewidth, colframe=blackish, colback=beaublue, boxsep=0mm, arc=3mm, left=1mm, right=1mm, right=1mm, top=1mm, bottom=1mm]
Some body-joints may be noisy and thus detrimental to recognition, and should not be selected for experiments \eg,
Configuration-D, which includes six body-joints such as the knees, elbows and hands, yields $93.03\%$, which outperforms more complex configurations.
\end{tcolorbox}
\vspace{-0.2cm}

As  Configuration-E includes eight body-joints such as the feet, knees, elbows and hands, we choose it for our further experiments as it represents a reasonable trade-off between performance and size of representations. This configuration scores $92.77\%$ accuracy. We see that if we utilize all the body-joints according to Configuration-I, performance of $91.65\%$ accuracy is still somewhat lower compared to $93.03\%$ accuracy for Configuration-D highlighting the issue of noisy body-joints. %To the best of our knowledge, this is the highest performance attained on this dataset.

In Figure \ref{fig:piv6}, we show the accuracy on our DCK kernel when HOSVD factors underlying our non-symmetric tensors are equalized by varying the EPN parameter $\gamma$. For $\gamma\!=\!1$,  EPN is disabled, which leads to $90.49\%$ accuracy only. For the optimal value of $\gamma\!=\!0.85$, the accuracy rises to $92.77\%$ which indicates the presence of the burstiness effect in temporal representations.

\subsection{SCK and DCK vs. the state of the art.}

Below, we compare the performance of our representations against the state of the art. \revised{Along with comparing SCK and DCK, we  also explore the complementarity of these representations %in capturing the action dynamics
 by combining them via the so-called late fusion, that is, a simple weighted concatenation of vectorized SCK and DCK.} % demonstrate our best results when combining our two tensor representations on the three datasets.

\begin{table}[t]%[htbp]
%%%%\vspace{-0.2cm}
%\renewcommand{\arraystretch}{0.95}
{
%\footnotesize
\parbox{.99\linewidth}{
\centering	
\begin{tabular}{ c | c | c | c }
 & SCK & DCK & SCK+DCK\\
\hline
\kern-0.1em acc., prot.~\cite{wu_actionlets} & $90.72\%$ & $86.30\%$ & $\mathbf{91.45\%}$\\
\kern-0.1em acc., prot.~\cite{li_msraction3d} & $93.52\%$ & $91.71\%$ & $\mathbf{93.96\%}$\\
size & 40,480 & 16,920 & 57,400\\
\hline
\end{tabular}
\vspace{0.2cm}\\
}
\parbox{.99\linewidth}{
\centering	
\begin{tabular}{ c | c | c | c }
 & SCK$\,\oplus$ & DCK$\,\oplus$ & SCK$\,\oplus$ + DCK$\,\oplus$\\
\hline
\kern-0.1em acc., prot.~\cite{wu_actionlets} & $97.50\%$ & $90.03\%$ & $\mathbf{98.10\%}$\\
\kern-0.1em acc., prot.~\cite{li_msraction3d} & $98.12\%$ & $94.28\%$ & $\mathbf{98.62\%}$\\
size & 81,200 & 67,680 & 148,880\\
\hline
\end{tabular}
\vspace{0.2cm}\\
}
\parbox{.99\linewidth}{
\centering
%
%\begin{tabular}{ c | c | c | c}
%\hline
%acc., protocol~\cite{wu_actionlets} & Actionlets $88.20\%$ \cite{wu_actionlets} & \multirow{2}{*}{$SE(3)$} {$89.48\%$}\multirow{2}{*}{ \cite{vemulapalli_SE3}} & Kin. desc. 91.07\%\cite{zanfir_movingpose}\\
%acc., protocol~\cite{li_msraction3d} & Random Forests $90.90\%$ \cite{zhu_fusingjoints} & $\qquad\!92.46\%$ & - \\
%\hline
%\end{tabular}
%
\begin{tabular}{ c | c }
accuracy, protocol~\cite{wu_actionlets} & accuracy, protocol~\cite{li_msraction3d}\\
\hline
Actionlets $88.20\%$ \cite{wu_actionlets} & $SE(3)$ $92.46\%$ \cite{vemulapalli_SE3}\kern-0.1em\\
$SE(3)$ $89.48\%$ \cite{vemulapalli_SE3} & Kendal Traj. $94.19\%$ \cite{kendal_traj}\\
Kin. desc. $91.07\%$ \cite{zanfir_movingpose} & Ker-RP-RBF $96.9\%$ \cite{lei_kermats} \\
\hline
\end{tabular}
}
}
%\vspace{-0.4cm}
\caption{Results of ({\em top}) SCK/DCK, ({\em middle}) our improved SCK$\,\oplus$ / DCK$\,\oplus$ and ({\em bottom}) the state of the art on MSR-Action3D.}\label{tab:msr}
\vspace{-0.2cm}
\end{table}

\begin{table}[t]%[htbp]
{
%\footnotesize
\parbox{.99\linewidth}{
\centering	
\begin{tabular}{rl | c | c }
																			   \kern-0.8em &\kern-0.5em								                                                        													& \kern-0.5em cross-subject \kern-0.5em & \kern-0.5em cross-view \kern-0.5em\\
\hline
\kern-0.1em \revised{SCK ($r\!=\!2$)}              \kern-0.8em & \kern-0.5em\multirow{4}{*}{\MyLRBrace{5.1ex}{\!}on 3D body-joints}\kern-0.1em															& \revised{$64.08\%$} & \revised{$65.24\%$} \\
\kern-0.1em \revised{SCK (no EPN)} 	               \kern-0.8em & \kern-0.5em																																																& \revised{$65.37\%$} & \revised{$67.18\%$} \\
\kern-0.1em SCK 					               \kern-0.8em & \kern-0.5em																																																& $69.20\%$ & $70.55\%$ \\
\kern-0.1em SCK$\,\oplus$                \kern-0.8em & \kern-0.5em															                                        													& $\mathbf{72.82}\%$ & $\mathbf{74.10}\%$ \\
\hline
\kern-0.1em \revised{SCK}                          \kern-0.8em & \kern-0.5em\revised{\multirow{2}{*}{\MyLLBrace{2.5ex}{\!}\pbox{3cm}{on 3D body-joints\\ +ST-GCN}}}\kern-0.1em												& \revised{$82.61\%$} & \revised{$89.52\%$} \\
\kern-0.1em \revised{SCK$\,\oplus$}                \kern-0.8em & \kern-0.5em															                                        													& \revised{$\mathbf{83.58}\%$} & \revised{$\mathbf{90.84}\%$} \\
\hline
\end{tabular}
\vspace{0.2cm}\\
}
\parbox{.99\linewidth}{
\centering	
\begin{tabular}{rl | c | c }
																			   \kern-0.8em &\kern-0.5em								                                                        													& \kern-0.5em cross-subject \kern-0.5em & \kern-0.5em cross-view \kern-0.5em\\
\hline
\multicolumn{2}{c|}{\kern-0.7em\revised{Two-stream+AP (ResNet-50)} } & \revised{$74.4\%$} & \revised{$83.3\%$}\\
\multicolumn{2}{c|}{\kern-0.7em\revised{Two-stream+MP (ResNet-50)} } & \revised{$65.8\%$} & \revised{$58.7\%$}\\
\hline
\multicolumn{2}{c|}{\kern-0.7em SCK$\,\oplus$                on RGB frames \revised{(ResNet-152)}\kern-0.6em}																															& $\mathbf{90.11}\%$ & $\revised{\mathbf{93.62}\%}$\\
\multicolumn{2}{c|}{\kern-0.6em\multirow{2}{*}{SCK$\,\oplus$} \multirow{2}{*}{\MyLLBrace{2.5ex}{\!}\pbox{4cm}{on 3D body-joints\\+RGB frames \revised{(ResNet-152)}}} \kern-0.6em}		& \multirow{2}{*}{$\mathbf{90.78}\%$} & \multirow{2}{*}{$\revised{\mathbf{94.15}\%}$}\\
																			   \kern-0.5em & \kern-0.5em										& & \\
\multicolumn{2}{c|}{\kern-0.6em\multirow{3}{*}{SCK$\,\oplus$} \multirow{3}{*}{\MyLLBrace{3.8ex}{\!}\pbox{4cm}{on 3D body-joints\\+RGB frames+optical flow\\\revised{(ResNet-152)}}} \kern-0.6em}		& \multirow{3}{*}{$\mathbf{91.56}\%$} & \multirow{3}{*}{$\revised{\mathbf{94.75}\%}$}\\
																			   \kern-0.5em & \kern-0.5em										& & \\
																			   \kern-0.5em & \kern-0.5em										& & \\
%size & 40,480 & 16,920 & 57,400\\
\hline
\end{tabular}
\vspace{0.2cm}\\
}
\parbox{.99\linewidth}{
\centering
\begin{tabular}{ l | c | c }
		& \kern-0.5em	cross-subject\kern-0.5em	 & \kern-0.5em	cross-view \kern-0.5em	\\
\hline
\kern-0.5em Second-order DA \cite{action_da} \revised{(ResNet-50)} & $75.35\%$ & $79.30\%$\\
\kern-0.5em Frames + CNN  \cite{quinhong} \revised{(VGG-19)} & $75.73\%$ & $79.62\%$\kern-0.5em\\
\kern-0.5em Clips + CNN + MTLN \cite{quinhong} \revised{(VGG-19)} & $79.57\%$ & $84.83\%$\kern-0.5em\\
\kern-0.5em VA-LSTM \cite{va_lstm} & $79.4\%$ & $87.6\%$\\
\kern-0.5em \revised{ST-GCN} \cite{stgcn} & \revised{$81.5\%$} & \revised{$88.3\%$}\\
\kern-0.5em DSP  \cite{action_advers} & $81.6\%$ & $88.7\%$\\
\kern-0.5em \revised{Multi-scale CNN} \cite{yuchao_ar} \revised{(ResNet-101)} & \revised{$84.6\%$} & \revised{$92.1\%$}\\
\kern-0.5em \revised{Multi-scale CNN} \cite{yuchao_ar} \revised{(ResNet-152)} & \revised{$85.0\%$} & \revised{$92.3\%$}\\
\kern-0.5em \revised{Deep Bilinear} \cite{act_deep_bilinear} \revised{(ResNet-101)} & \revised{$85.4\%$} & \revised{$90.7\%$}\\
\hline
\end{tabular}
}
}
%\vspace{-0.4cm}
\caption{\revised{Results on our SCK and the improved SCK$\,\oplus$ on ({\em top}) skeleton sequences and ({\em middle}) two-stream networks. We also indicate  results on the baseline two-stream network with standard average pooling ({\em AP}) and maximum pooling ({\em MP}). We indicate backbones in parentheses. ({\em bottom}) The state of the art on NTU-RGBD.}}\label{tab:ntu}
\vspace{-0.2cm}
\end{table}

\begin{table}[t]%htbp % left bottom right top
%\renewcommand{\arraystretch}{0.85}
%\footnotesize
\parbox{.99\linewidth}{
\centering
\begin{tabular}{ c | c | c | c}
 & \revised{SCK+ST-GCN} & \revised{SCK$\,\oplus$+ST-GCN} & \revised{ST-GCN} \\
\hline
\revised{top-1} & \revised{$31.2\%$} & \revised{$\mathbf{31.8}\%$} & \revised{$30.7\%$}\\
\revised{top-5} & \revised{$53.7\%$} & \revised{$\mathbf{54.9}\%$} & \revised{$52.8\%$}\\
\hline
\end{tabular}
}
%
%\vspace{-0.2cm}
%\vspace{-0.4cm}
\caption{\revised{SCK and SCK$\,\oplus$ combined with ST-GCN \vs ST-GCN \cite{stgcn} alone on Kinetics \cite{kinetics_400} skeletons extracted by OpenPose \cite{cao2018openpose}.}}\vspace{-0.2cm}
\label{tab:kinect}
\end{table}

On the Florence3D-Action dataset, we present our best results in Table \ref{tab:flor}. Note that the model parameters for the evaluation was selected by cross-validation. Linearizing a sequence compatibility kernel using these parameters resulted in a tensor representation of size $26,565$ dimensions\footnote{\label{foot:foot1}This is the length of a vector per sequence after unfolding our tensor represent./removing duplicate coefficients from the symmetries in the tensor.}, and produced an accuracy of $92.98\%$ accuracy. As for DCK, our model used Configuration-E (described in Figure~\ref{fig:piv4}) resulting in a representation of dimensionality $16,920$, and achieved a performance of $92\%$. However, somewhat better results were attained by Configuration-D, that is, $92.27\%$ accuracy for size of $9,450$.  Combining SCK and DCK in Configuration-E yields $95.23\%$ accuracy, a $4.5\%$ improvement over the state of the art on this dataset, as listed in Table \ref{tab:flor}, which highlights the complementary nature of SCK and DCK. %To the best of our knowledge, this is the highest performance attained on this dataset.

Action recognition results on the UTKinect-Action dataset are presented in Table \ref{tab:utk}. For our experiments on this dataset, we kept all the parameters the same as those used on the Florence3D dataset. SCK and DCK representations yielded  $96.08\%$ and $97.5\%$ accuracy, respectively. Combining SCK and DCK yielded $98.2\%$ accuracy outperforming marginally a more complex approach \cite{vemulapalli_SE3} based on the Lie group algebra, %on $SE(3)$ matrix descriptors 
 dynamic time warping and Fourier temporal pyramids. %, which we avoid completely.

In Table~\ref{tab:msr}, we present our results on the MSR-Action3D dataset. %Again, we kept all the model parameters the same as those used on the Florence3D dataset. 
Conforming to the prior literature, we use two evaluation protocols, that is, (i) the protocol described in actionlets~\cite{wu_actionlets}, for which the authors utilize the entire dataset with its 20 classes during the training and evaluation, and (ii) approach of~\cite{li_msraction3d}, for which the authors divide the data into three subsets and report the average in classification accuracy over these subsets. SCK yields the state-of-the-art accuracy of $90.72\%$ and $93.52\%$ for the two evaluation protocols, respectively. Combining SCK with DCK outperforms other approaches listed in the table and yields $91.45\%$ and $93.96\%$ accuracy for the two protocols, respectively. %\todo{How were the parameters set for evaluations on these datasets?}

\subsection{SCK$\,\oplus$ and DCK$\,\oplus$ vs. the state of the art.}
\label{exp:sck-dck-info}
Our extended SCK$\,\oplus$ is trained with $3Z_2\!=\!15$, $Z_3\!=Z_4\!=\!5$ and $Z_5\!=\!3$ while DCK$\,\oplus$ follows the same setting as DCK, except that we introduce quantity $Z_6\!=\!4$ which is the number of pivots encoding the subsequence position within the sequence, as dictated by Eq. \eqref{eq:ker_dyn_pp}. For the Florence3D-Action dataset, Table \ref{tab:flor} shows that aggregating over subsequences across various scales results in SCK$\,\oplus$ outperforming SCK by $\sim$3.5\%, DCK$\,\oplus$ outperforming DCK by $\sim$3.4\% and the combined kernel SCK$\,\oplus$ + DCK$\,\oplus$ outperforming SCK+DCK by $\sim$2.2\%. Table \ref{tab:utk} shows the similar trend for the UTKinect-Action dataset, for which SCK$\,\oplus$ outperforms SCK by $\sim$2.4\%, DCK$\,\oplus$ outperforms DCK by $\sim$0.6\% and the combined kernel SCK$\,\oplus$ + DCK$\,\oplus$ outperforms SCK+DCK by $\sim$1.0\%. Note that the results on UTKinect-Action should be considered as already saturated. Furthermore, Table \ref{tab:msr} shows that on MSR-Action3D, SCK$\,\oplus$ outperforms SCK by $\sim$6.8\%, DCK$\,\oplus$ outperforms DCK by $\sim$3.7\% and the combined kernel SCK$\,\oplus$ + DCK$\,\oplus$ outperforms SCK+DCK by $\sim$7.5\%.

\vspace{0.05cm}
\noindent{\textbf{Fine-grained Action Recognition. }} In what follows, we employ NTU-RGBD, a partially fine-grained dataset, and MPII Cooking Activities containing many fine-grained classes.  

\vspace{-0.0cm}
\begin{tcolorbox}[width=1.0\linewidth, colframe=blackish, colback=beaublue, boxsep=0mm, arc=3mm, left=1mm, right=1mm, right=1mm, top=1mm, bottom=1mm]
Our SCK$\,\oplus$ kernel is designed to capture specific subsequences of variable lengths. Kernels $G_{\sigma_2},\cdots,G_{\sigma_5}$ from Section \ref{sec:ker3} capture higher-order statistics of joint locations in subsequences, the temporal
alignment of pose snippets, the global alignment of subsequences, and the match of subsequence lengths.  SCK$\,\oplus$ uses EPN in Eq. (\ref{eq:rawcod3}-\ref{eq:rawcod6}) which makes %SCK$\,\oplus$ 
it act as a detector of spectral higher-order occurrences. Thus, SCK$\,\oplus$ addresses all hallmarks of modern fine-grained recognition systems: {\em it captures higher-order statistics} describing visual contents/objects and {\em discarding burstiness} \cite{koniusz2018deeper} (co-occurrence detection).
\end{tcolorbox}
\vspace{-0.2cm}

Moreover, our SCK$\,\oplus$ kernel  captures higher-order occurrences of features representing spatio-temporal evolution of skeletons (for 3D body-joints) and/or frame-based classifier scores (semantic information) by feeding them into kernels $G_{\sigma_2^{(1)}},\cdots,G_{\sigma_2^{(Q)}}$ from Eq. \eqref{eq:ker3a} for $Q$ modalities.

Table \ref{tab:ntu} (top) shows that, our SCK$\,\oplus$ yields some $\sim$3.6\% improvement over SCK and reaches $72.82\%$ accuracy on the NTU-RGBD dataset in the cross-
subject setting for the 3D body-joints as input. We expect that aggregating over subsequences can encode local fine-grained motion details essential for the good performance. Similar observations hold for the cross-view setting.

\begin{table}[t]%htbp % left bottom right top
%\renewcommand{\arraystretch}{0.85}
%\footnotesize
\parbox{.99\linewidth}{
\centering
\begin{tabular}{ l | c | c | c | c  }
 & \multirow{2}{*}{-} & \multirow{2}{*}{\kern-0.5em+IDT\kern-0.5em} & \multirow{2}{*}{\kern-0.6em+sec-ord\kern-0.6em} & \kern-0.5em+sec-ord\kern-0.5em \\
 & 	&  &  & \kern-0.5em+IDT\kern-0.5em \\
\hline
\kern-0.6em \revised{Two-stream+AP (VGG-19)}\kern-0.6em 							& \kern-0.6em\revised{$38.1\%$}\kern-0.6em	 					& - & - & - \\
\kern-0.6em \revised{Two-stream+AP (ResNet-152)}\kern-0.6em 					& \kern-0.6em\revised{$45.3\%$}\kern-0.6em					 & - & - & - \\
\kern-0.6em \revised{Subsequences+AP (I3D)}\kern-0.6em 								& \kern-0.6em\revised{$52.7\%$}\kern-0.6em					 & - & - & - \\
\hline
\kern-0.6em HOK \cite{hok} \revised{(VGG-16)} \kern-0.6em    		   	  & \kern-0.6em$60.1\%$\kern-0.6em 					 & - & $69.1\%$ & \kern-0.5em$73.1\%$\kern-0.5em \\
\kern-0.6em SCK$\,\oplus$ \revised{(VGG-19)}\kern-0.6em 							& \kern-0.6em$70.1\%$\kern-0.6em 					 & - & $74.0\%$ & \kern-0.5em$76.1\%$\kern-0.5em \\
\kern-0.6em SCK$\,\oplus$ \revised{(ResNet-152)}\kern-0.6em 					& \kern-0.6em$71.4\%$\kern-0.6em 					 & - & $75.5\%$ & \kern-0.5em$77.4\%$\kern-0.5em \\
\kern-0.6em \revised{SCK$\,\oplus$ \revised{(I3D)}}\kern-0.6em 			  & \kern-0.6em\revised{$77.8\%$}\kern-0.6em & \kern-0.5em\revised{$\mathbf{80.4}\%$}\kern-0.5em & - & -  \\
\hline
\end{tabular}
\vspace{0.2cm}\\
}
\parbox{.99\linewidth}{
\centering
\begin{tabular}{ c | c }
\hline
\kern-0.6em KRP-FS $70.0\%$ \cite{anoop_subsp_repr} \revised{(VGG-19)}\kern-0.5em & \kern-0.5emKRP-FS+IDT $76.1\%$ \cite{anoop_subsp_repr} \revised{(VGG-19)}\kern-0.9em\\
\kern-0.6em GRP $68.4\%$ \cite{anoop_generalized} \revised{(VGG-19)}\kern-0.5em & \kern-0.5emGRP+IDT $75.5\%$ \cite{anoop_generalized} \revised{(VGG-19)} \kern-0.9em\\
\hline
\end{tabular}
}
\caption{\revised{Results (mAP\%) for ({\em top}) our HOK \cite{hok} and improved SCK$\,\oplus$. We also indicate  results on the baseline two-stream network with standard average pooling ({\em AP}). We indicate backbones in parentheses. ({\em bottom}) The state of the art on MPII Cooking Activities.}}
\vspace{-0.2cm}
\label{tab:mpii}
\end{table}

\begin{table}[t]%htbp % left bottom right top
%\renewcommand{\arraystretch}{0.85}
%\footnotesize
\parbox{.99\linewidth}{
\centering
\begin{tabular}{ l| c | c | c | c }
 & \kern-0.5em{\em sp1}\kern-0.5em & \kern-0.5em{\em sp2}\kern-0.5em & \kern-0.5em{\em sp3}\kern-0.5em & \kern-0.5emmean acc.\kern-0.5em \\
\hline
\kern-0.5em \revised{Two-stream+AP (ResNet-152)}\kern-0.5em     & \kern-0.5em$65.30\%$\kern-0.5em & \kern-0.5em$62.20\%$\kern-0.5em & \kern-0.5em$62.55\%$\kern-0.5em & \kern-0.5em$63.35\%$\kern-0.5em\\
\kern-0.5em \revised{Two-stream+MP (ResNet-152)}\kern-0.5em     & \kern-0.5em$61.38\%$\kern-0.5em & \kern-0.5em$60.58\%$\kern-0.5em & \kern-0.5em$60.06\%$\kern-0.5em & \kern-0.5em$60.66\%$\kern-0.5em\\
\hline
\kern-0.5emSCK$\,\oplus\,$\revised{(ResNet-152)}\kern-0.5em     & \kern-0.5em$72.55\%$\kern-0.5em & \kern-0.5em$70.85\%$\kern-0.5em & \kern-0.5em$71.63\%$\kern-0.5em & \kern-0.5em$71.67\%$\kern-0.5em\\
\kern-0.5emSCK$\,\oplus\,$\revised{(ResNet-152)}+IDT\kern-0.5em & \kern-0.5em$74.20\%$\kern-0.5em & \kern-0.5em$73.73\%$\kern-0.5em & \kern-0.5em$73.40\%$\kern-0.5em & \kern-0.5em$73.77\%$\kern-0.5em\\
\kern-0.5em\revised{SCK$\,\oplus ($r=2$)\,$(I3D)+IDT}\kern-0.5em & \kern-0.5em\revised{$85.61\%$}\kern-0.5em & \kern-0.5em\revised{$84.54\%$}\kern-0.5em & \kern-0.5em\revised{$85.25\%$}\kern-0.5em & \kern-0.5em\revised{$85.13\%$}\kern-0.5em\\
\kern-0.5em\revised{SCK$\,\oplus\,$(I3D)+IDT}\kern-0.5em & \kern-0.5em\revised{$86.31\%$}\kern-0.5em & \kern-0.5em\revised{$85.63\%$}\kern-0.5em & \kern-0.5em\revised{$86.41\%$}\kern-0.5em & \kern-0.5em\revised{$\mathbf{86.11}\%$}\kern-0.5em\\
\hline
\end{tabular}
\vspace{0.2cm}\\
}
\parbox{.99\linewidth}{
\centering
\begin{tabular}{ c | c }
\hline
\kern-0.5em DSP $72.4\%$ \cite{action_advers} \revised{(ResNet-152)}\kern-0.5em  & \kern-0.5em ShuttleNet+MIF $71.08\%$ \cite{long_term_dep}\kern-0.5em\\
\kern-0.5em DSP+IDT $74.3\%$ \cite{action_advers} \revised{(ResNet-152)}\kern-0.5em & \revised{\kern-0.5em I3D $80.2\%$} \cite{i3d_net}\kern-0.5em\\
\hline
\end{tabular}
}
\caption{\revised{Evaluations of ({\em top}) our improved SCK$\,\oplus$. We  also indicate  results on baseline two-stream network with standard average pooling ({\em AP}) and maximum pooling ({\em MP}). We indicate backbones in parentheses. ({\em bottom}) The state of the art on HMDB-51.}}
\vspace{-0.2cm}
\label{tab:hmdb51}
\end{table}

Table \ref{tab:ntu} (middle) shows that our SCK$\,\oplus$ attains $90.11\%$ accuracy on the NTU-RGBD dataset in the cross-
subject setting on the RGB frames (classifier scores) as input. With the 3D body-joints added, results increase to $90.78\%$. Lastly, adding optical flow as input to our SCK$\,\oplus$ yields $91.56\%$. This is $\sim10.0\%$ improvement over competing methods from Table \ref{tab:ntu} (bottom).

Table \ref{tab:mpii} shows that our SCK$\,\oplus$ yields some 1.4\% mAP improvement over other state-of-the-art methods \cite{anoop_generalized,anoop_subsp_repr} on the MPII Cooking Activities dataset. Further improvements are attained by combining SCK$\,\oplus$ with the second-order descriptor ({\em sec-ord}) \cite{anoop_subsp_repr} and the IDT representation, which yields $77.4\%$ mAP. This compares favorably with other methods in the table. %This is about $\sim\!1.3\%$ improvement over the kernelized subspaces approach ({\em KRP-FS+IDT}) combined with IDT representation. 
We also note that SCK$\,\oplus$ outperforms the HOK descriptor \cite{hok} which is a variant of SCK with a suboptimal linearization of an {\em fc} layer. 
\revised{Finally, applying SCK$\,\oplus$ over I3D-based subsequences yields state-of-the-art $80.4\%$ mAP (we comment on the reasons below).}

\vspace{0.05cm}
\noindent{\textbf{Video Classification. }} Table \ref{tab:ntu} confirms that the classifier scores extracted from CNNs rather than mere 3D body-joints are a more informative input for SCK$\,\oplus$. Thus, we perform additional evaluations on the HMDB-51 dataset. Table \ref{tab:hmdb51} (top) shows that SCK$\,\oplus$ and SCK$\,\oplus$+IDT, trained with the two-stream ResNet-152 backbone, score $71.67$ and $73.77\%$ accuracy which is on a par with other best methods listed in Table \ref{tab:hmdb51} (bottom). \revised{Furthermore, applying the I3D backbone on SCK$\,\oplus$ yields state-of-the-art $86.11\%$ accuracy. We believe that training I3D on subsequences of various lengths and strides, as detailed in Section \ref{sec:setup} (bottom), is a more discriminative strategy than average-pooling of frame-wise features in standard two-stream networks. As SCK$\,\oplus$ is designed to combine subsequences of various lengths and strides rather than sequences, it captures informative higher-order occurrences of multiple complementary features, and also preserves a degree of individual statistics by factoring out one variable at a time \eg, see the discussion in Figure \ref{fig:phis1}.}

\revised{
\vspace{0.05cm} 
\noindent{\textbf{Kinetics-400. }} Table \ref{tab:kinect} shows that our SCK and SCK$\,\oplus$ are complementary to powerful networks such as ST-GCN \cite{stgcn}. We work with Kinetics skeletons extracted with \cite{cao2018openpose} and compare our method to the baseline ST-GCN \cite{stgcn}. We use the standard training/evaluation protocol (but we use skeletons rather than RGB or optical flow frames). As SCK and SCK$\,\oplus$ are shallow representations based on higher-order aggregation, it is unrealistic to expect them to outperform deep networks. However, SCK and SCK$\,\oplus$ capture very different statistics compared to deep networks, being highly complementary. Table \ref{tab:kinect} shows that we attain $1.1\%$ and $2.1\%$ gain over ST-GCN alone by concatenating both representations.
}

\revised{
\vspace{0.05cm}
\noindent{\textbf{Signature Lengths.}} 
Section \ref{exp:sck-dck-info} indicates the number of pivots for SCK$\,\oplus$ on NTU (skeleton-based experiments) to amount to $d\!=\!3Z_2\!+\!Z_3\!+\!Z_4\!+\!Z_5\!=\!15\!+\!5\!+\!5\!+\!3\!=\!28$. The unique number of coefficients in the super-symmetric tensor of order $r$ follows the formula $\binom{d+r-1}{r}$ discussed just below Eq. \eqref{eq:ker1c}. As we obtain a tensor per joint, and we concatenate unique parts of tensors $j\!=\!1,\cdots,J$, we have $\binom{d+r-1}{r}\!\cdot\!J$ coefficients in total in our representation. For SCK$\,\oplus$ on NTU with $J\!=\!25$ body joints, we obtain $4060\!\times\!25\!=\!101500$ coefficients for SCK$\,\oplus$. For SCK and SCK ($r\!=\!2$) on NTU, we set $d\!=\!3Z_2\!+\!Z_3\!=24\!+\!5\!=\!29$ and obtain $112375$ and $10875$ coefficients, respectively. For Kinetics skeletons with $J\!=\!18$ body joints, OpenPose returns only two Cartesian coordinates, so we set $d\!=\!2Z_2\!+\!Z_3\!+\!Z_4\!+\!Z_5\!=\!20\!+\!5\!+\!5\!+\!3\!=\!33$ which yields $4545\!\times\!18\!=\!117810$ coefficients.

For SCK$\,\oplus$ (NTU) on (i) RGB frames and (ii) RGB frames+optical flow, we obtain $d\!=\!Z_2\!+\!Z_3\!+\!Z_4\!+\!Z_5\!=\!60\!+\!5\!+\!5\!+\!3\!=\!73$ and $d\!=\!2Z_2\!+\!Z_3\!+\!Z_4\!+\!Z_5\!=\!73$ (for the latter case, we reduce the size of vector of classifier scores $2\!\times$ by the PCA). As we do not use any body joints here, we obtain $67525$ coefficients. When we concatenate these representations with the skeleton-based one, we obtain $67525\!+\!101500\!=\!169025$ coefficients per video.

On SCK$\,\oplus$ given MPII and HMDB-51 datasets, we obtain $171700$ and $125580$ coefficients after reducing  the size of vectors of RGB frame-wise and optical flow classifier scores from $2\!\times\!64$ to $100$ and from $2\!\times\!51$ to $90$, respectively.
}

\revised{
\vspace{0.05cm}
\noindent{\textbf{Parameters in SCK$\,\oplus$.}} 
The main parameters shared between SCK and SCK$\,\oplus$ are evaluated in Figures \ref{fig:par1} and \ref{fig:par2}. The parameters for SCK$\,\oplus$ that we start with are indicated in Section \ref{sec:ker3} (bottom). To select the best parameters, we cross-validate one parameter at a time while keeping the rest fixed. For NTU, we aggregated over subsequence lengths (using the Matlab notation) of  $14\!:\!1\!:\!110$, $14\!:\!2\!:\!110$, $14\!:\!4\!:\!110$ and $14\!:\!6\!:\!110$, and we obtained $73.10\%$, $72.82\%$, $72.41\%$ and $71.65\%$ accuracy, respectively. For subsequence lengths $30\!:\!2\!:\!110$ and $30\!:\!2\!:\!80$, we obtained $72.54\%$ and $72.12\%$ accuracy. These evaluations show that SCK$\,\oplus$ is not overly sensitive to its parameters. For smaller skeleton-based datasets, we aggregate subsequences in range $6\!:\!2\!:\!24$, whereas on HMDB-51 we use $6\!:\!8\!:\!62$, and for MPII we use $48\!:\!16\!:\!96$.}

\vspace{0.05cm}
\noindent{\textbf{Processing Time.}}
For SCK/DCK, processing a sequence with unoptimized MATLAB code on a single i5 core takes 0.2s and 1.2s, respectively. For SCK$\,\oplus$ / DCK$\,\oplus$, processing one sequence takes 0.5s and 3.0s. Training on full MSR-Action3D with the SCK+DCK takes about 13 min, whereas with the SCK$\,\oplus$ + DCK$\,\oplus$, it takes about 35 min. In comparison, extracting $SE(3)$ features \cite{vemulapalli_SE3} takes 5.3s per sequence, processing on the full MSR-Action3D dataset takes $\sim$50 min., whereas with post-processing (time warping and Fourier pyramids) it takes about 72 min. Thus, SCK+DCK is $\sim\!5.4\!\times$ faster while SCK$\,\oplus$ + DCK$\,\oplus$ is $\sim\!2\!\times$ faster. 
% When $SE(3)$ features are combined with Discrete Time Warping and Fourier pyramids, processing extends to 72mins.
Section \hyperref[{sec:complexity}]{C} contains the computational complexity analysis.
\begin{figure*}[t]
% ensure that we have normalsize text
%\normalsize
\fontsize{8}{9}\selectfont
% Store the current equation number.
%\setcounter{MYtempeqncnt}{\value{equation}}
% Set the equation number to one less than the one
% desired for the first equation here.
% The value here will have to changed if equations
% are added or removed prior to the place these
% equations are referenced in the main text.
%\setcounter{equation}{19}
%
%

\vspace{-0.15cm}
\hrulefill
\vspace{-0.3cm}

\begin{align}
& \!K_D(\piA,\piB)=%\nonumber\\
%
%&\!\!=
\!\frac{1}{\Lambda}\!\!\!\!\!\sum\limits_{\substack{(i,s)\in\idxJ\!,\\(i',s')\in\idxJ\!,\\i'\!\!\neq\!i\!,s'\!\!\neq\!s}}\sum\limits_{\substack{(\!j,t)\in\idxJ\!,\\(\!j'\!\!,t'\!)\in\idxJ,\\j'\!\!\neq\!j\!,t'\!\!\neq\!t}}\!\!\!\!G'_{\sigma'_1}(i\!-\!j\!, i'\!\!-\!j'\!)\,G_{\sigma'_2}\left(\left(\vx_{is}\!-\!\vx_{i's'}\!\right)\!-\!\left(\vy_{jt}-\vy_{j't'}\right)\right)G'_{\sigma'_3}(\frac{s-t}{N},\!\frac{s'-t'}{N})\cdot%\nonumber\\[-24pt]
%&\qquad\qquad\qquad\qquad\qquad\qquad\qquad\qquad\qquad\qquad\qquad\qquad\qquad\;\;\;\cdot 
G'_{\sigma'_4}(s\!-\!s'\!,t\!-\!t'\!)\nonumber\\%[2pt]
&\qquad\qquad\qquad\,=\!\frac{1}{\Lambda}\!\!\sum\limits_{\substack{i,i'\!\in\idx{J}\!:\\i'\!\neq i}}\sum\limits_{\substack{s,s'\!\in\idx{N}\!:\\s'\!\!\neq\!s}}\sum\limits_{\substack{t,t'\!:\\t'\!\!\neq\!t}}\!G_{\sigma'_2}\big(\!\left(\vx_{is}\!-\!\vx_{i's'}\!\right)\!-\!\left(\vy_{jt}-\vy_{j't'}\right)\!\big)\,G_{\sigma'_3}\big(\frac{s-t}{N}\big)G_{\sigma'_3}\big(\frac{s'-t'}{N}\big)
%\cdot{\RaiseBiggBar{-8pt}{_{\substack{\\[-10pt]j\!=\!i\\j'\!\!=\!i'\!}}}}%\nonumber\\[-24pt]
%&\qquad\qquad\qquad\qquad\qquad\qquad\qquad\qquad\qquad\qquad\quad\;
\cdot G_{\sigma'_4}(s\!-\!s'\!)\,G_{\sigma'_4}(t\!-\!t'\!) {\RaiseBiggBar{-2pt}{_{\substack{\\[-10pt]j\!=\!i\\j'\!\!=\!i'\!}}}}\nonumber\\%[2pt]
&\qquad\qquad\qquad\,\approx\!\frac{1}{\Lambda}\!\!\sum\limits_{\substack{i,i'\!\in\idx{J}\!:\\i'\!\neq i}}\sum\limits_{\substack{s,s'\!\in\idx{N}\!:\\s'\!\!\neq\!s}}\sum\limits_{\substack{t,t'\!\in\idx{N}\!:\\t'\!\!\neq\!t}}\!\vphi\left(\vx_{is}\!-\!\vx_{i's'}\!\right)^T\!\vphi\left(\vy_{it}-\vy_{i't'}\right)\!\cdot\!\vz\big(\frac{s}{N}\big)^T\!\vz\big(\frac{t}{N}\big)\!\cdot\!\vz\big(\frac{s'\!}{N}\big)^T\!\vz\big(\frac{t'\!}{N}\big)\cdot%\nonumber\\[-24pt]
%&\qquad\qquad\qquad\qquad\qquad\qquad\qquad\qquad\qquad\qquad\quad\;\cdot 
G_{\sigma'_4}(s\!-\!s'\!)\,G_{\sigma'_4}(t\!-\!t'\!)\nonumber\\%[2pt]
&\qquad\qquad\qquad\,=\!\frac{1}{\Lambda}\!\!\sum\limits_{\substack{i,i'\!\in\idx{J}\!:\\i'\!\neq i}}\sum\limits_{\substack{s,s'\!\in\idx{N}\!:\\s'\!\!\neq\!s}}\sum\limits_{\substack{t,t'\!\in\idx{N}\!:\\t'\!\!\neq\!t}}\!\!\left<\!G_{\sigma'_4}(s\!-\!s'\!)\left(\vphi(\vx_{is}\!\!-\!\!\vx_{i's'})
\!\cdot\!\vz\big(\frac{s}{N}\big)^T\!\right)\!\kronstack\vz\big(\frac{s'\!}{N}\big)\!\right.,%\nonumber\\[-24pt]
%&\qquad\qquad\qquad\qquad\qquad\qquad\qquad\qquad
\left.
G_{\sigma'_4}(t\!-\!t'\!)\Big(\vphi(\vy_{it}\!\!-\!\!\vy_{i't'})
\!\cdot\!\vz\big(\frac{t}{N}\big)^T\!\Big)\!\kronstack\vz\big(\frac{t'\!}{N}\big)\!\right>\nonumber\\
&\qquad\qquad\qquad\,=\!\!\!\!\sum\limits_{\substack{i,i'\!\in\idx{J}\!:\\i'\!\neq i}}
\!\!
\left<
\!\frac{1}{\sqrt{\Lambda}}\!\!\sum\limits_{\substack{s,s'\!\in\idx{N}\!:\\s'\!\!\neq\!s}}\!\!
G_{\sigma'_4}(s\!-\!s'\!)\left(\vphi(\vx_{is}\!\!-\!\!\vx_{i's'})
\!\cdot\!\vz\big(\frac{s}{N}\big)^T\!\right)\!\kronstack\vz\big(\frac{s'\!}{N}\big)\!\right.,%\nonumber\\[-24pt]
%&\qquad\qquad\qquad\qquad\qquad\!
\left.
\!\frac{1}{\sqrt{\Lambda}}\!\!\sum\limits_{\substack{t,t'\!\in\idx{N}\!:\\t'\!\!\neq\!t}}\!\!
G_{\sigma'_4}(t\!-\!t'\!)\,\Big(\vphi(\vy_{it}\!\!-\!\!\vy_{i't'})
\!\cdot\!\vz\big(\frac{t}{N}\big)^T\!\Big)\!\kronstack\vz\big(\frac{t'\!}{N}\big)\!\right>
\label{eq:supp1}
\end{align}

\vspace{-0.15cm}
\hrulefill
\vspace{-0.3cm}

\begin{align}
&\!\!K_D^{*}(\piA,\piB)%\!%\nonumber\\
=%\label{eq:supp2}\\
%&
%&
\!\!\!\!\!\sum\limits_{\substack{i,i'\!\in\idx{J}\!:\\i'\!\neq i}}
\!\!\!
\left<
\!\tG\bigg(\!\frac{1}{\sqrt{\Lambda}}\!\!\!\!\!\!\sum\limits_{\quad\substack{s,s'\!\in\idx{N}\!:\\s'\!\!\neq\!s}\!\!\!\!}\!\!\!\!\!
G_{\sigma'_4}(s\!-\!s'\!)\left(\vphi(\vx_{is}\!\!-\!\!\vx_{i's'})
\!\cdot\!\vz\big(\frac{s}{N}\big)^T\!\right)\!\kronstack\vz\big(\frac{s'\!}{N}\big)\!\right.\!\!\bigg),%\nonumber\\[-6pt]
%&\qquad\qquad\qquad\qquad
%\!\!
\tG\bigg(\!\left.
\!\frac{1}{\sqrt{\Lambda}}\!\!\!\!\!\!\sum\limits_{\quad\substack{t,t'\!\in\idx{N}\!:\\t'\!\!\neq\!t}\!\!\!\!}\!\!\!\!\!
G_{\sigma'_4}(t\!-\!t'\!)\Big(\vphi(\vy_{it}\!\!-\!\!\vy_{i't'})
\!\cdot\!\vz\big(\frac{t}{N}\big)^T\!\Big)\!\kronstack\vz\big(\frac{t'\!}{N}\big)\!\bigg)\!\right>\nonumber\\[-14pt]
&\label{eq:supp2}
\end{align}
\vspace{-0.45cm}
%

%
% Restore the current equation number.
%\setcounter{equation}{20}
% IEEE uses as a separator
\hrulefill
% The spacer can be tweaked to stop underfull vboxes.
\vspace*{4pt}\vspace{-0.45cm}
\end{figure*}
\section{Conclusions}\label{ref:conc}

We have presented two kernel-based tensor representations, % for action recognition from 3D skeletons, 
namely the sequence compatibility kernel (SCK) and dynamics compatibility kernel (DCK). SCK captures the higher-order correlations between 3D coordinates of the body-joints and their temporal variations. As SCK factors out the temporal variable, expensive Fourier temporal pyramid matching/dynamic time warping are not needed. %, commonly used in action recognition. %for generating sequence-level action representations. 
Further, our DCK kernel captures the action dynamics by modeling the spatio-temporal co-occurrences of the body-joints. %This tensor representation also factors out the temporal variable, whose length depends on each sequence. 

Additionally, we have presented a highly effective extension of SCK, termed SCK$\,\oplus$, which aggregates over subsequences of multiple lengths, focusing on actions within subsequences. We have demonstrated that SCK$\,\oplus$ can aggregate over 3D body-joints and/or frame-wise classifier scores from CNNs to capture higher-order statistics between various features extracted from body-skeletons, classifier scores, and temporal positions. % which is the hallmark of modern fine-grained recognition systems.

\vspace{0.00cm}
\begin{tcolorbox}[width=1.0\linewidth, colframe=blackish, colback=beaublue, boxsep=0mm, arc=3mm, left=1mm, right=1mm, right=1mm, top=1mm, bottom=1mm]
Section \hyperref[{sec:epn_interp}]{D} shows that (Tensor) Eigenvalue Power Normalization indeed acts as a spectrum-based metric with $\binom{Z_*}{r}$ subspace-based detectors of higher-order occurrence of datapoints of dim. $Z_*$, more specifically, detectors that capture asymmetry of projections into `positive' and `negative' parts of each subspace. %something long speculated in \cite{me_tensor2,me_tensor} but used at face value in fine-grained systems. 
As $\binom{Z_*}{3}\!\gg\!\binom{Z_*}{2}$, third-order tensors capture more dependencies than autocorrelation matrices, improving fine-grained systems.
\end{tcolorbox}
\vspace{-0.20cm}

%Our experiments on six regular and/or fine-grained datasets for skeleton- and/or video-based inputs substantiate the effectiveness of our representations, demonstrating state-of-the-art performance.% on six challenging action recognition datasets.
%kernel representations demonstrate state of the art performance on  results \eg, on the Florence3D-Action dataset. They are also both novel propositions which, despite somewhat complex maths, boil down to simple sums and outer-products algorithm-wise.
%\renewcommand*\appendixpagename{Appendix}
%\appendix

%\begin{appendices}

\renewcommand{\appendixname}{Remaining Details/Derivations}
\appendix
%\appendixheaderoff

%\renewcommand{\thesection}{A.\arabic{section}}
%\renewcommand{\thesection}{\arabic{section}}{}

%\newcounter{MYtempeqncnt}

%\stepcounter{chapter}
\subsection*{A. Linearizing Dynamics Compatibility Kernel}
\label{sec:dck_der}
In what follows, we derive the linearization of DCK. Let us recall that $G_{\sigma}(\vu-\vub)=\exp(-\enorm{\vu - \vub}^2/{2\sigma^2})$, $G'_{\sigma}(\valpha,\vbeta)=G_{\sigma}(\valpha)G_{\sigma}(\vbeta)$ and $G_{\sigma}(\vi-\vj)=\delta(\vi-\vj)$ if $\sigma\!\rightarrow\!0$, therefore $\delta(\boldsymbol{0})=1$ and $\delta(\vu)=0$ if $\vu\!\neq\!\boldsymbol{0}$. Moreover, $\Lambda=J^2$ is a normalization constant and $\idxJ=\idx{J}\times\idx{N}$. We recall that kernel $G_{\sigma'_2}(\vx-\vy)\approx \phi(\vx)^T\phi(\vy)$ while $G_{\sigma'_3}(\frac{s-t}{N})\approx\vz(s/N)^T\vz(t/N)$. Thus, we obtain 
Eq. \eqref{eq:supp1} which expresses $K_D(\piA,\piB)$ as a sum over dot-products on third-order non-symmetric tensors. We  introduce operator $\tG$ into Eq. \eqref{eq:supp1} to amend the dot-product with a distance which handles burstiness. We obtain a modified kernel in Eq. \eqref{eq:supp2}
%
%
%
%
%From Equation \eqref{eq:supp2} 
based on which the following notation is introduced:
\begin{align}
&\tV_{ii'\!}\!=\!\mygthree{\tX_{ii'\!}}\!\text{, where }\label{eq:supp3}\\
&\tX_{ii'\!}\!=\!\!\frac{1}{\sqrt{\Lambda}}\!\!\sum\limits_{\substack{s,s'\!\in\idx{N}\!:\\s'\!\!\neq\!s}}\!\!
G_{\sigma'_4}(s\!-\!s'\!)\left(
\vphi(\vx_{is}\!\!-\!\!\vx_{i's'})
\!\cdot\!\vz\big(\frac{s}{N}\big)^T\!\right)\!\kronstack\vz\big(\frac{s'\!}{N}\big),\nonumber
\end{align}
and the summation over the pairs of body-joints in Eq. \eqref{eq:supp2} is replaced by the concatenation along the fourth mode to obtain representations $\big[\tV_{ii'\!}\big]_{i>i'\!:\,i,i'\in\idx{J}}^{\oplus_4}$ and $\big[\tVH_{ii'\!}\big]_{i>i'\!:\,i,i'\in\idx{J}}^{\oplus_4}$ for $\piA$ and $\piB$. Thus, $K_D^{*}$ becomes:
\vspace{-0.1cm}
\begin{align}
& \!\!\!\!\!\!\!\!\!\!\!\!K_D^{*}(\piA,\piB)=\left<\sqrt{2}\big[\tV_{ii'\!}\big]_{i>i'\!:\,i,i'\in\idx{J}}^{\oplus_4}\!,\sqrt{2}\big[\tVH_{ii'\!}\big]_{i>i'\!:\,i,i'\in\idx{J}}^{\oplus_4}\right>\!\!\!\!\label{eq:supp4}
\end{align}
As Eq. \eqref{eq:supp4} suggests, we avoid repeating the same evaluations in our representations: we stack only unique pairs of body-joints $i\!>\!i'\!$. Moreover, we ensure we run computations temporally only for $s\!>\!s'\!$. In practice, we have to evaluate only $\binom{JN}{2}$ unique spatio-temporal pairs in Eq. \eqref{eq:supp4} rather than naive  $J^2N^2$ per sequence. The final representation is of $Z'_2\!\cdot\!\binom{JZ'_3\!}{2}$ size, where $Z'_2$ and $Z'_3$ are the numbers of pivots for approximation of $G_{\sigma'_2}\!$ and $G_{\sigma'_3}$. 

We assume that all sequences have $N$ frames for simplification of presentation. Our formulations are equally applicable to sequences of arbitrary lengths \eg,~$M$ and $N$. Thus, we apply in practice $G'_{\sigma'_3}(\frac{s}{M}\!-\!\frac{t}{N},\frac{s'}{M}\!-\!\frac{t'}{N})$ and $\Lambda\!=\!J^2MN$ in Eq. \eqref{eq:supp1}.

\revised{Moreover, a displacement  between any pair of joints $\vx,\vy\!\in\!\mbr{3}$ lies within the Cartesian coordinate system, thus $\vx\!-\!\vy\!\in\!\mbr{3}$. In practice, in place of generic $G_{\sigma'_2}$, we use the sum kernel $G^{'}_{\sigma'_2}(\vx\!-\!\vy)\!=\!G_{\sigma'_2}(x_1\!\!-\!y_1)\!+\!G_{\sigma'_2}(x_2\!\!-\!y_2)\!+\!G_{\sigma'_2}(x_3\!\!-\!y_3)$ so the kernel $G^{'}_{\sigma'_2}(\vx\!-\!\vy)\!\approx\![\phi(x_1\!); \phi(x_2\!); \phi(x_3\!)]^T\![\phi(y_1\!); \phi(y_2\!); \phi(y_3\!)]$. However, for the simplicity of notation, we refer to it in our formulations by its generic form $G_{\sigma'_2}(\vx\!-\!\vy)\!\approx\!\phi(\vx)^T\phi(\vy)$, as we can simply define $\phi(\vx)\!=\![\phi(x_1\!); \phi(x_2\!); \phi(x_3\!)]$.} %Note that $(x)$, $(y)$, $(z)$ are spatial xyz-components of displacement vec. \eg, $\vx_{is}\!-\!\vx_{i's'}$.

\vspace{-0.2cm}
\subsection*{B. Positive Definiteness of SCK and DCK}
SCK/DCK are sums over products of RBF subkernels. According to \cite{taylor_kermet}, sums, products and linear combinations (for non-negative weights) of positive definite kernels yield positive definite kernels. 
Moreover, subkernel $G_{\sigma'_2}\left(\left(\vx_{is}\!-\!\vx_{i's'}\!\right)\!-\!\left(\vy_{jt}-\vy_{j't'}\right)\right)$ employed by DCK in Eq. \eqref{eq:supp1} (top) can be rewritten as:
\begin{align}
&G_{\sigma'_2}\left(\vz_{isi's'}\!-\!\vz'_{jtj't'}\!\right),\label{eq:supp5}\\
&\text{ where }\vz_{isi's'}\!=\!\vx_{is}\!-\!\vx_{i's'}\text{ and }\vz'_{jtj't'}\!=\!\vy_{jt}-\vy_{j't'}.\nonumber
\end{align}

The RBF kernel $G_{\sigma'_2}$ is positive definite (PD) by definition and the mappings from $\vx_{is}$ and $\vx_{i's'}$ to $\vz_{isi's'}$ and from $\vy_{jt}$ and $\vy_{j't'}$ to $\vz'_{jtj't'}$, respectively, are unique. Thus, the entire kernel is PD.

Whitening on SCK results in a positive (semi)definite (PSD) kernel as we employ the Power-Euclidean kernel \eg, if $\mX$ is PD then $\mX^\gamma$ stays also PD for $0\!<\!\gamma\!\leq\!1$ because $\mX^\gamma\!=\!\mU\mLambda^\gamma\mV$ and element-wise rising of eigenvalues to the power of $\gamma$ gives us $\diag(\mLambda)^\gamma\!\geq\!0$. Thus, the sum over dot-products of positive (semi)definite matrices raised to the power of $\gamma$ stays PSD/PD.%is positive (semi)definite.

\ifdefined\arxiv
\newcommand{\PowH}{3.0cm}
\newcommand{\PowHB}{2.875cm}
\newcommand{\PowW}{3.65cm}
\else
\newcommand{\PowH}{3.4cm}
\newcommand{\PowHB}{3cm}
\newcommand{\PowW}{4.1cm}
\fi
\begin{figure*}[t]%htbp % left bottom right top
\centering
\vspace{-0.3cm}
\hspace{-0.3cm}
\begin{subfigure}[t]{0.245\linewidth}
\centering\includegraphics[trim=0 0 0 0, clip=true, height=\PowH]{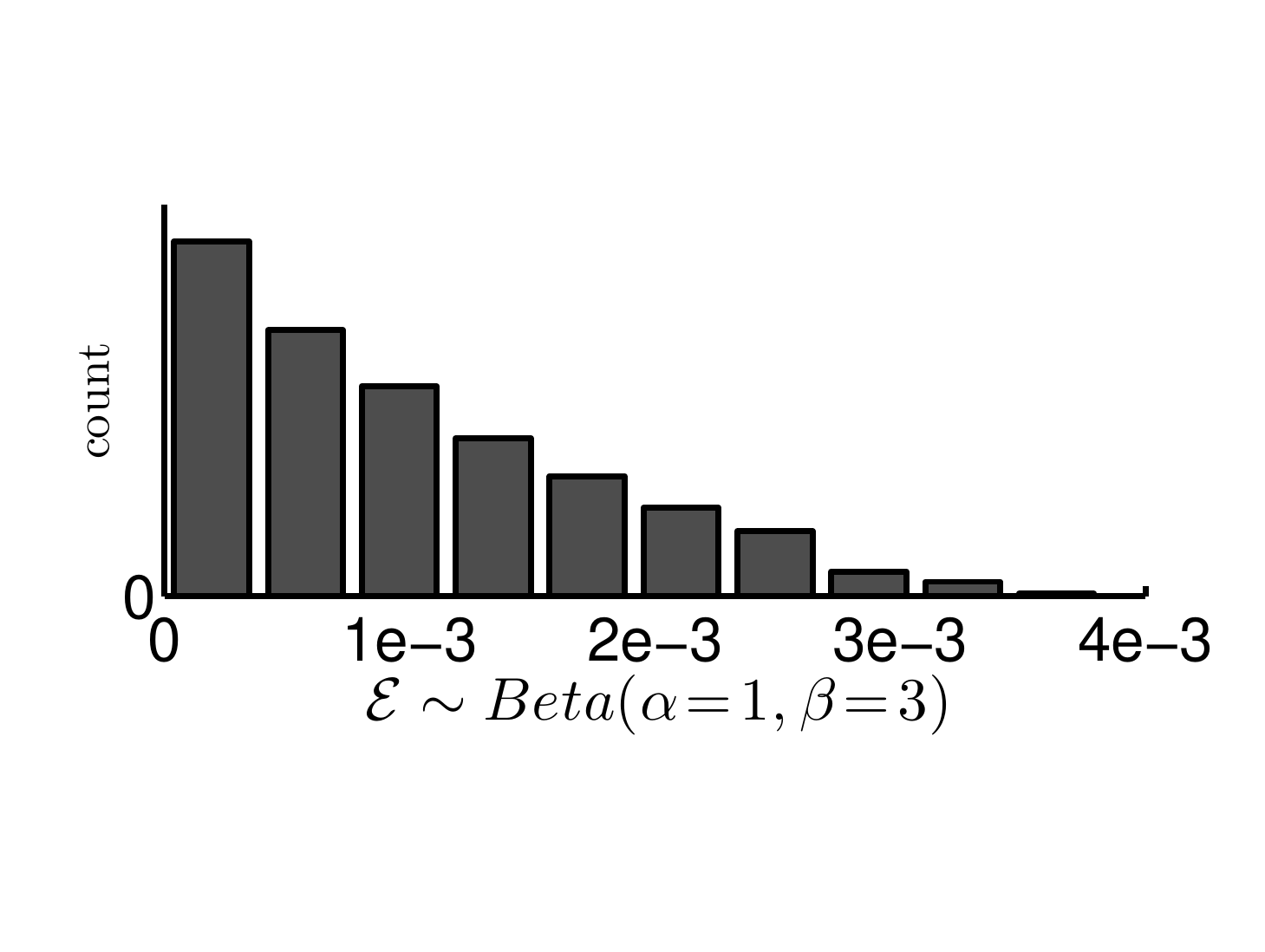}
\caption{Initial spectral dist.$\!\!\!\!\!\!\!\!$}\label{fig:dist1}
\end{subfigure}
\begin{subfigure}[t]{0.245\linewidth}
\centering\includegraphics[trim=0 0 0 0, clip=true, height=\PowH]{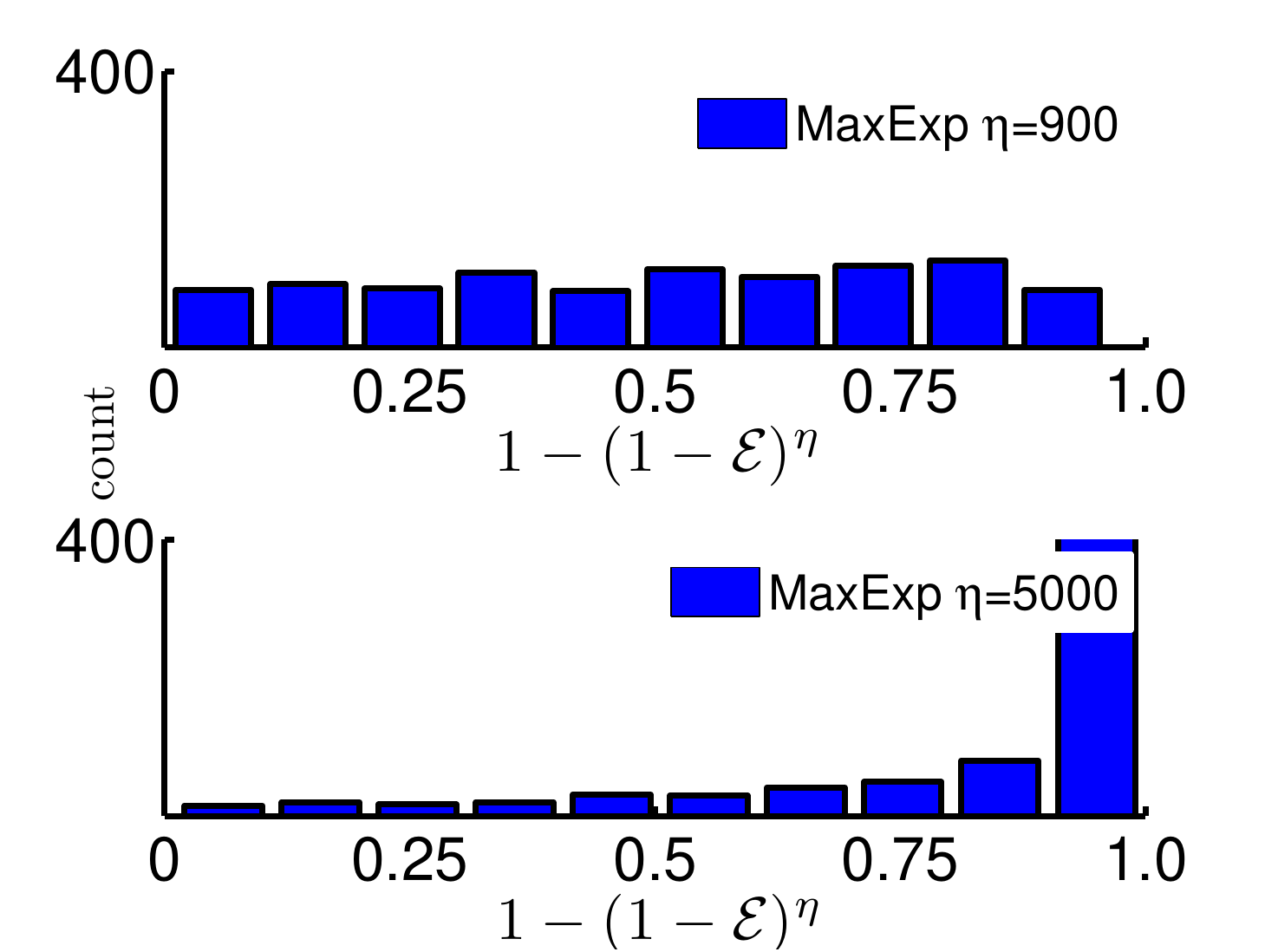}%\hspace{-0.5cm}
\caption{Pushforward (MaxExp)$\!\!\!\!\!\!\!\!$}\label{fig:dist2}
\end{subfigure}
\begin{subfigure}[t]{0.245\linewidth}
\centering\includegraphics[trim=0 0 0 0, clip=true, height=\PowH]{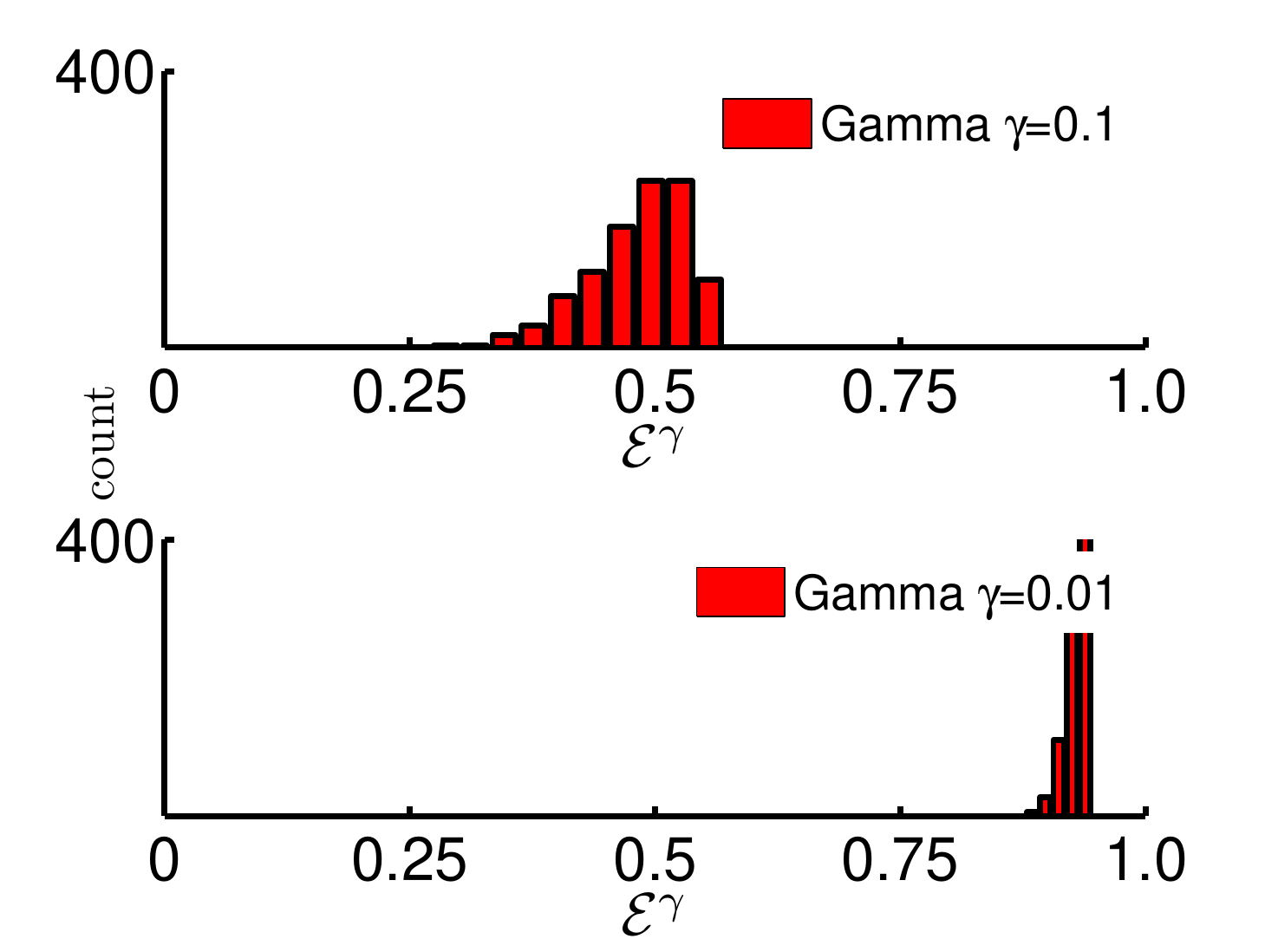}%\hspace{-0.8cm}
\caption{Pushforward (Gamma)$\!\!\!\!\!\!\!\!$}\label{fig:dist3}
\end{subfigure}
\begin{subfigure}[t]{0.245\linewidth}
\centering\includegraphics[trim=20 190 10 250, clip=true, height=\PowH]{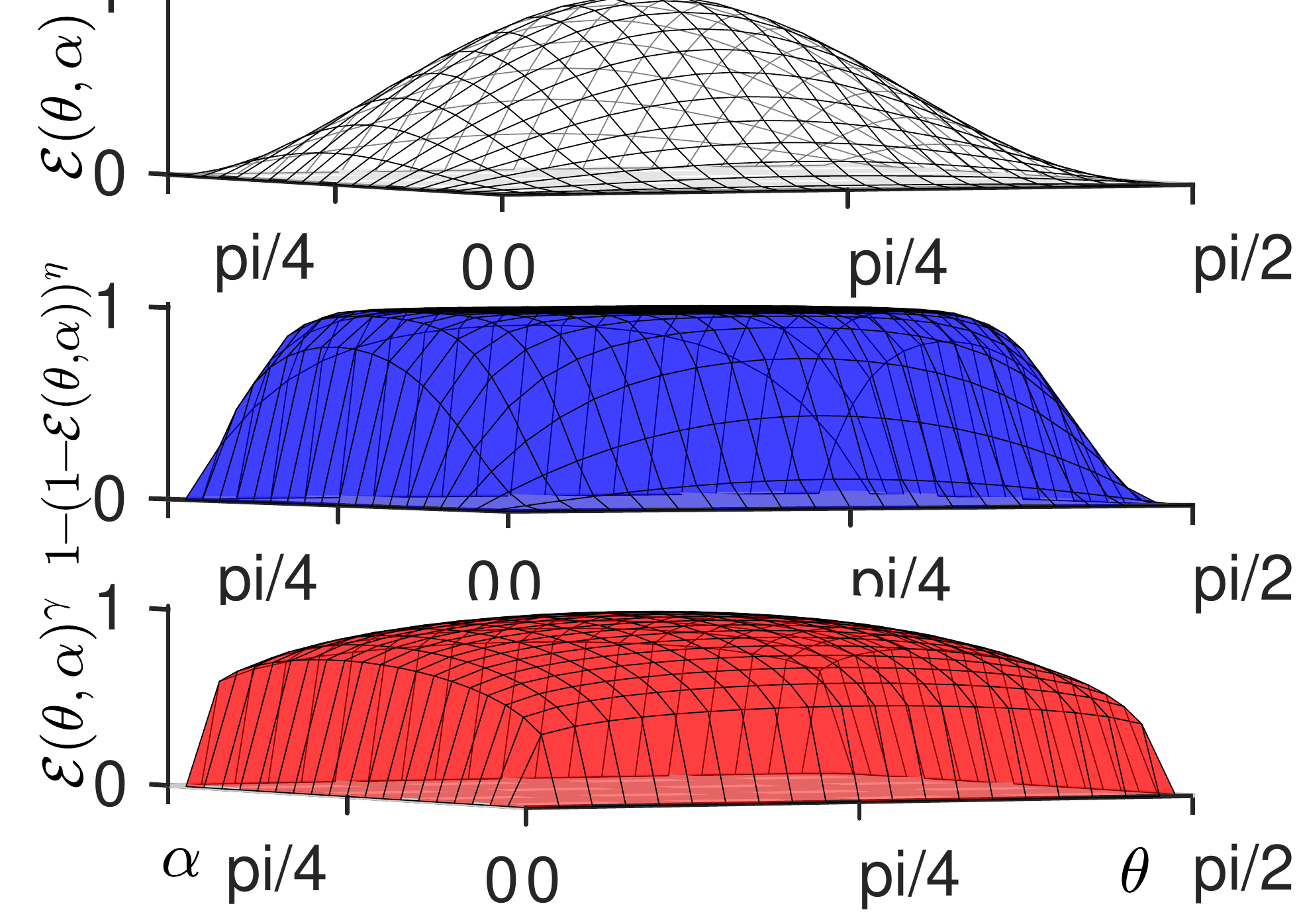}%\hspace{-0.8cm}
\caption{Spectral detectors$\!\!\!\!\!\!\!\!$}\label{fig:det1}
\end{subfigure}
\caption{\revised{The intuitive principle of the EPN. Given a discrete spectrum following a Beta distribution in Fig. \ref{fig:dist1}, 
the pushforward measures by MaxExp and Gamma in Fig. \ref{fig:dist2} and \ref{fig:dist3} are very similar for large $\eta$ (and small $\gamma$). Note that both EPN functions in bottom plots whiten the spectrum (map most values to be close to 1) thus removing burstiness. Fig. \ref{fig:det1} illustrates the principle of detecting higher-order occurrence(s) in one of $\binom{Z_*}{r}$  subspaces represented by $\cE_{\vu,\vv,\vw}$ (we write $\cE$ for simplicity). Fig. \ref{fig:det1} ({\em top}) No EPN: $\cE(\theta,\alpha)$, ({\em middle}) MaxExp: $1\!-\!(1\!-\!\cE(\theta,\alpha))^\eta$ and ({\em bottom}) Gamma: $\cE(\theta,\alpha)^\gamma$. Note how MaxExp/Gamma reach high detection values close to borders. Refer Section \ref{sec:epn_interp} for def. of $\cE(\theta,\alpha)$.}}
\vspace{-0.3cm}
\label{fig:epn-dist}
\end{figure*}

\vspace{-0.2cm}
\subsection*{C. Computational Complexity}
\label{sec:complexity}
Non-linearized SCK with ker. SVM have complexity $\bigoh(JN^2T^\rho)$ given $J$ body joints, $N$ frames per sequence, $T$ sequences, and $2\!<\!\rho\!<\!3$ which concerns complexity of kernel SVM. Linearized SCK with linear SVM takes $\bigoh(JNTZ_*^r)$ for total of $Z_*$ pivots and tensor of order $r\!=\!3$. Note that $N^2T^\rho\!\gg\!NTZ_*^r$. %For $N\!=\!50$ and $Z_*\!=\!20$, \eg~$Z_*\!=\!3Z_2\!+\!Z_3$ given $G_{\sigma'_2}$ and $G_{\sigma'_3}$, linearization is $3.5\!\times$ (or $32\!\times$) faster than the exact kernel if $T\!=\!557$ (or $T\!=\!5000$, respectively).
For $N\!=\!50$ and $Z_*\!=\!20$,  this is $3.5\!\times$ (or $32\!\times$) faster than the exact kernel for $T\!=\!557$ (or $T\!=\!5000$) used in our experiments.

Non-linearized DCK+kernel SVM enjoys $\bigoh(J^2N^4T^\rho)$ complexity. Linearized DCK+SVM enjoys $\bigoh(J^2N^2TZ^3)$ for $Z$ pivots per kernel, \eg~$Z\!=\!Z_2\!=\!Z_3$ given $G_{\sigma'_2}$ and $G_{\sigma'_3}$. As $N^4T^\rho\!\gg\!N^2TZ^3$, the linearization is $~11000\!\times$ faster than the exact kernel, for say $Z\!=\!5$. %for say $Z\!=\!5$ is $~11000\!\times$ faster than the exact kernel. 
Slice-wise EPN applied to SCK has negligible cost  $\bigoh(JTZ_*^{\omega+1})$, where $2\!<\!\omega\!<\!2.376$ concerns complexity of eigenvalue decomposition applied per tensor slice.

Note that EPN incurs negligible cost (see~\cite{tensor_arxiv} for details). 
EPN applied to DCK utilizes HOSVD and results in complexity $\bigoh(J^2TZ^4)$. As HOSVD is performed by truncated SVD on matrices obtained from unfolding $\tV_{ii'\!}\in\mbr{Z\times Z\times Z}\!$ along a chosen mode, $\bigoh(Z^4)$ represents the complexity of truncated SVD on matrices $\mV_{ii'\!}\in\mbr{Z\times Z^2}\!$ which have rank less than or equal $Z$.

Linearized SCK$\,\oplus$ with linear SVM also takes $\bigoh(JNTZ_*^r)$ for a total of $Z_*$. However, $Z_*\!=\!3Z_2\!+\!Z_3\!+\!Z_4\!+\!Z_5$ thus $Z_*\!=\!28$. The linearized DCK$\,\oplus$ takes $\bigoh(J^2N^2TZ^3Z_6)$ where $Z_6\!=\!4$ in our experiments. EPN applied to SCK$\,\oplus$ and DCK$\,\oplus$ results in complexity $\bigoh(JTZ_*^{2(r\!-\!1)})$ and $\bigoh(J^2TZ^4Z_6)$.

\vspace{-0.2cm}
\subsection*{D. What is (Tensor) Eigenvalue Power Normalization?}
\label{sec:epn_interp}

Below, we show that EPN in fact retrieves factors which quantify whether there is at least one  datapoint $\vphi({\vx_n})$ from $n\!\in\!\idx{N}$ projected into each subspace spanned by $r$-tuples of eigenvectors from matrices $\vec{A}_1\!=\!\vec{A}_2\!=\!\cdots\!=\!\vec{A}_r$. For brevity, assume order $r\!=\!3$, a super-symmetric case, and a $3$-tuple of eigenvectors $\vu$, $\vv$, and $\vw$ from $\vec{A}$. Note that $\vu\!\perp\!\vv, \vv\!\perp\!\vw$ and $\vu\!\perp\!\vw$. Moreover, note that if we have $Z_*\!$ unique eigenvectors, we can enumerate $\binom{Z_*}{r}$ $r$-tuples and thus $\binom{Z_*}{r}$ subspaces $\mbr{r}\!\subset\!\mbr{Z_*}$.
For brevity, let $||\vphi({\vx})||_2\!=\!1$ and $\vphi({\vx})\!\geq\!0$. Also, we write $\vphi_n$ instead of $\vphi({\vx})$ for $n\!\in\!\idx{N}$. Next, let us write our super-symmetric tensor as:
\begin{align}
& \tX\!=\!\frac{1}{N}\!\sum\limits_{n\in \idx{N}}{\kronstack}_r\,\vphi_n,
\label{eq:hod}
\end{align}
and the `diagonalization' of $\tX$ w.r.t. by eigenvec. $\vu$, $\vv$, and $\vw$ as:
\vspace{-0.45cm}
\begin{align}
& \cE_{\vu,\vv,\vw}=((\tX\otimes_{1}\!\vu)\,\otimes_{1}\!\vv)\otimes_{3}\!\vw,
\label{eq:hoddiag}
\end{align}
where $\cE_{\vu,\vv,\vw}$ is a coefficient from the core tensor $\tE$ for eigenvectors $\vu$, $\vv$, and $\vw$. 
Now, we combine Eq. \ref{eq:hod} and \ref{eq:hoddiag} and obtain:
\vspace{-0.4cm}
\begin{align}
& \cE_{\vu,\vv,\vw}=\bigg(\Big(\big(\frac{1}{N}\!\sum\limits_{n\in \idx{N}}{\kronstack}_3\,\vphi_n\big)\otimes_{1}\!\vu\Big)\!\otimes_{2}\!\vv\bigg)\!\otimes_{3}\!\vw\nonumber\\
& \qquad\quad=\frac{1}{N}\!\sum\limits_{n\in \idx{N}}\left<\vphi_n, \vu\right>\left<\vphi_n, \vv\right>\left<\vphi_n, \vw\right>
\label{eq:hoddiag2}
\end{align}

We assume $\vphi_n$ is projected into subspace spanned by $\vu,\vv$ and $\vw$ when $\psi'\!_n\!=\!\left<\vphi_n, \vu\right>\left<\vphi_n, \vv\right>\left<\vphi_n, \vw\right>$ is maximized. As our $\vu$, $\vv$, and $\vw$ are orthogonal w.r.t. each other and $||\vphi_n||_2\!=\!1$, simple Lagrange eq.  $\mathcal{L}\!=\!\Pi_{i\!=\!1}^r\vec{e}_i^T\!\vphi_n\!+\!\lambda(||\vphi_n||^2_2\!-\!1)$  yields maximum of $\kappa\!=\!(1/\sqrt{r})^r$ at $\vphi_n\!=\![(1/\sqrt{r}),\cdots,(1/\sqrt{r})]^T$. For each $n\!\in\!\idx{N}$, we store $\psi_n\!=\!\psi'\!_n/\kappa$ in vector $\vec{\psi}$.

\vspace{-0.05cm}
\begin{tcolorbox}[width=1.0\linewidth, colframe=blackish, colback=beaublue, boxsep=0mm, arc=3mm, left=1mm, right=1mm, right=1mm, top=1mm, bottom=1mm]
Assume that $\vec{\psi}\!\in\!\{0,1\}^{N}$ stores $N$ outcomes of drawing from Bernoulli distribution under the i.i.d. assumption for which the probability $p$ of an event $(\psi_n\!=\!1)$ and $1\!-\!p$ for $(\psi_n\!=\!0)$ is estimated as an expected value, $p\!=\!\avg_n\psi_n$ (even if $0\!\leq\!\vec{\psi}\!\leq\!1$ in reality). Then the probability of at least one projection event $(\psi_n\!=\!1)$ into the subspace spanned by $r$-tuples in $N$ trials becomes:

\vspace{-0.6cm}
{\fontsize{8}{9}\selectfont
\begin{equation}
\cEH_{\vu,\vv,\vw}\!=\!1\!-\!(1\!-\!p)^{N}=1\!-\!\left(1\!-\!\frac{\cE_{\vu,\vv,\vw}}{\kappa}\right)^{N}\!\!\!\!\approx\!\left(\frac{\cE_{\vu,\vv,\vw}}{\kappa}\right)^\gamma\!\!.
\label{eq:my_maxexp_tuples}
\end{equation}
}
\vspace{-0.4cm}

\noindent{Thus}, each of $\binom{Z_*}{r}$ subspaces spanned by $r$-tuples acts as a detector of projections into this subspace. The middle part of Eq. \eqref{eq:my_maxexp_tuples} (so-called MaxExp pooling) and its connection to the right-hand part of Eq. \eqref{eq:my_maxexp_tuples} (so-called Gamma) are detailed in \cite{koniusz2018deeper}. 
In fact, our $\vec{\psi}$ can be negative so extending Eq. \eqref{eq:my_maxexp_tuples} to $\sgn(\cE_{\vu,\vv,\vw})\Big(1\!-\!(1\!-\!\frac{|\cE_{\vu,\vv,\vw}|}{\kappa})^{N\!+\!\eta}\Big)$ makes our model a detector of asymmetry between projections into `positive' and `negative' parts of each subspace, and $\eta$ compensates for non-binary $\vec{\psi}$.
\end{tcolorbox}
\vspace{-0.15cm}

\revised{Figure \ref{fig:epn-dist} illustrates that MaxExp and Gamma are in fact very similar. Figure \ref{fig:dist1} shows an initial Beta distribution of spectrum. Figures \ref{fig:dist2} and \ref{fig:dist3} (bottom) show that for sufficiently large parameters $\eta$ and $\gamma$, both MaxExp and Gamma shift most of the distribution values to be approximately equal $1$. Figure \ref{fig:dist3} illustrates the effect of EPN on eigenvalue $\cE_{\vu,\vv,\vw}$ (denoted as $\cE$ for simplicity) representing a single subspace spanned by eigenvectors $\vu,\vv,\vw$ such that $\vu\!\perp\!\vv, \vv\!\perp\!\vw$ and $\vu\!\perp\!\vw$.  As a single projection into the subspace is defined as $\psi\!_n\!=\!\left<\vphi_n, \vu\right>\left<\vphi_n, \vv\right>\left<\vphi_n, \vw\right>/\kappa$, we note this is the product of three projections of $\vphi_n$ onto $\vu,\vv,\vw$, respectively, measured by the cosine (dot-product). Thus, we parametrize such a projection by the spherical coordinates, that is:
\vspace{-0.4cm}
\begin{align}
%&\pi_\vu(\theta,\alpha)\!=\!\cos(\theta)\!\cdot\!\sin(\alpha),\nonumber\\
%&\pi_\vv(\theta,\alpha)\!=\!\sin(\theta)\!\cdot\!\sin(\alpha),\nonumber\\
%&\pi_\vu(\alpha)\!=\!\cos(\alpha),
%
&\pi_\vu(\theta,\alpha)\!=\!\cos(\theta)\!\cdot\!\sin(\alpha),\,\pi_\vv(\theta,\alpha)\!=\!\sin(\theta)\!\cdot\!\sin(\alpha),\nonumber\\
&\pi_\vu(\alpha)\!=\!\cos(\alpha),
\end{align}

\noindent{where} the azimuthal coordinate $\theta$ runs from 0 to $2\pi$ and the polar coordinate $\alpha$ runs from 0 to $\pi$. We rewrite the projection as:
\begin{align}
&\pi_{\vu,\vv,\vw}(\theta,\alpha)\!=\!\pi_\vu(\theta,\alpha)\!\cdot\!\pi_\vv(\theta,\alpha)\!\cdot\!\pi_{\vw}(\alpha)/\kappa.
\end{align}
We note that $\pi_{\vu,\vv,\vw}(\theta,\alpha)$ and $\psi\!_n$ are isomorphic as $||\vphi_n||_2\!=\!1$, thus it suffices to note $\cE_{\vu,\vv,\vw}\!\sim\!\pi_{\vu,\vv,\vw}(\theta,\alpha)$ and show the EPN pushforward output of $\cE$ to understand how EPN behaves around the boundaries of the spanning vectors $\vu,\vv,\vw$.
Figure \ref{fig:det1} (top) shows that $\cE$ by itself has a weak response in the proximity of the spanning vectors $\vu,\vv,\vw$. However, MaxExp and Gamma in Figures \ref{fig:det1} middle and bottom manage to boost  projections in the proximity of the spanning vectors in the similar manner to each other, both behaving like spectral detectors.
}

To conclude, let us consider the dot-product between Power Normalized tensors $\tX$ and $\tY$ computed according to Eq. (\ref{eq:rawcod3}-\ref{eq:rawcod5}). Then:

\vspace{-0.55cm}
{\fontsize{8}{9}\selectfont
\begin{align}
& \!\!\!\!\left<\tVT(\tX),\tVT(\tY)\right>\!=\!\left<\!\sum\limits_{\substack{\vu\in\mU(\tX)\\\vv\in\mV(\tX)\\\vw\in\mW(\tX)}}\!\!\!\!\!\!\!\cEH_{\vu,\vv,\vw}\vu\vv^T\!\kronstack\!\vw,
\!\!\sum\limits_{\substack{\vu'\!\in\mU(\tY)\\\vv'\!\in\mV(\tY)\\\vw'\!\in\mW(\tY)}}\!\!\!\!\!\!\!\cEH'_{\vu'\!,\vv'\!,\vw'}\vu'\!\vv'^T\!\kronstack\!\vw'\!
 \right>\nonumber\\
& \qquad\qquad\qquad\!\!=\!\!\!\sum\limits_{\substack{\vu\in\mU(\tX)\\\vv\in\mV(\tX)\\\vw\in\mW(\tX)}}\sum\limits_{\substack{\vu'\!\in\mU(\tY)\\\vv'\!\in\mV(\tY)\\\vw'\!\in\mW(\tY)}}\!\!\!\!\!\!\!\cEH_{\vu,\vv,\vw}\cEH'_{\vu'\!,\vv'\!,\vw'}\,\left<\vu,\vu'\right>\left<\vv,\vv'\right>\left<\vw,\vw'\right>.\nonumber\\[-16pt]
&\label{eq:my_tens_subsp}
\end{align}
}

\vspace{-0.45cm}
\noindent{Eq.} \eqref{eq:my_tens_subsp} shows that all subspaces of $\tX$ and $\tY$ spanned by $r$-tuples ($3$-tuples in this example) are compared against each other for alignment by the cosine distance. When two subspaces $[\vu\,\vv\,\vw]^T$ and $[\vu'\,\vv'\,\vw']^T$ are aligned, for a strong similarity between these subspaces, a detection of at least one $\vphi_n$ and $\vphi'_n$ evidenced by $\cEH_{\vu,\vv,\vw}$ and $\cEH'_{\vu'\!,\vv'\!,\vw'}$ is also needed. We term Eq. \eqref{eq:my_tens_subsp} together with Eq. (\ref{eq:rawcod3}-\ref{eq:rawcod5}) as Tensor Power Euclidean dot-product which has the  associated Tensor Power Euclidean metric $||\tX\!-\!\tY||_{\mathcal{T}}\!=\!||\tVT(\tX)-\tVT(\tY)||_F$.

%\end{appendices}

%\IEEEraisesectionheading{\section{Introduction}\label{sec:introduction}}

%\ifCLASSOPTIONcaptionsoff
%  \newpage
%\fi

\vspace{-0.2cm}
{\small
\bibliographystyle{IEEEtran}
\bibliography{egbib}
}

\vspace{-1.13cm}
\begin{IEEEbiography}[{\includegraphics[width=1in,height=1.25in,clip,keepaspectratio]{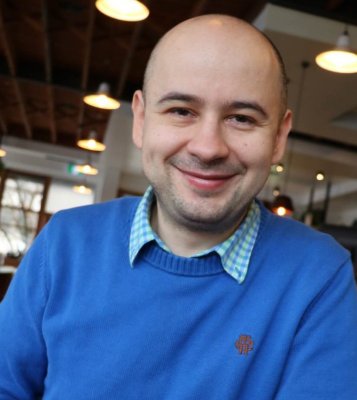}}]{Piotr Koniusz.}
A Senior Researcher in Machine Learning Research Group at Data61/CSIRO (NICTA), and a Senior Honorary Lecturer at the Australian National University (ANU). He was a postdoctoral researcher in the team LEAR, INRIA, France. He received his BSc in Telecommunications and Software Engineering in 2004 from the Warsaw University of Technology, Poland, and completed his PhD in Computer Vision in 2013 at CVSSP, University of Surrey, UK. 
%His interests include visual categorization, (kernel) learning on graphs/tensors and deep learning.
\end{IEEEbiography}

\vspace{-1.42cm}
\begin{IEEEbiography}[{\includegraphics[width=1in,height=1.25in,clip,keepaspectratio]{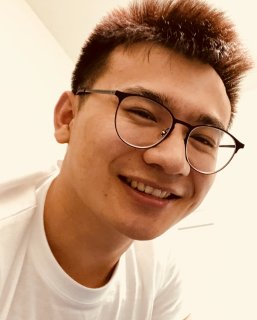}}]{Lei Wang.}
He received the M.E. degree in software engineering from The University of Western Australia (UWA), Australia, in 2018.
Since then, he has worked as a Computer Vision Researcher at iCetana Pty Ltd. He is currently a PhD student at the Australian National University and Data61/CSIRO under the supervision of Dr. Piotr Koniusz. 
His research interests include human action recognition in videos, machine learning and computer vision.
\end{IEEEbiography}

\vspace{-1.45cm}
\begin{IEEEbiography}[{\includegraphics[width=1in,height=1.25in,clip,keepaspectratio]{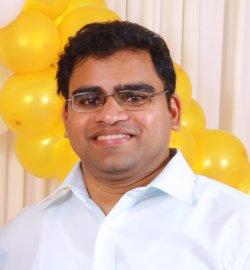}}]{Anoop Cherian.}
A Research Scientist at Mitsubishi Electric Research Labs (MERL) Cambridge, MA and an Adjunct Researcher at the Australian Centre for Robotic Vision (ACRV) at the Australian National University, Canberra.  Before joining ANU, he was a postdoctoral researcher in the LEAR team at INRIA, Grenoble.  He received my MS and PhD in 2010 and 2013 from the University of Minnesota, Minneapolis, USA. He received his undergraduate (honors) degree in computer science and engineering at the National Institute of Technology (NIT), Calicut, India in 2002.
\end{IEEEbiography}

\end{document}